%% file: cvpr.tex
\documentclass[final]{cvpr}

\usepackage{times}
\usepackage{epsfig}
\usepackage{graphicx}
\usepackage{amsmath}
\usepackage{amssymb}
\usepackage{multirow}
\usepackage{makecell}
\usepackage{booktabs}
\usepackage{nopageno}

\usepackage[pagebackref=true,breaklinks=true,colorlinks,bookmarks=false]{hyperref}
\makeatletter
\def\blfootnote{\gdef\@thefnmark{}\@footnotetext}
\makeatother

\input{vcl-shortcuts.tex}
\def\cvprPaperID{4173} 
\def\confYear{CVPR 2021}

\begin{document}

\title{Neural Camera Simulators}

\author{
Hao Ouyang*\\
HKUST

\and
Zifan Shi*\\
HKUST

\and
Chenyang Lei\\
HKUST

\and
Ka Lung Law\\
SenseTime

\and
Qifeng Chen\\
HKUST
}

\maketitle

\blfootnote{*Joint first authors}

\begin{abstract}

We present a controllable camera simulator based on deep neural networks to synthesize raw image data under different camera settings, including exposure time, ISO, and aperture.  The proposed simulator includes an exposure module that utilizes the principle of modern lens designs for correcting the luminance level. It also contains a noise module using the noise level function and an aperture module with adaptive attention to simulate the side effects on noise and defocus blur.  To facilitate the learning of a simulator model, we collect a dataset of the 10,000 raw images of 450 scenes with different exposure settings.  Quantitative experiments and qualitative comparisons show that our approach outperforms relevant baselines in raw data synthesize on multiple cameras. Furthermore, the camera simulator enables various applications, including large-aperture enhancement, HDR, auto exposure, and data augmentation for training local feature detectors. Our work represents the first attempt to simulate a camera sensor's behavior leveraging both the advantage of traditional raw sensor features and the power of data-driven deep learning. 
The code and the dataset are available at \href{https://github.com/ken-ouyang/neural_image_simulator}{https://github.com/ken-ouyang/neural\_image\_simulator}.

\end{abstract}

\vspace{-2mm}

\section{Introduction}

\begin{figure}[t!]
\centering
\hspace*{-2mm}
\includegraphics[width=1.02\linewidth]{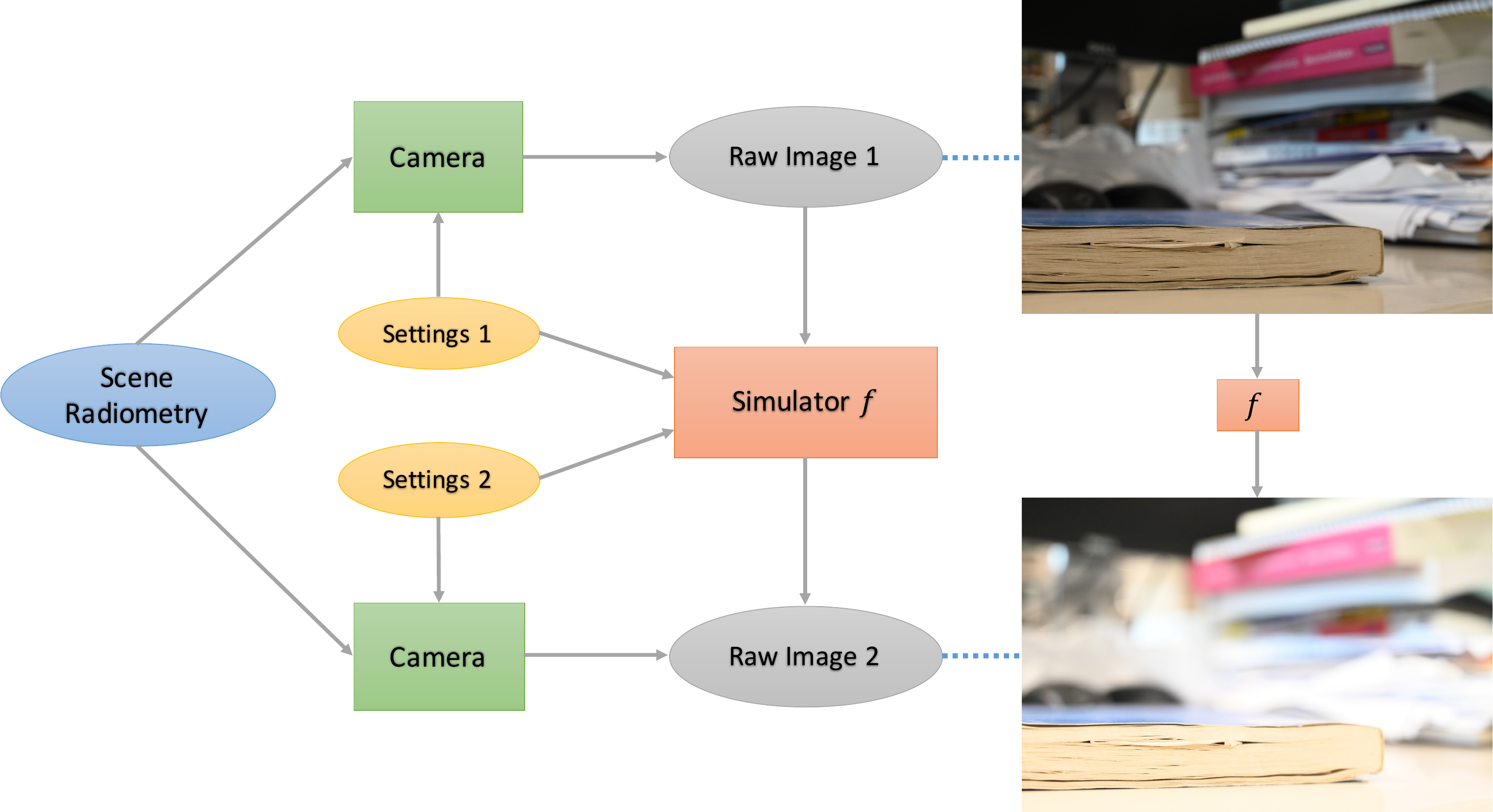}
\vspace{-2mm}
\caption{The problem formulation of simulating images with different camera settings. Given the same scene, a camera can capture different images with different camera settings, where the degrees of luminance, noise, and blur vary a lot. Our simulator aims to model the mappings between raw images with different camera settings.}
\label{fig:intro}
\end{figure}

Controllable photo-realistic image generation is a new trending research topic~\cite{tewari2020state}. Most recent works focus on learning a certain scene representation conditioned on different viewpoints~\cite{mildenhall2020nerf, meshry2019neural} or lighting~\cite{sun2019single, zhou2019deep}. However, in the physical image formulation pipeline, apart from scene, light, and view angle, the camera settings are also important components, which are yet to be investigated. Modern cameras introduce various settings for capturing sceneries, among which the exposure settings (i.e., exposure time or shutter speed, ISO, and aperture size) are most commonly adjusted. As in Fig.~\ref{fig:intro}, cameras capture the scene radiometric characteristics and record them as raw image data. Capturing with different exposure settings not only leads to luminance changes but also results in different side effects: ISO affects the noise presented, and aperture size decides the defocus blur. This paper studies a challenging new task of controllable exposure synthesis with different camera settings. Given the original image, we learn a simulator $f$  that maps from the old settings to new settings. To capture the essence of changing exposure settings, we directly explore our simulation on raw data rather than monitor-ready images so that any camera signal processing pipeline can be applied~\cite{karaimer2016software}.

Several features presented in raw data are beneficial for the exposure simulation: 
\begin{itemize}
\renewcommand{\labelitemi}{\textbullet}
  \item  Raw sensor data directly captures the physical information, such as the light intensity of a scene. To simulate a new illuminance level, we can analyze the change of captured light intensity under new settings based on the physical prior and the design of the modern lens~\cite{smith2005modern}. 
  \item  The noise distribution model on raw sensor data is relatively robust because of the physical imaging model~\cite{foi2008practical}.
  \item   In a close shot scene, the original divisions of blurry and sharp regions provide additional supervision information. It is possible to learn and locate the blurry regions to magnify the defocus blur.
\end{itemize}
Therefore, we propose a model that consists of three modules, which progressively learn the change of exposure, noise, and defocus blur. We first adopt an exposure correction module by analyzing the light intensity and refine it using linear regression. In the second module, we utilize the traditional noise-level-function model~\cite{foi2009clipped,foi2008practical}, and train a deep network to further modify the distribution based on real data. Finally, we propose a new attention module to focus on the blurry regions for aperture enhancement.  In this paper, we do not consider cases such as removing defocus blur (from large to small aperture in the close shot) or motion blur (with moving objects). They are too scene-dependent to find a general representation.

Since no existing dataset contains enough image sequences that are captured at the same scene with different settings, we collect around 10,000 images of 450 such sequences. The dataset is collected with two modern cameras in diverse environments. Extensive experiments on this dataset demonstrate that our proposed model can progressively generate both visually and statistically similar results at new exposure settings. 

The simulator can benefit many low-level computer vision tasks such as generating high-dynamic-range (HDR) photos and enhancing defocus blur. It may also generate realistic images for data augmentation in training deep neural networks. We demonstrate four applications of the simulator: magnifying the defocus blur, generating HDR images,  providing environments for training auto-exposure algorithms, and data augmentation for training local feature detectors. The contribution of this paper can be summarized as:
\begin{itemize}
\renewcommand{\labelitemi}{\textbullet}
  \item To the best of our knowledge, we are the first to systematically study the problem of controllable image generation with camera exposure settings, which are important components in the physical image formulation pipeline. By leveraging the physical imaging prior and the power of data-driven deep learning, the model achieves better results than all baselines. 
  \item We demonstrate four applications (large-aperture enhancement, HDR, auto-exposure, and data augmentation) using the proposed simulator.
  \item  We collect a large dataset of raw data sequences of the same scenes under different settings. We believe it can benefit many other computer-vision tasks.
\end{itemize}

\section{Related Work}
\subsection{Raw data processing and simulation}
Learning-based raw data processing has attracted lots of research interests in recent years~\cite{chen2018learning,xu2019towards,zhang2019zoom}. Most of the newly proposed methods focus on generating higher quality monitor-ready images by imitating specific camera image signal processors (ISP). Researchers have explored utilizing a collection of local linear filter sets~\cite{jiang2017learning,lansel2014learning} or deep neural networks~\cite{jiang2017learning} to approximate the complex nonlinear operations in traditional ISP. Several follow-up works extend the usage of raw image data to other applications such as low-light image enhancement~\cite{chen2018learning}, real scene super-resolution~\cite{xu2019towards}, optical zoom adjustment~\cite{zhang2019zoom} and reflection removal~\cite{lei2020polarized,Lei_2021_RFC}. However, a real camera ISP varies with the camera's exposure settings~\cite{karaimer2016software}, which further increases the complexity and leads to the divergence in the training of the learning-based method. In this paper, we focus on the raw-to-raw data generation to make sure that errors in estimating the complicated camera ISP will be excluded.

Previous works have also proposed several ways to synthesize raw data. Gharbi et al.~\cite{gharbi2016deep} propose to follow the Bayer pattern to assign color pixels from the sRGB images to synthesize mosaicked images. Attracted by the more robust noise model in raw images, Brooks et al.~\cite{brooks2019unprocessing} propose to reverse the camera ISP to synthesize raw data for training denoise. However, reverting images from sRGB space is not suitable in our case because guessing the complicated ISP steps (especially the tone curve) introduces additional inaccuracies in simulating the illuminance level. Abdelhamed et al.~\cite{abdelhamed2019noise} utilize a flow-based method to synthesize more realistic noise based on camera ISO. Instead of focusing only on noise, our model explores raw data generation in a more general way, which also includes exposure and defocus manipulation. Researchers have also proposed several traditional methods~\cite{liu2021isetauto, liu2020neural, lian2018image, farrell2012digital, farrell2003simulation} to simulate images based on camera parameters.  Farrell et al.~\cite{farrell2012digital, farrell2003simulation} study the image formulation pipelines to directly infer the simulation for high-quality images for display. Liu et al. ~\cite{liu2020neural} explore simulation on pixel sizes, color filters, acquisition, and post-acquisition processing to enhance performance in autonomous driving tasks. On the other hand, our work focuses on combining traditional simulation and data-driven deep learning approaches to simulate challenging cases such as defocus blur.  

\subsection{Aperture supervision}
Researchers have proposed a variety of methods for enhancing defocus blur.  The most common practice is first to estimate a blurriness map~\cite{bae2007defocus,tai2009single,tang2013defocus,ma2018deep,park2017unified} and then refine and propagate it to other pixels.  The blurriness map can be retrieved by estimating blur kernels~\cite{bae2007defocus}, analyzing gradients~\cite{tai2009single}, or learning using deep networks on large-scale datasets~\cite{ma2018deep}. Several follow-up works focus on detecting just-noticable-blur~\cite{shi2015just} or blur in small regions~\cite{shi2015break} by analyzing small patches. Other methods~\cite{chen2017automatic,farhat2017intelligent} first learn to segment the foreground objects segmentation and followed by re-blurring. Defocus magnification is also associated with depth estimation.  Some works~\cite{barron2015fast, wang2018deeplens, wadhwa2018synthetic} render large-aperture images by estimating high-quality depth images as blurriness maps from a single image or stereo image pairs. Srinivasan et al.~\cite{srinivasan2018aperture} collect a large dataset of aperture stacks on flowers for inferencing the monocular depth. Instead of estimating the accurate depth value, we train models on image pairs with different apertures for learning how to locate the blurry regions. Moreover, the magnification level on the defocus blur is directly related to the value of the new aperture settings.

\begin{figure*}[!t]
\centering
\includegraphics[width=0.99\linewidth]{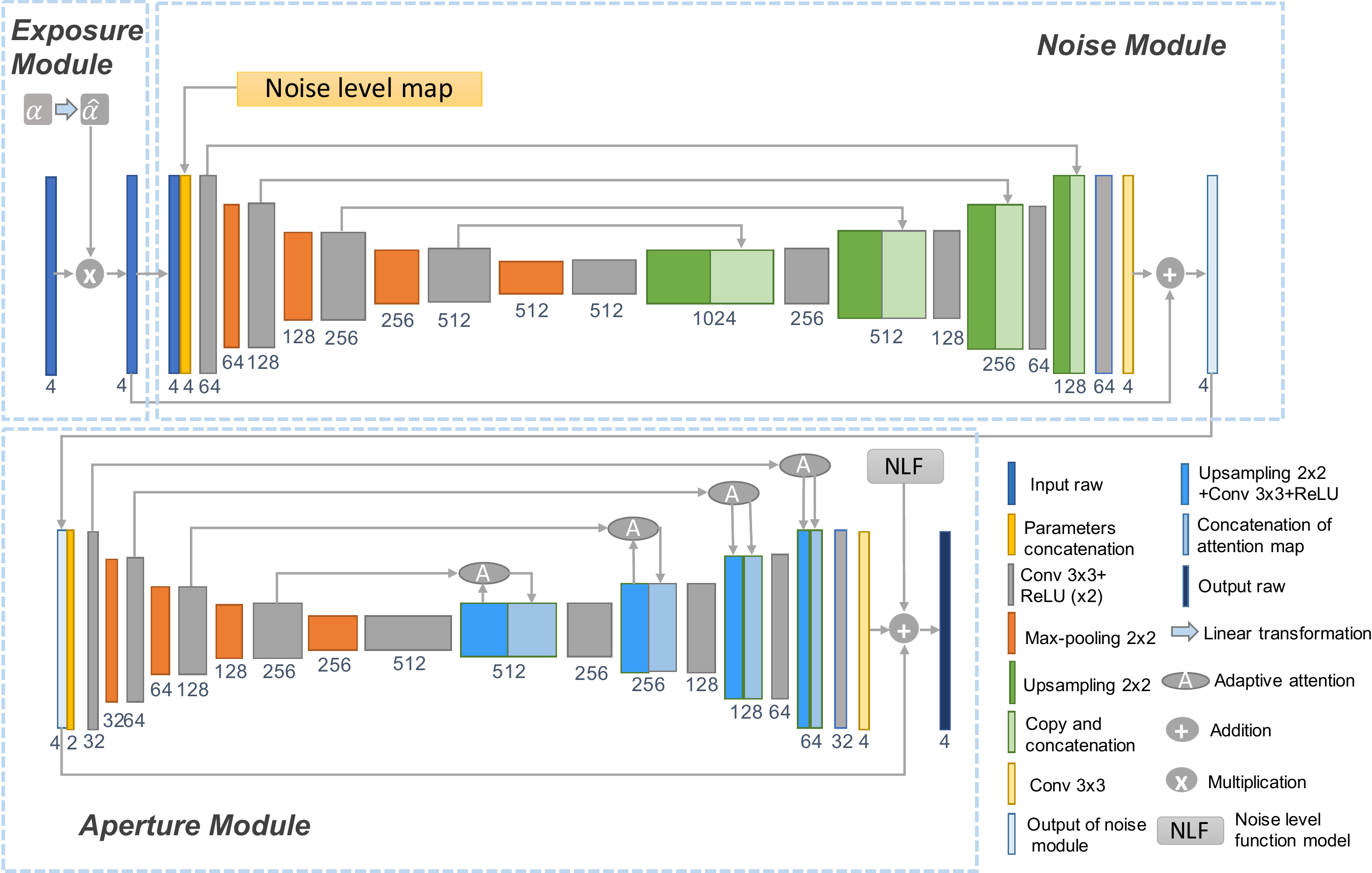}
\caption{Our simulator synthesizes images under different camera settings based on three modules: exposure module, noise module, and aperture module. The exposure module applies a linear transformation on the input and output camera settings, and thus bringing the brightness of input and output raw to the same level. Noise module takes noise level map of input raw as additional input channels and aims to remove the noise. The aperture module utilizes the input and output aperture parameters to simulate the side effect caused by the aperture. Finally, we adopt the NLF~\cite{foi2008practical} model to model the noise distribution under the output camera setting.}
\label{fig:method1}
\end{figure*}
\section{Method}

The overview of our framework is shown in Fig.~\ref{fig:method1}. Please refer to the caption for the general pipeline.  Note that we adopt the color processing pipeline from~\cite{karaimer2016software} for visualizing raw images in sRGB color space.

\subsection{Exposure module}
Our exposure correction module aims at adjusting the luminance level of input raw $I_1 \in\mathbb{R}^{\frac{H}{2} \times \frac{W}{2} \times 4}$ (The raw image is unpacked to a four-channel image where the width and height is half as original W and H and normalized between 0 and 1 after black-level subtraction) to the same level of the output $I_{2} \in \mathbb{R}^{\frac{H}{2} \times \frac{W}{2} \times 4}$. Unlike images in sRGB color space, raw images are linearly dependent on the received photon number. Adjusting camera settings also introduces an approximately proportional relationship on the luminance intensity between two images~\cite{ray2000manual}. Then we respectively analyze the effect of each camera parameter:

\mypara{Exposure time} controls the length of time when the shutter is open~\cite{allen2012manual}. The number of photons that reach the sensor is proportional to the time. Given the input and output exposure time $t_1$ and $t_2$, the multiplier $\alpha_{t}$ is defined as $t_2 / t_1$.

\mypara{ISO} controls the light sensitivity of the sensor. Doubling the sensitivity also doubles the recorded light value~\cite{ray2000manual}. Given the input and output ISO $g_1$ and $g_2$, the multiplier $\alpha_{g}$ is defined as $g_2 / g_1$.

\mypara{Aperture size} controls the area of the hole which the light can travel through~\cite{ray2000manual}. F-number $n$ is usually adopted to represent the size and $\frac{1}{n^2}$ is proportional to the hole area. Given input and output f-numbers $n_1$ and $n_2$, the multiplier $\alpha_{n}$ is defined as $n_1^2 / n_2^2$.

In image metadata, the aperture f-number is already approximated (e.g., $\sqrt{2}^3$ recorded as 2.8). We may further retrieve the original input and output exposure stop $s_1$ and $s_2$ for a more accurate estimation. We have $ s_1 = Round(6\log_2(n_1))$ and $ s_2 = Round(6\log_2(n_2))$ and the final $\alpha_{n} = 2^{(s_1-s_2)/3}$ (details in the \textbf{supplement}).

Combining these three parameters, we get the final $\alpha$: 
\begin{align}
  \alpha = \alpha_{t}\alpha_{g}\alpha_{n}.
\end{align}
The calculated $\alpha$ may not be accurate enough since camera settings such as exposure time is not perfectly controlled and has a biased shift. Following the formulation in~\cite{abdelhamed2018high}, we can adopt a linear model to further improve the accuracy of our observed data:
\begin{align}
  \hat \alpha &= w\alpha, \\
  I_{exp} &= clip((I_1 + b)* \hat \alpha ),
\end{align}
where $clip=\min(\max(y,0),1)$ and $I_{exp}$ is the output of the exposure stage. We initialize $w$ to 1 and $b$ to 0. Note that b is for compensating the inaccuracy of black level.  We optimize exposure module using the $L_1$ loss $||I_{exp} - I_2||_1$.

\mypara{Over-exposed regions} When input $I_1$ contains pixels that are over-exposed, the real value can be much higher than the recorded value. In this case, the multiplier proposed in the above part does not apply. Thus when training, we exclude pixels with values higher than 0.99.

\subsection{Noise module}
Though noise distribution in a monitor-ready image in sRGB color space can be very complicated because of the non-linear post-processing and compression, the noise in raw color space is a well-studied problem. Noise in raw data is usually represented by heteroscedastic Gaussian composed of two elements: shot noise due to the photon arrival and read noise caused by readout circuitry~\cite{foi2008practical}. Recent works in denoising proves the effectiveness of using this noise level function (NLF) model for noise synthesis~\cite{brooks2019unprocessing}. We also adopt this noise model as a prior in our noise module. We first reduce the primary noise in the $I_{exp}$ and then add back the NLF noise. Modern cameras provide calibrated NLF parameters for each ISO settings under normal temperature. The input read noise $\lambda_{read}^1$, input shot noise $\lambda_{shot}^1$, output read noise $\lambda_{read}^2$ and output shot noise $\lambda_{shot}^2$ can be directly retrieved from the camera metadata. We can represent the observed pixel value $y_1$ and $y_2$ in input and output raw data using the true signal $x_1$ and $x_2$:
\begin{align}
  y_1 &  \sim \mathcal{N}(\mu = x_1, \sigma_1^2 = \lambda_{read}^1+\lambda_{shot}^1x_1), \\
  y_2 &  \sim \mathcal{N}(\mu = x_2, \sigma_2^2 = \lambda_{read}^2+\lambda_{shot}^2x_2),
\end{align}
where $\mathcal{N}$ represents Gaussian distribution. After the exposure correction step, the distribution of pixels in $I_{exp}$ becomes: 
\begin{equation}
    \hat y_1 \sim \mathcal{N}(\mu = \hat \alpha (x_1+b), \hat \sigma_1^2 = \hat \alpha^2 \lambda_{read}^1+ \hat \alpha \lambda_{shot}^1x_1).
\end{equation}
We approximately estimate the noise level map of the input noise and output noise based on the first-stage output $I_{exp}$ and the modified input noise parameters $\hat \sigma_1$. The noise level map is concatenated to the input $I_{exp}$ as an additional prior. We adopt the same U-net structure used in~\cite{brooks2019unprocessing} and initialize it using their pre-trained model. As suggested by~\cite{zhang2017beyond}, we empirically find that using residual $I_{2}-I_{exp}$ is more effective in learning the high-frequency information. $L_1$ loss $||I_{rs}-(I_{2}-I_{exp})||_1$ is adopted for this module. We add back $I_{exp}$ to the learned residual $I_{rs}$ for retrieving the final output $I_{ns}$ in this stage. When training this stage, we adopt a special pair selection strategy to ensure that the output raw data contains little noise.

\begin{figure}[t!]
\centering
\hspace*{-2.5mm}
\includegraphics[width=1.05\linewidth]{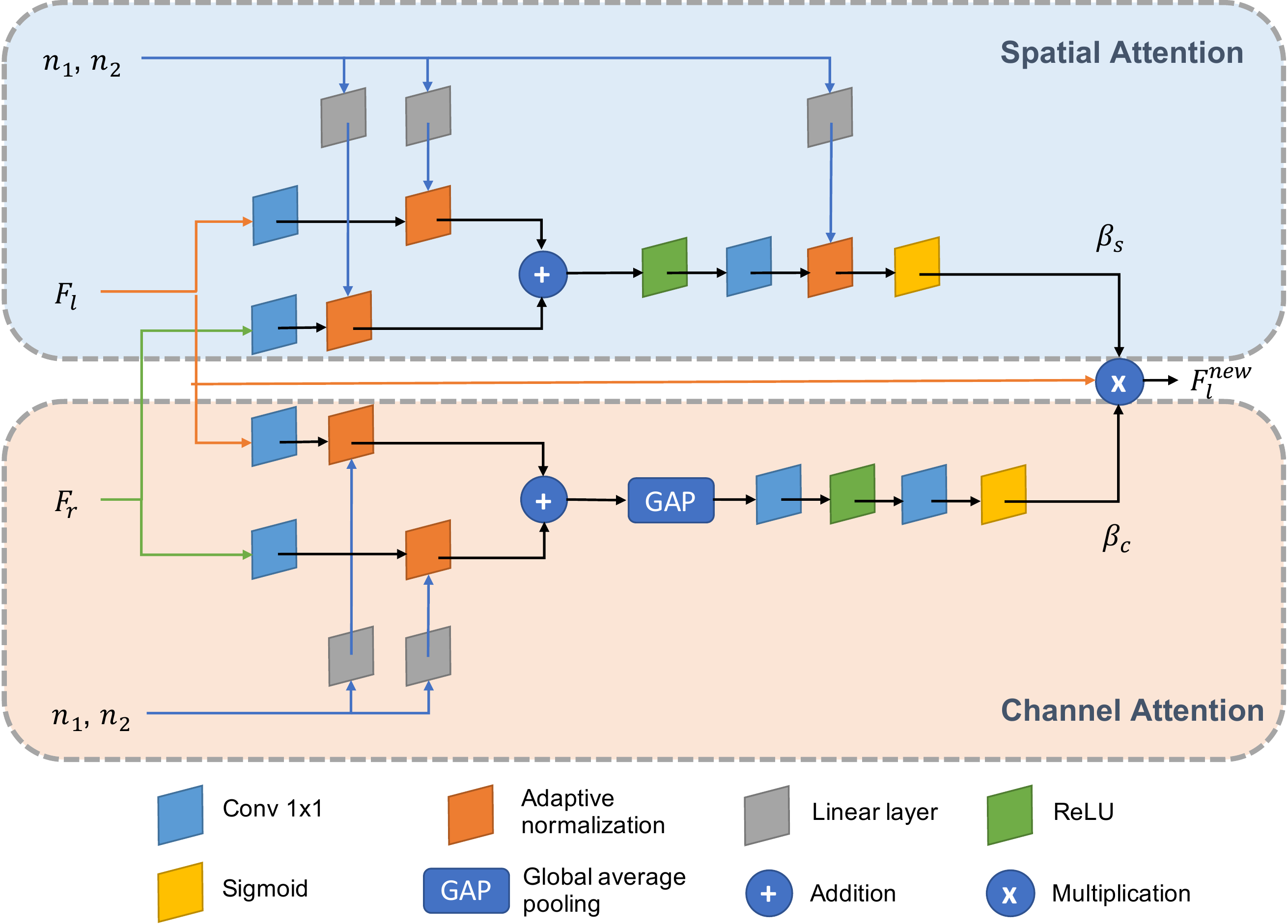}
\caption{Adaptive attention module comprises two parts: spatial attention module and channel attention module. The input and output aperture f-numbers are used to adjust the parameters in the normalization layer.}
\label{fig:attention}
\end{figure}
\subsection{Aperture module}

Increasing aperture size not only brightens the images but also enlarges the defocus blur because it shallows the depth-of-field. Recovering details from defocus blur highly depend on scene content. However, its inverse direction, which is also known as defocus magnification, is learnable by enhancing only the blurry regions. Our aperture model is designed to simulate the blur circle according to the new aperture settings. We adopt an adaptive attention module, as shown in Fig.~\ref{fig:attention}, which calibrates the features by considering spatial attention and channel attention at the same time. Uniquely, this module takes the input and output aperture f-number $n_1$ and $n_2$ to adjust the parameters of the normalization layer. The adaptive aperture layer is formulated as:
\begin{align}
   \mathcal{A}(x) = & (w_{\sigma}[n_1,n_2]+b_{\sigma})\left(\frac{x-\mu(x)}{\sigma(x)}\right) + (w_{\mu}[n_1,n_2]+b_{\mu}),
\end{align}
where $w_{\sigma}$, $w_{\mu}$, $b_{\sigma}$ and $b_{\mu}$ are parameters to be learnt for linear transformations. $\mu$ and $\sigma$ are mean and variance respectively. ''[ ]" denotes the concatenation of f-numbers. $x$ is the instance in the feature maps. 

Since we built adaptive attention module on a U-net architecture, the inputs of the adaptive attention module are original feature maps $F_{l}\in \mathbb{R}^{H_{l} \times W_{l} \times C_{l}}$ and gating signal $F_{r}\in \mathbb{R}^{H_{r}\times W_{r}\times C_{r}}$. Utilizing the above adaptive aperture layer, we can design the spatial and channel-wise attention in our case. 

Spatial attention module explores the inter-pixels relationships in feature maps $F_{l}$ and $F_{r}$, and computes an attention map $\beta_{s}$ to rescale feature maps $F_{l}$. We first transform $F_{l}$ and $F_{r}$ independently with a $1\times 1$ convolutional layer and an aperture layer. After a ReLU activation on the addition of the transformed feature maps, another $1\times 1$ convolutional layer and aperture layer are applied to get the final spatial attention map $\beta_{s}$ followed by a sigmoid activation.

In the channel attention module, channel-wise relationships are captured to rescale $F_{l}$ in the channel dimension. After the linear transformations on $F_{l}$ and $F_{r}$ same as above, a global average pooling layer is used to squeeze the feature maps to a feature vector $v$ with $C_l$ dimension. The excitation on $v$ is completed by two convolutional layers, a ReLU activation, and a sigmoid activation, which gives us the channel attention map $\beta_{c}$. With the spatial and channel attention maps, $F_{l}$ will be calibrated as:
\begin{align}
  F_{l}^{new}=F_{l}\times \beta_{s}\times \beta_{c}.
\end{align}
The modified feature $F_{l}^{new}$ can provide effective information when concatenating with up-sampled features in the U-net structure. The input and output f-numbers are concatenated as additional channels to the output $I_{ns}$ in the noise module.  We also adopt residual learning~\cite{zhang2017beyond} and utilize $L_1$ loss in this module.  Adding the residual learned in the aperture module back to $I_{ns}$ gives us the final output of the aperture module. To complete the entire simulation process, we add the synthesized output NLF noise to the output raw data.

\begin{figure*}[!t]
    \centering 
    \begin{tabular}{@{}c@{\hspace{0.8mm}}c@{\hspace{0.8mm}}c@{\hspace{0.8mm}}c@{\hspace{0.8mm}}c@{}}
     Input& Exposure Module & Noise Module & Full Model & Ground Truth \\
     \includegraphics[width=0.19\linewidth]{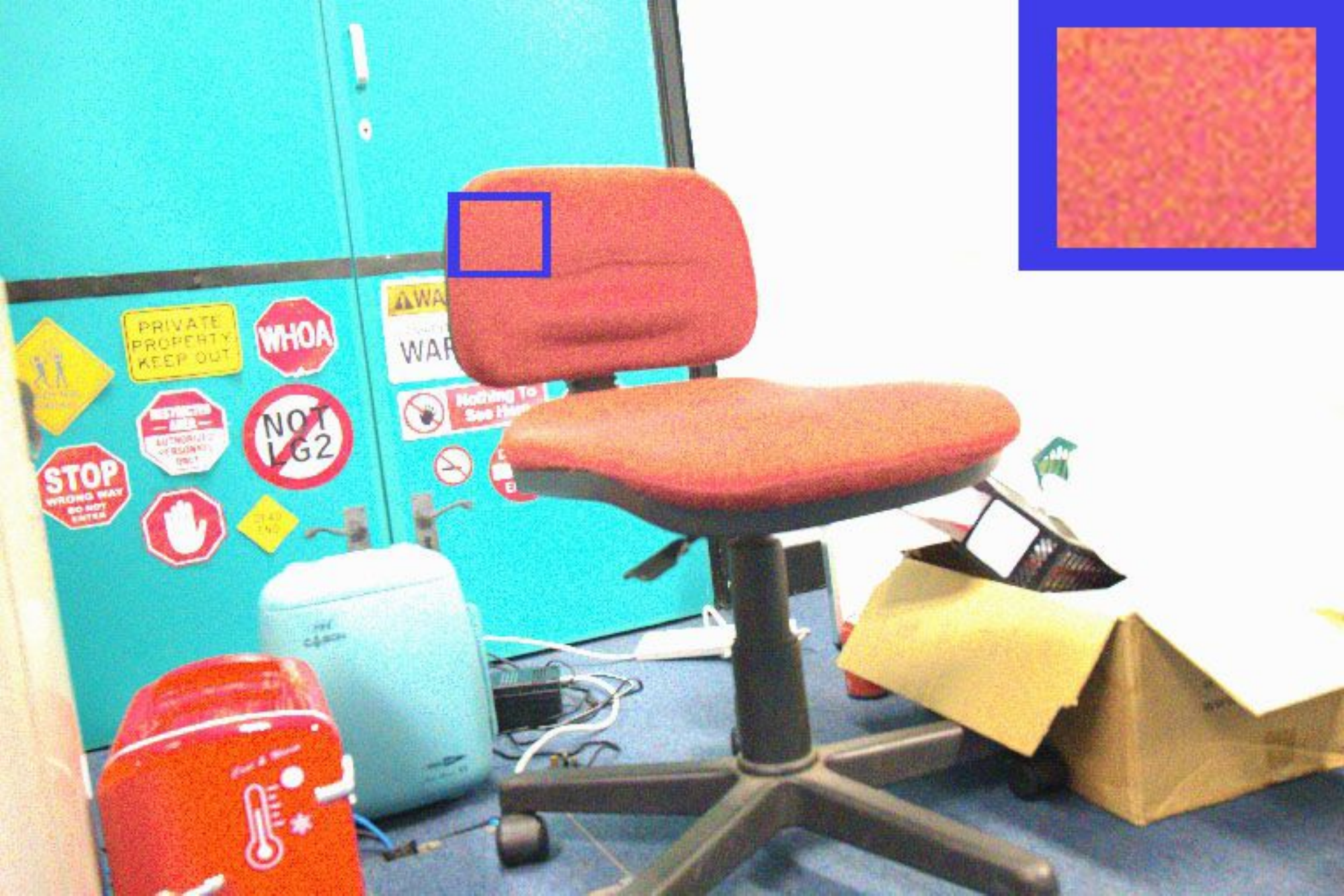} &
     \includegraphics[width=0.19\linewidth]{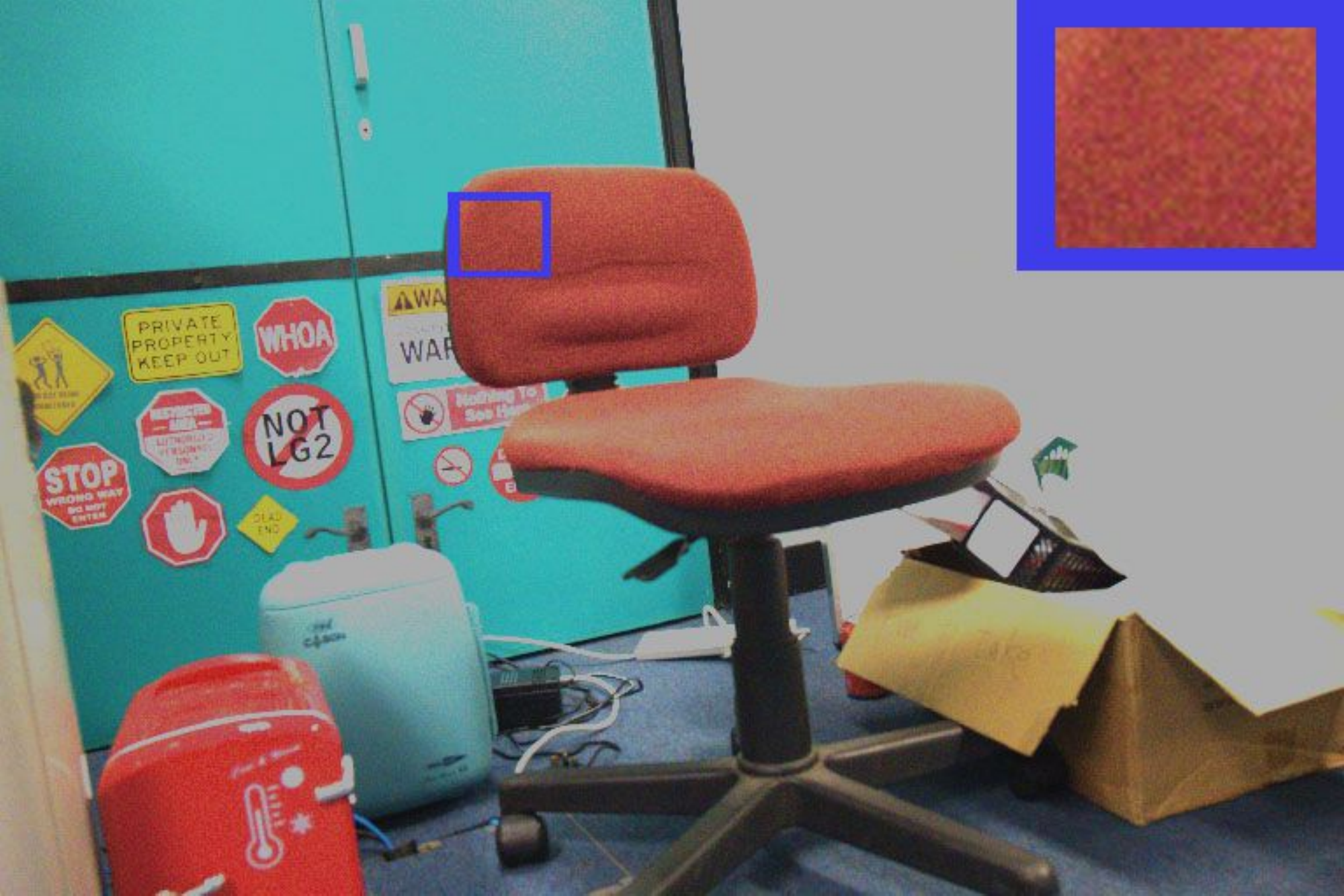} & \includegraphics[width=0.19\linewidth]{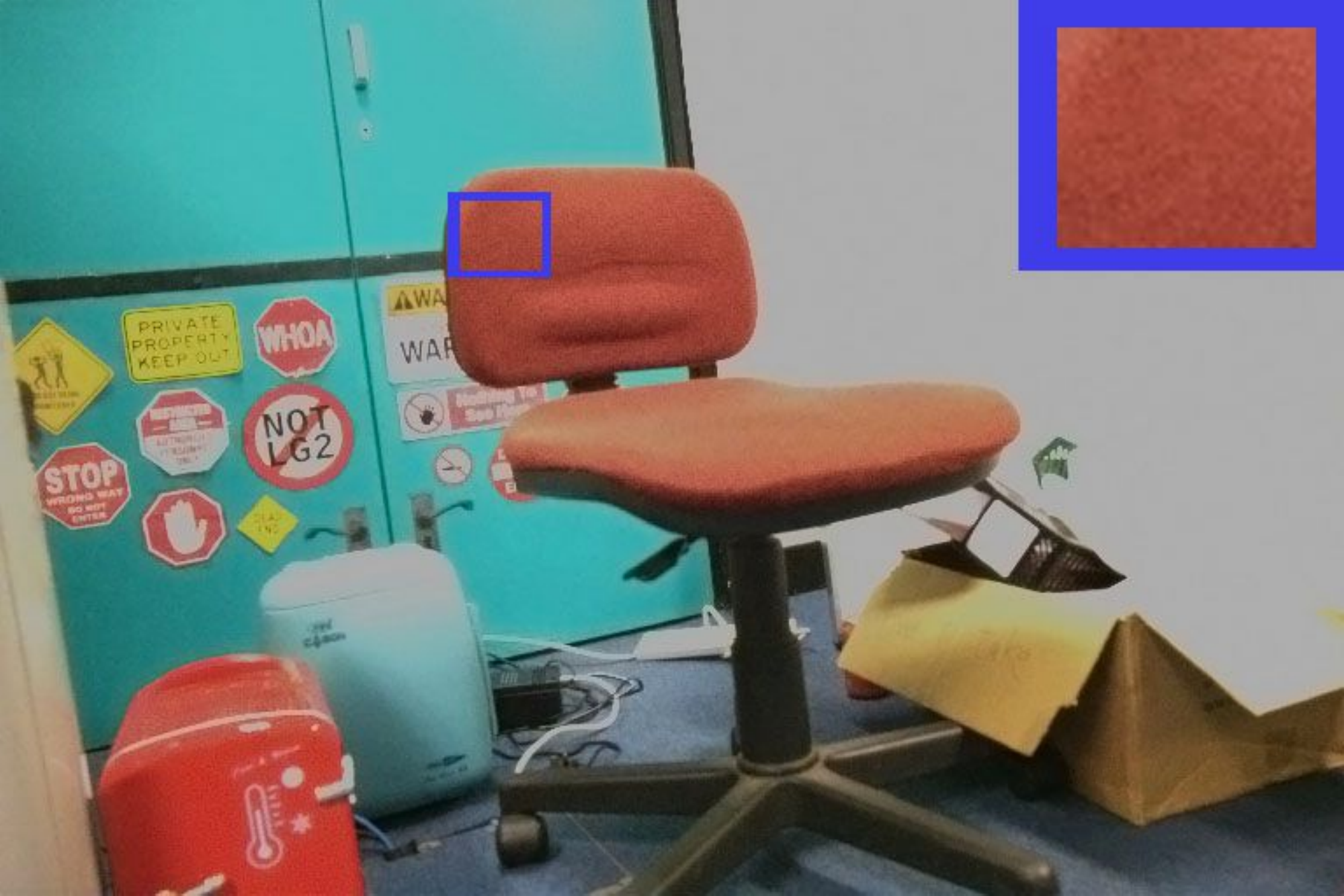}  & \includegraphics[width=0.19\linewidth]{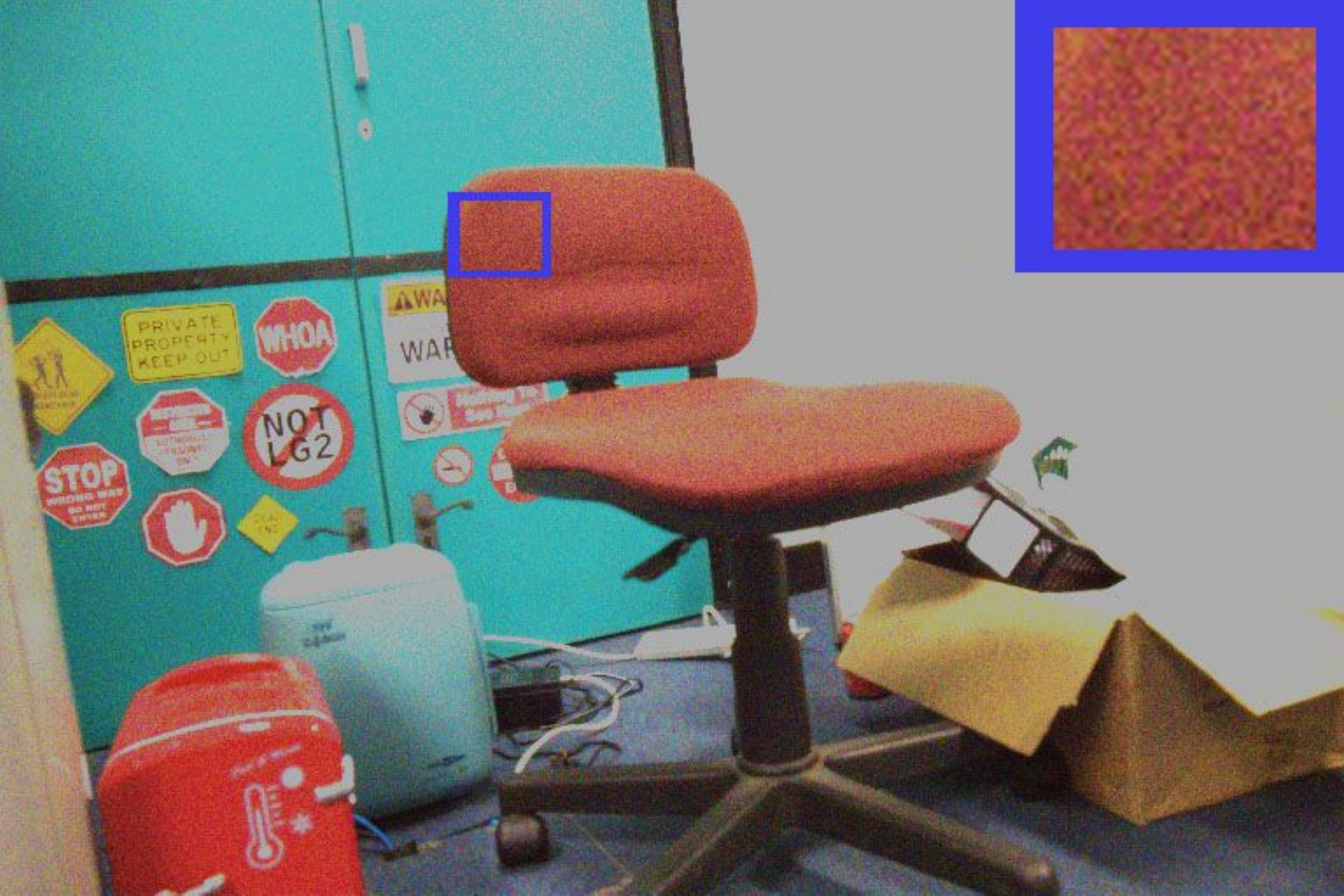} &    \includegraphics[width=0.19\linewidth]{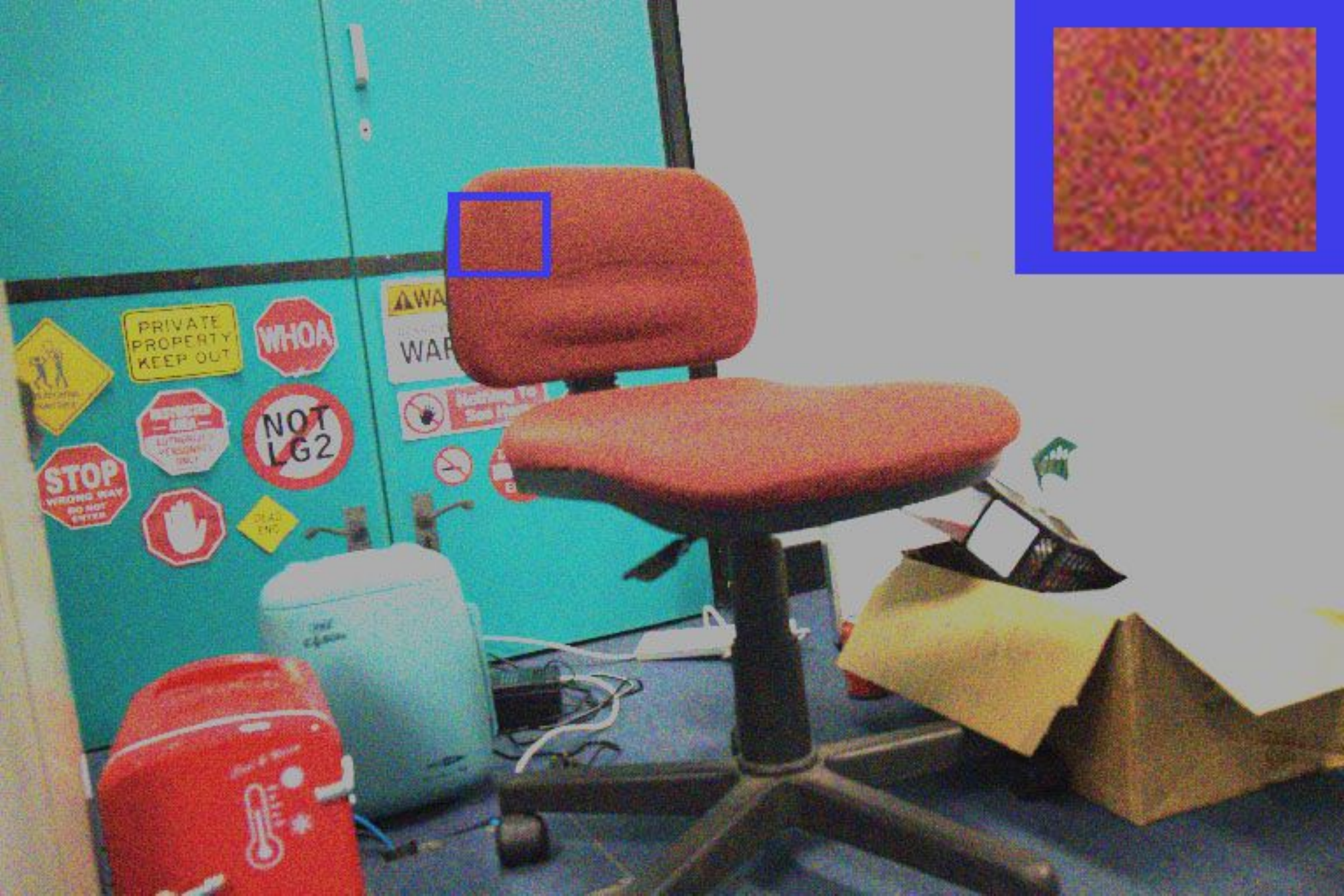}   \\

    \includegraphics[width=0.19\linewidth]{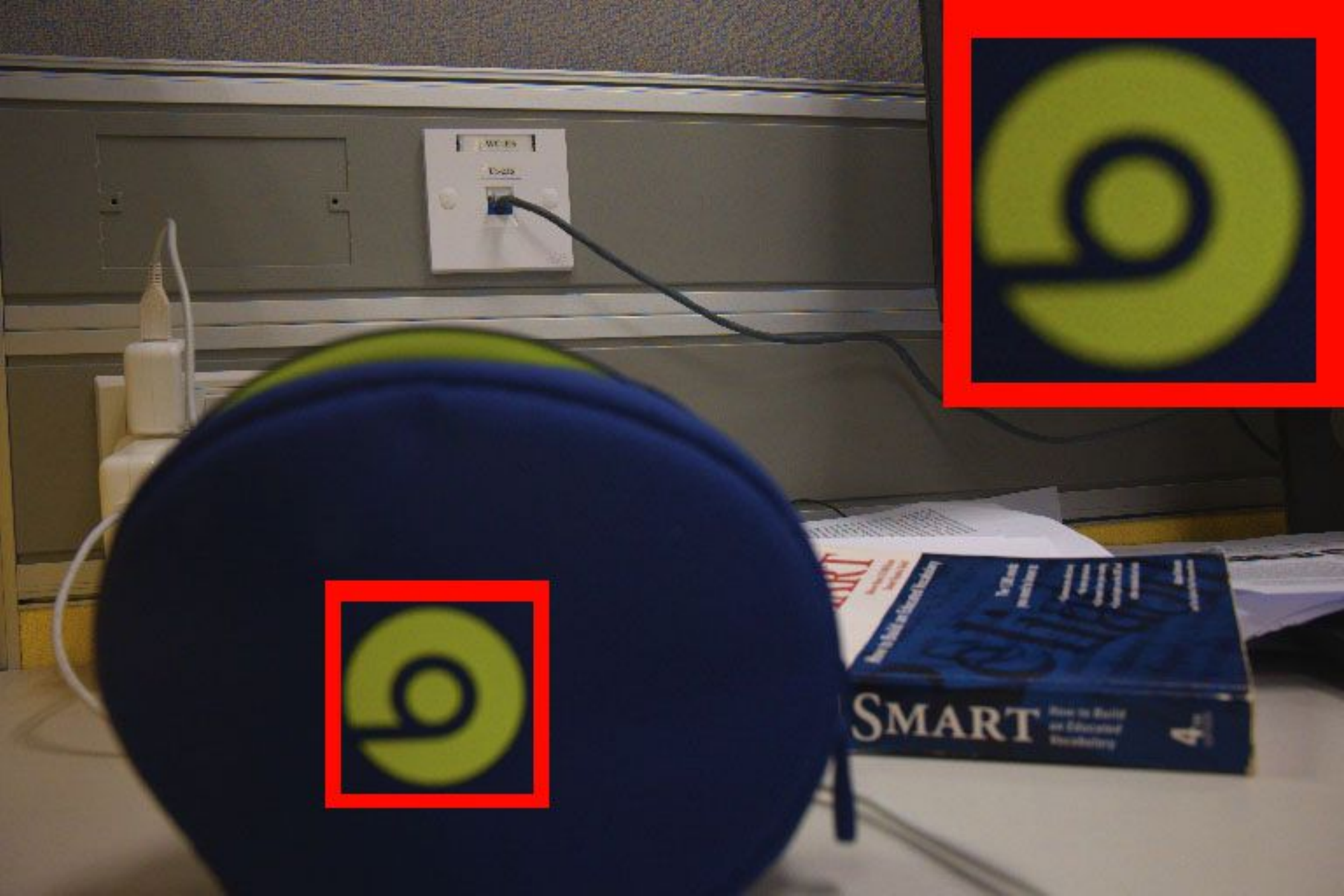} &
     \includegraphics[width=0.19\linewidth]{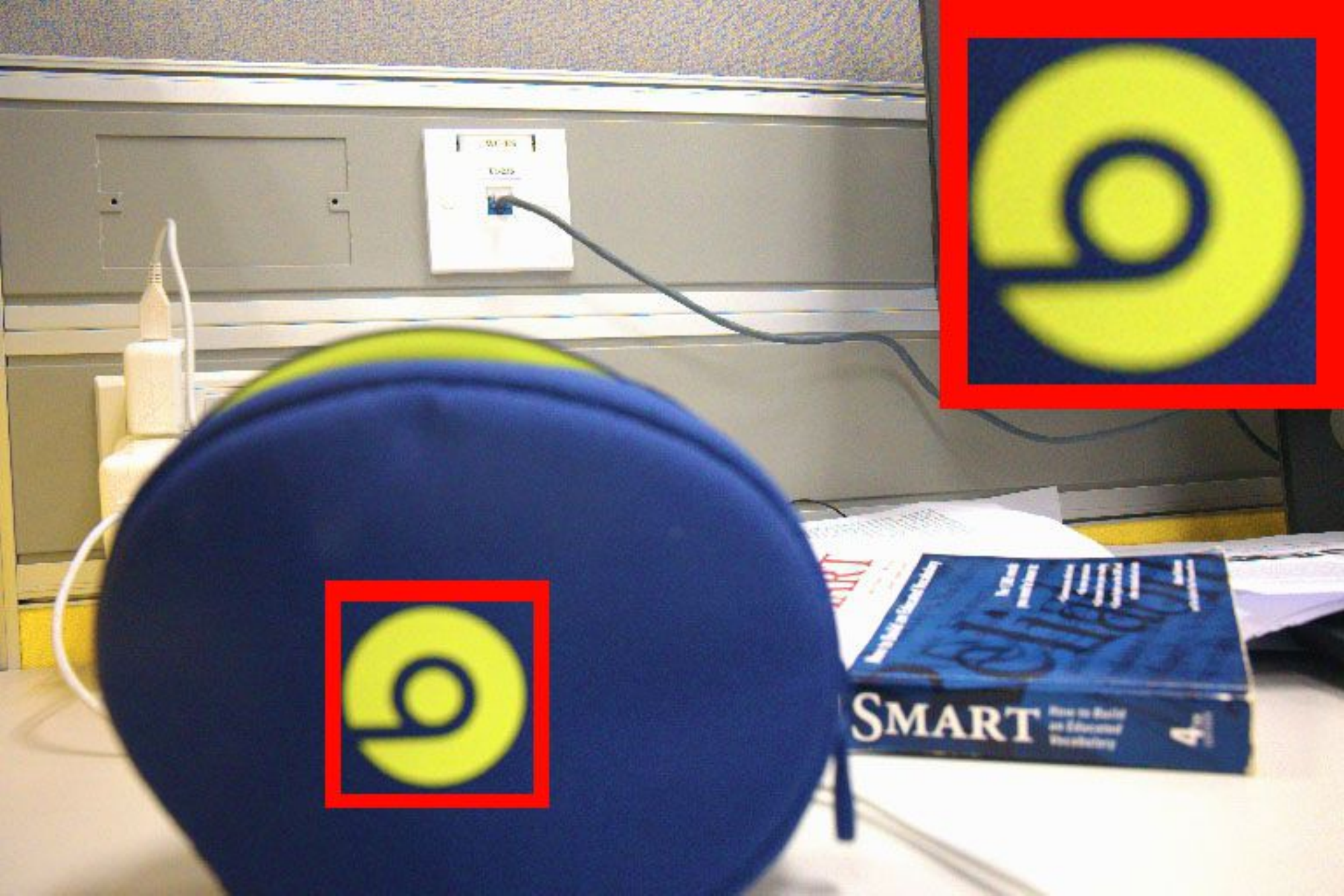}    &    \includegraphics[width=0.19\linewidth]{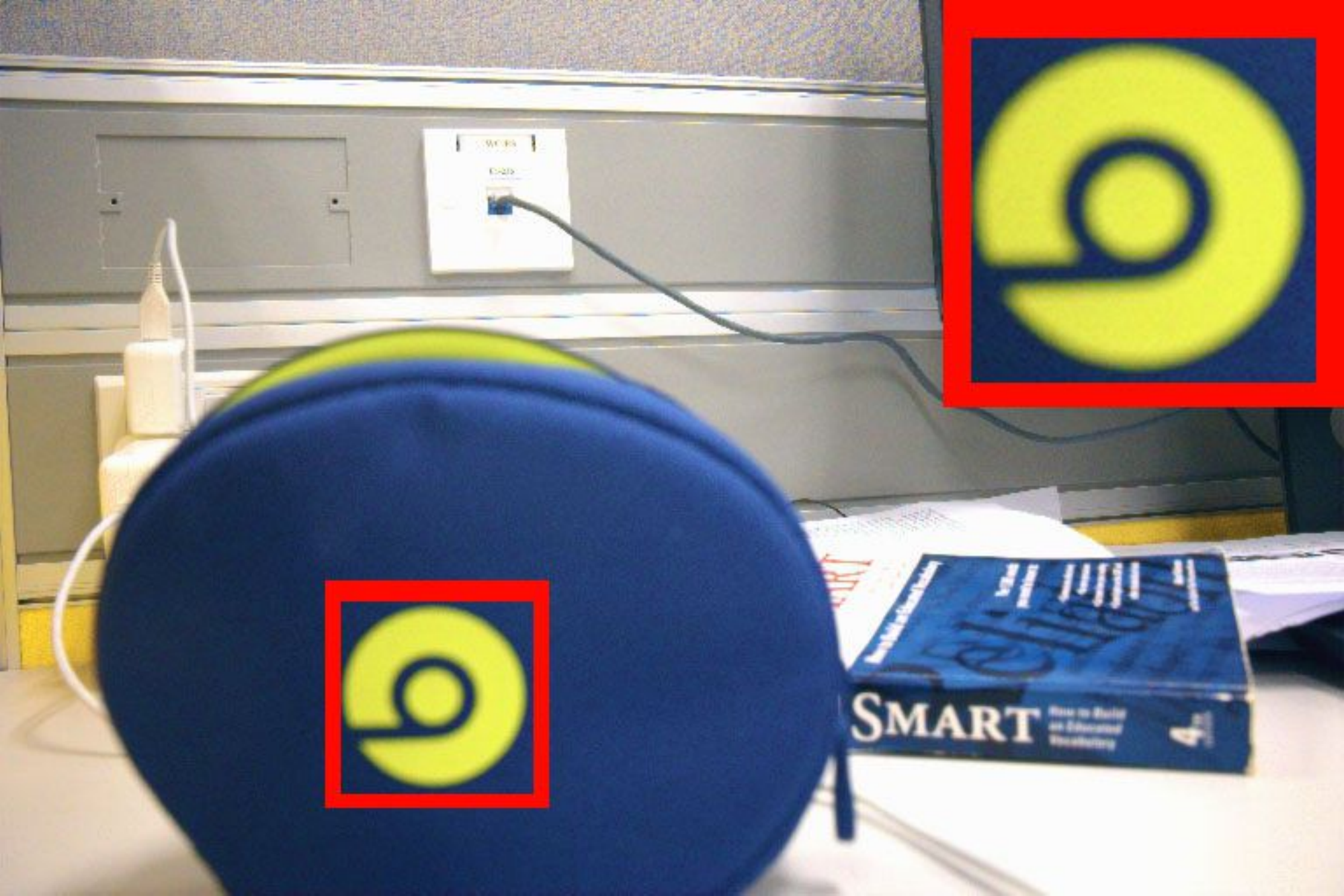}  & \includegraphics[width=0.19\linewidth]{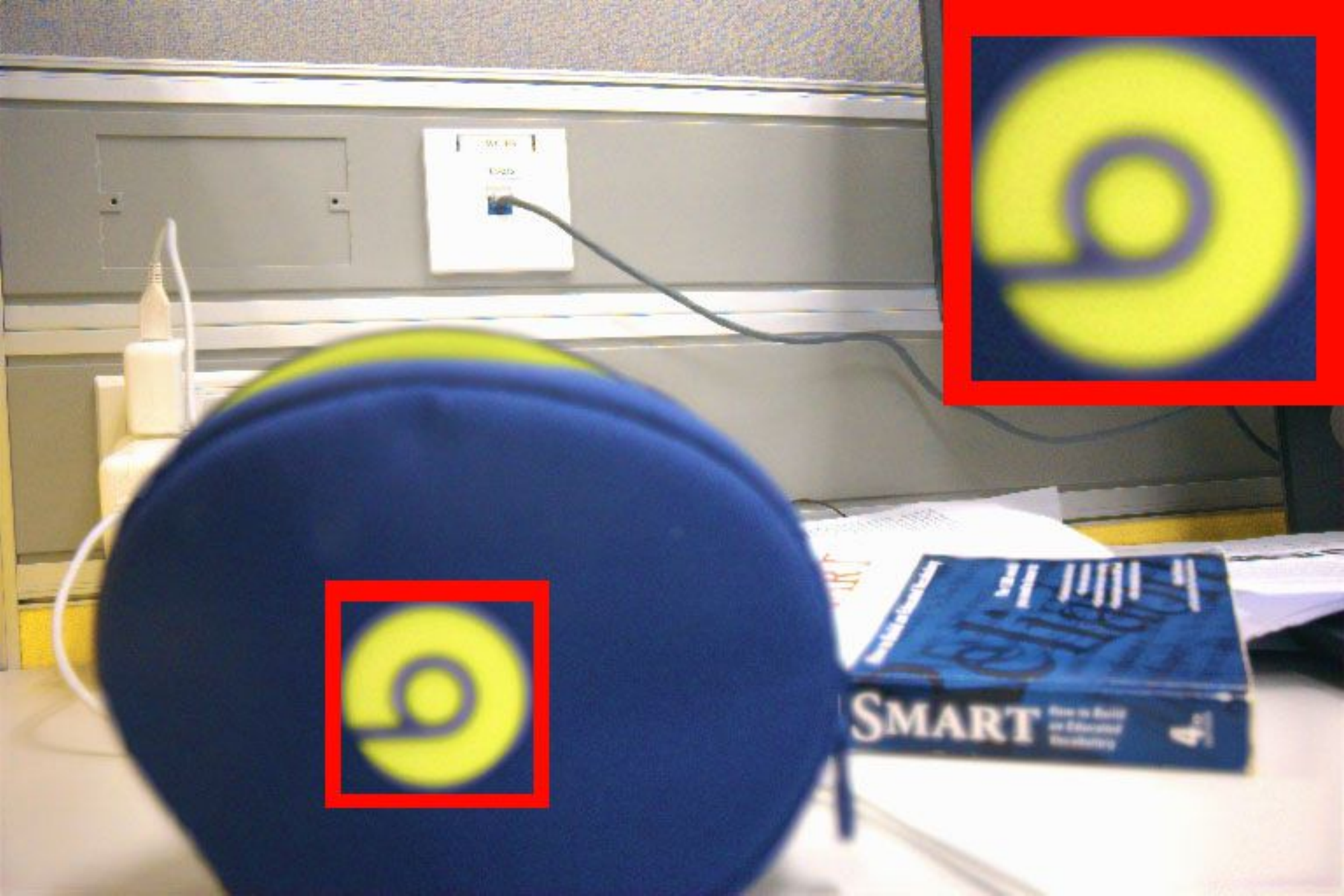} &    \includegraphics[width=0.19\linewidth]{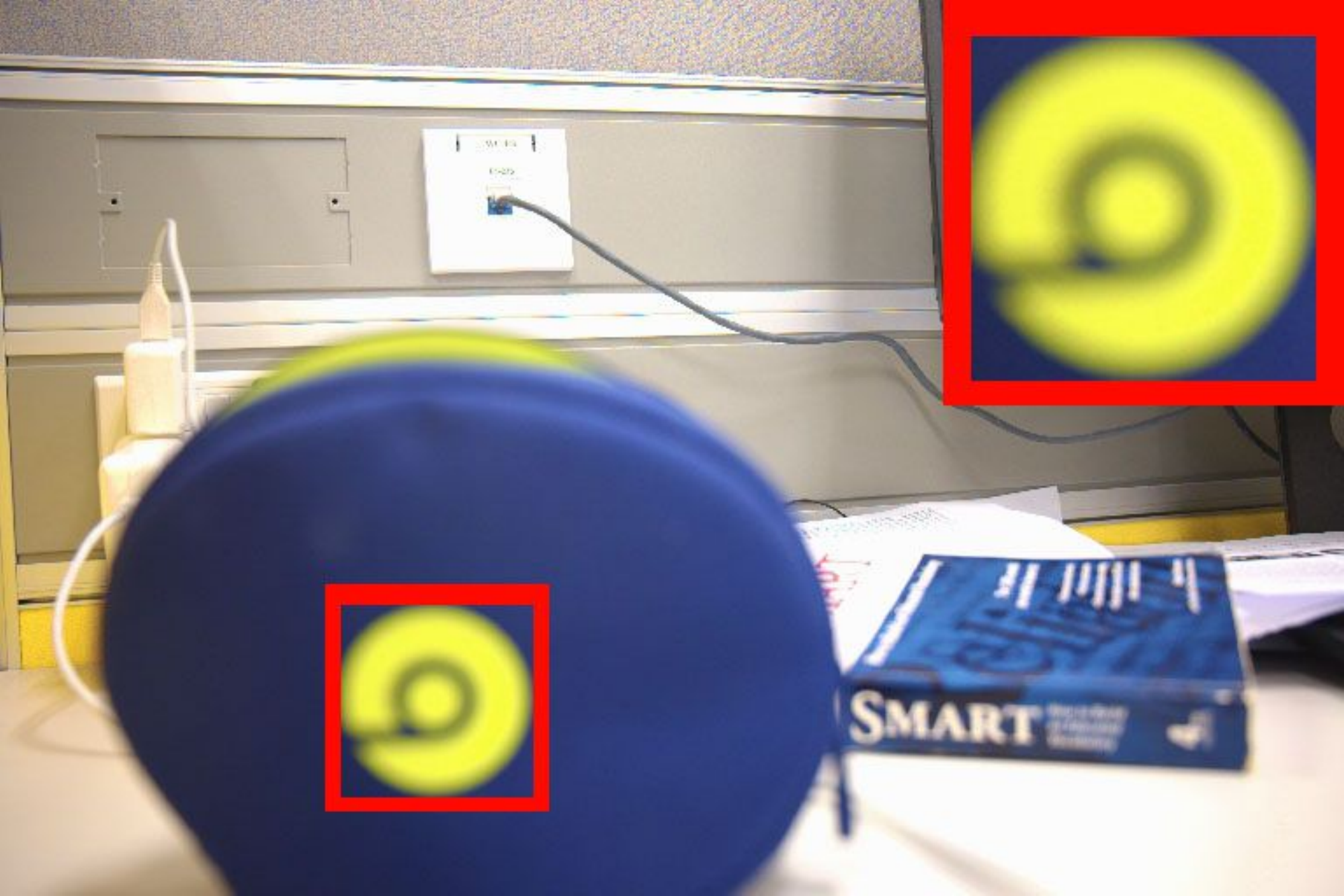}   \\
     
    \includegraphics[width=0.19\linewidth]{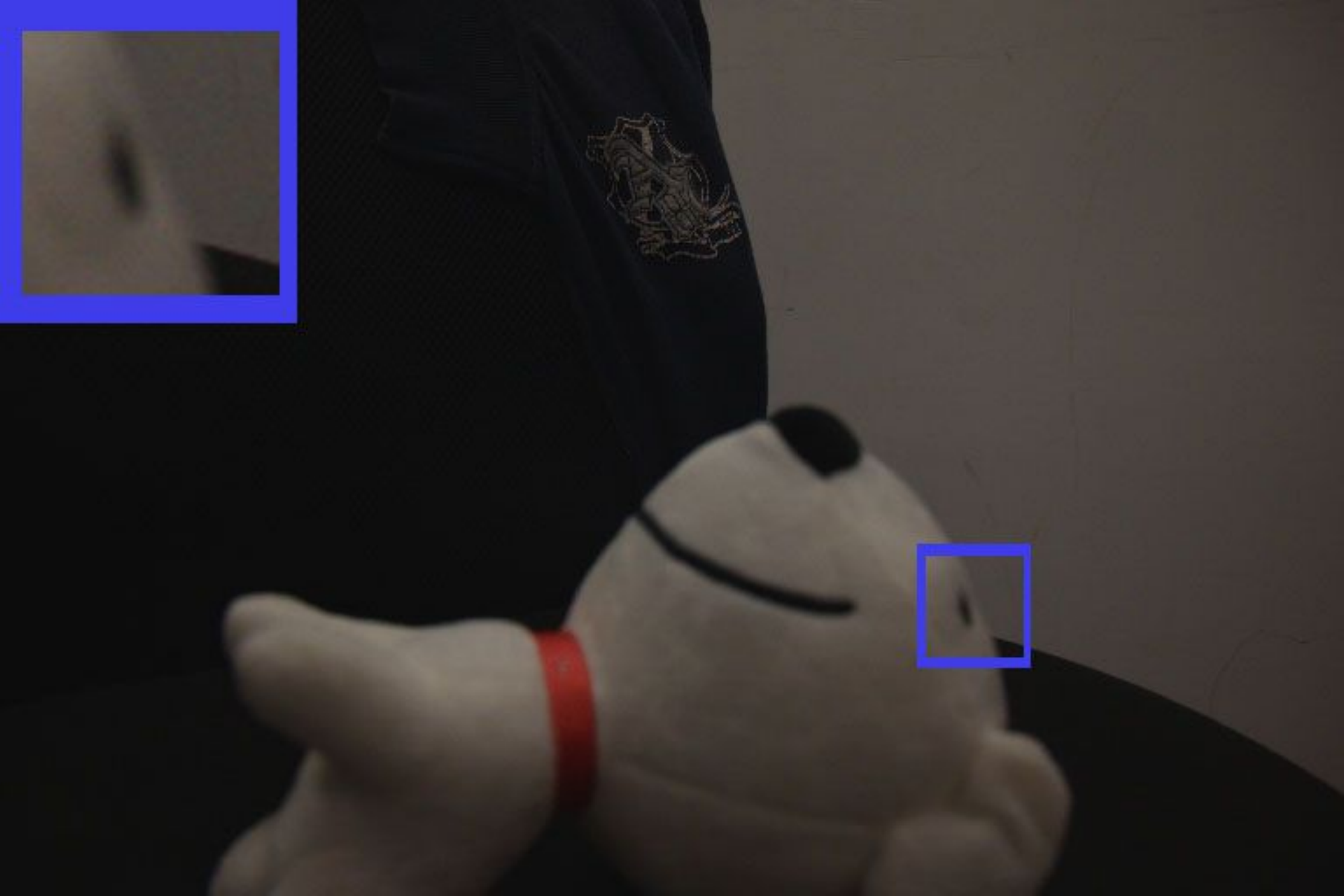} &
     \includegraphics[width=0.19\linewidth]{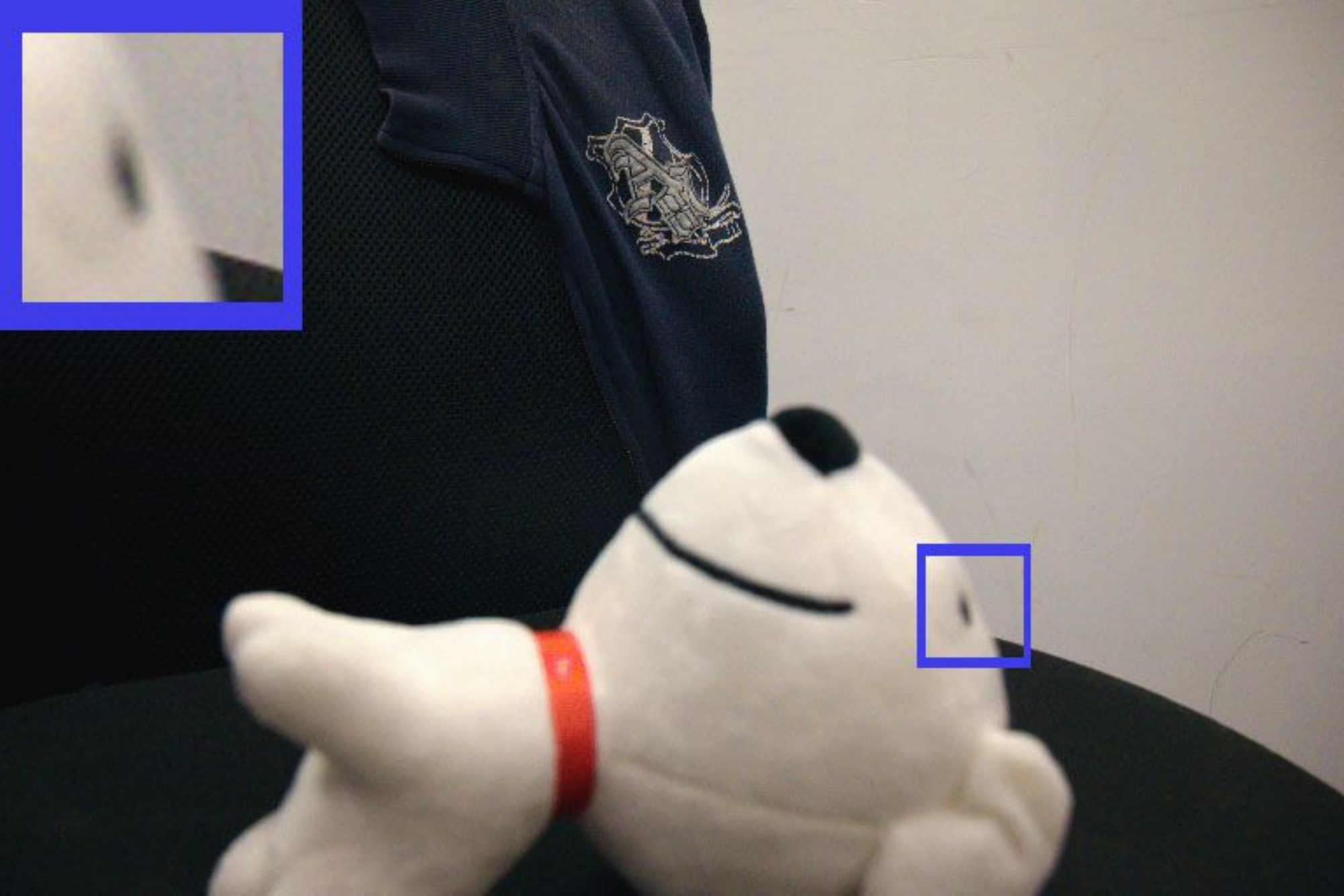}    &   \includegraphics[width=0.19\linewidth]{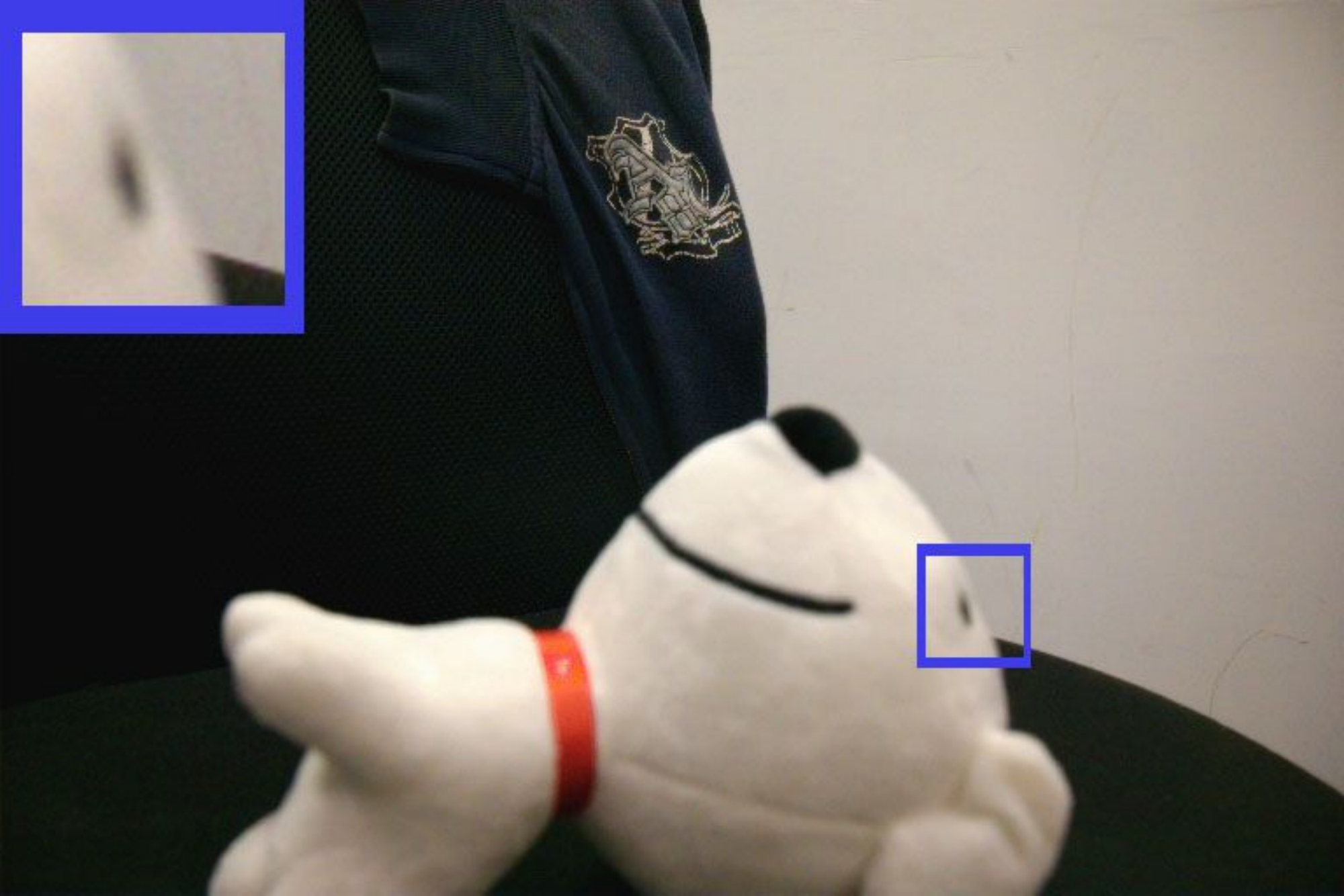}  & \includegraphics[width=0.19\linewidth]{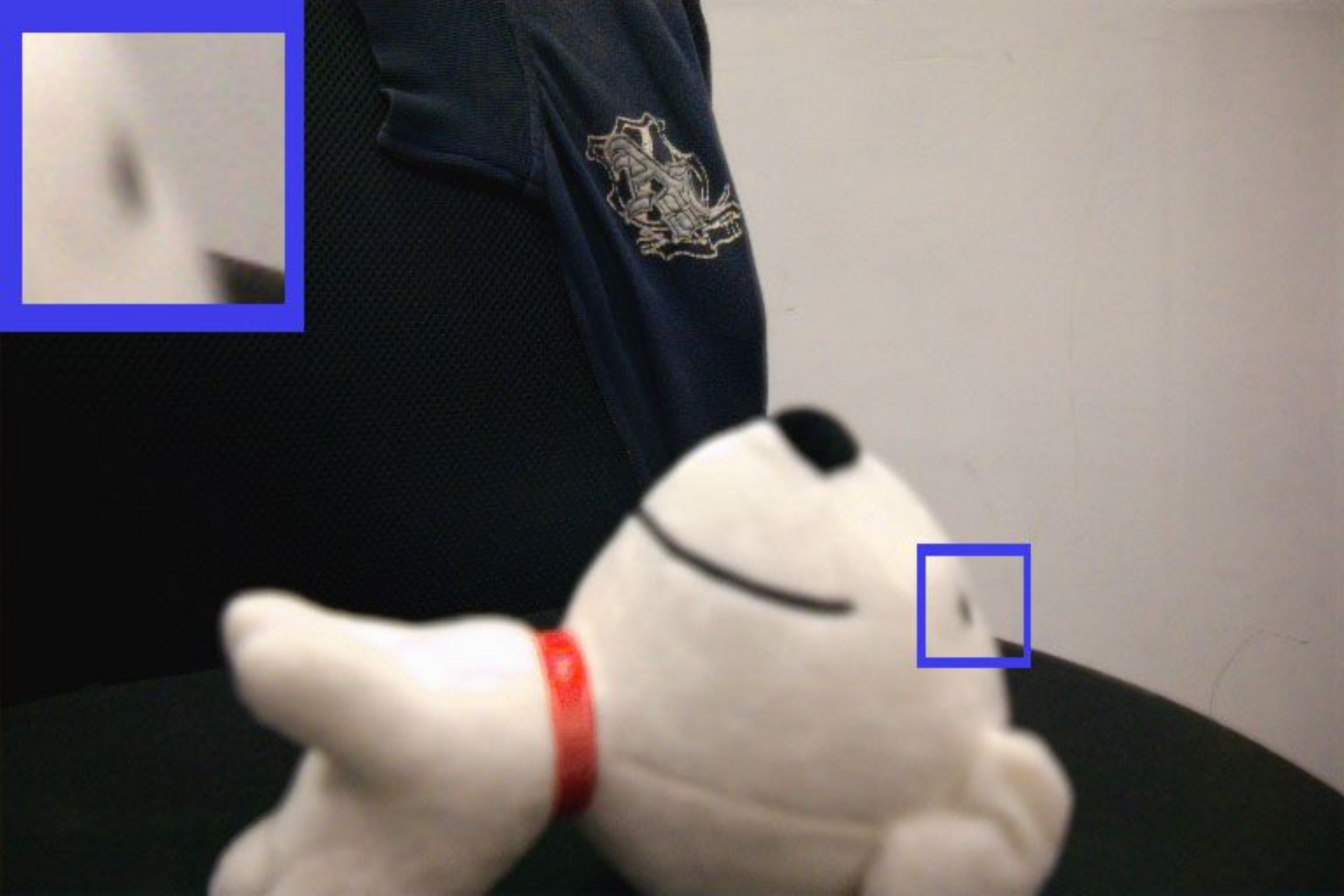} &    \includegraphics[width=0.19\linewidth]{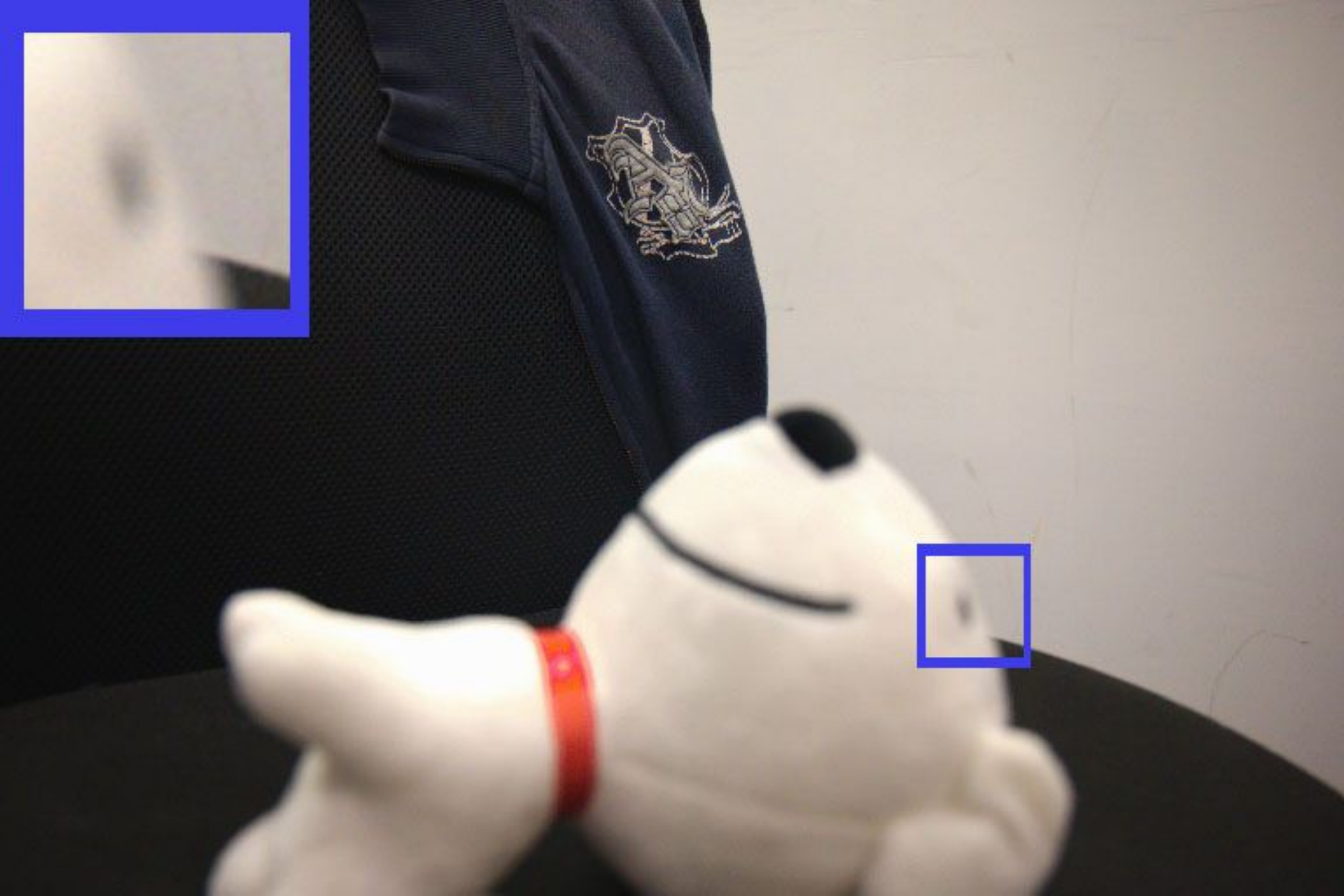}   \\
     
    \includegraphics[width=0.19\linewidth]{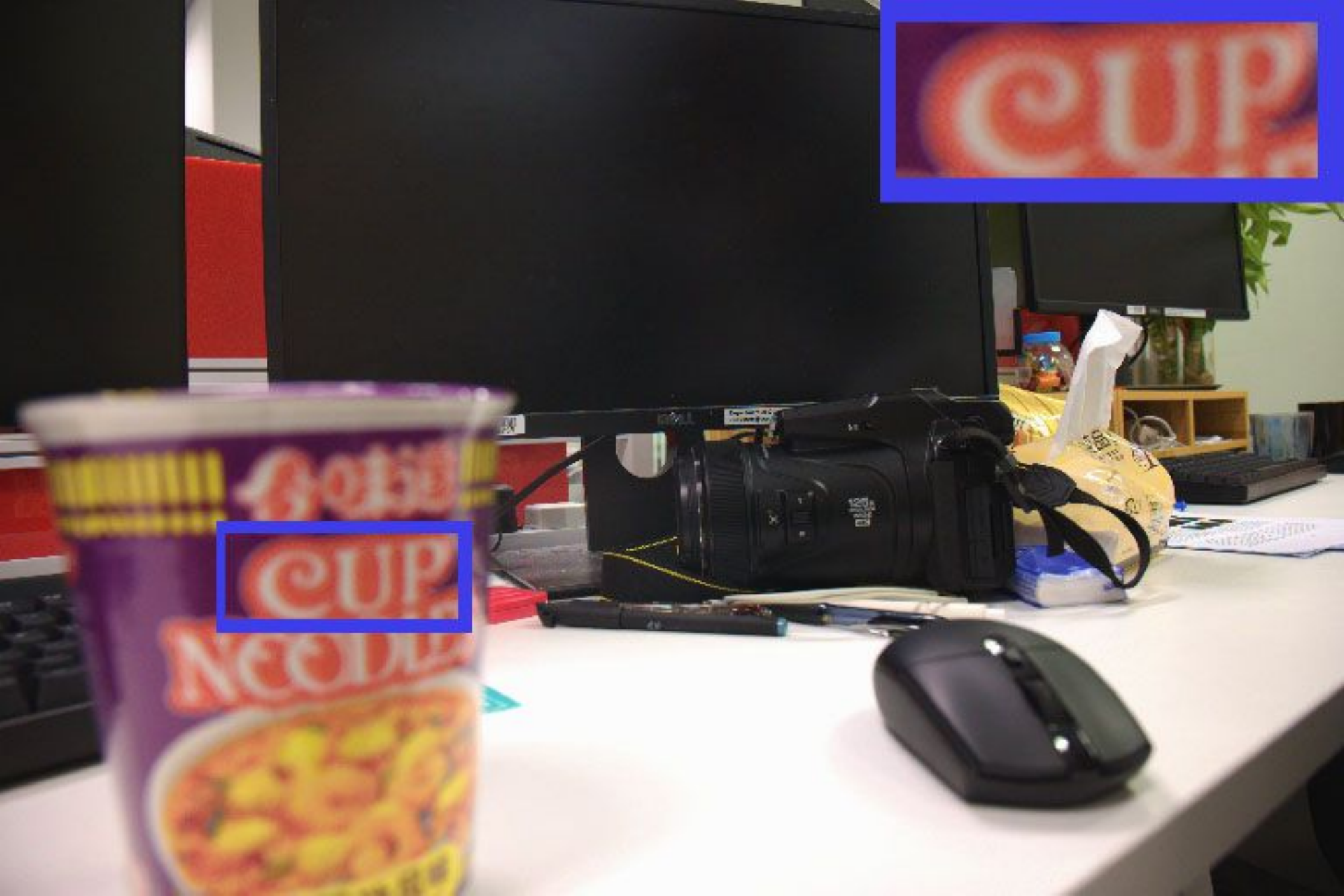} &
     \includegraphics[width=0.19\linewidth]{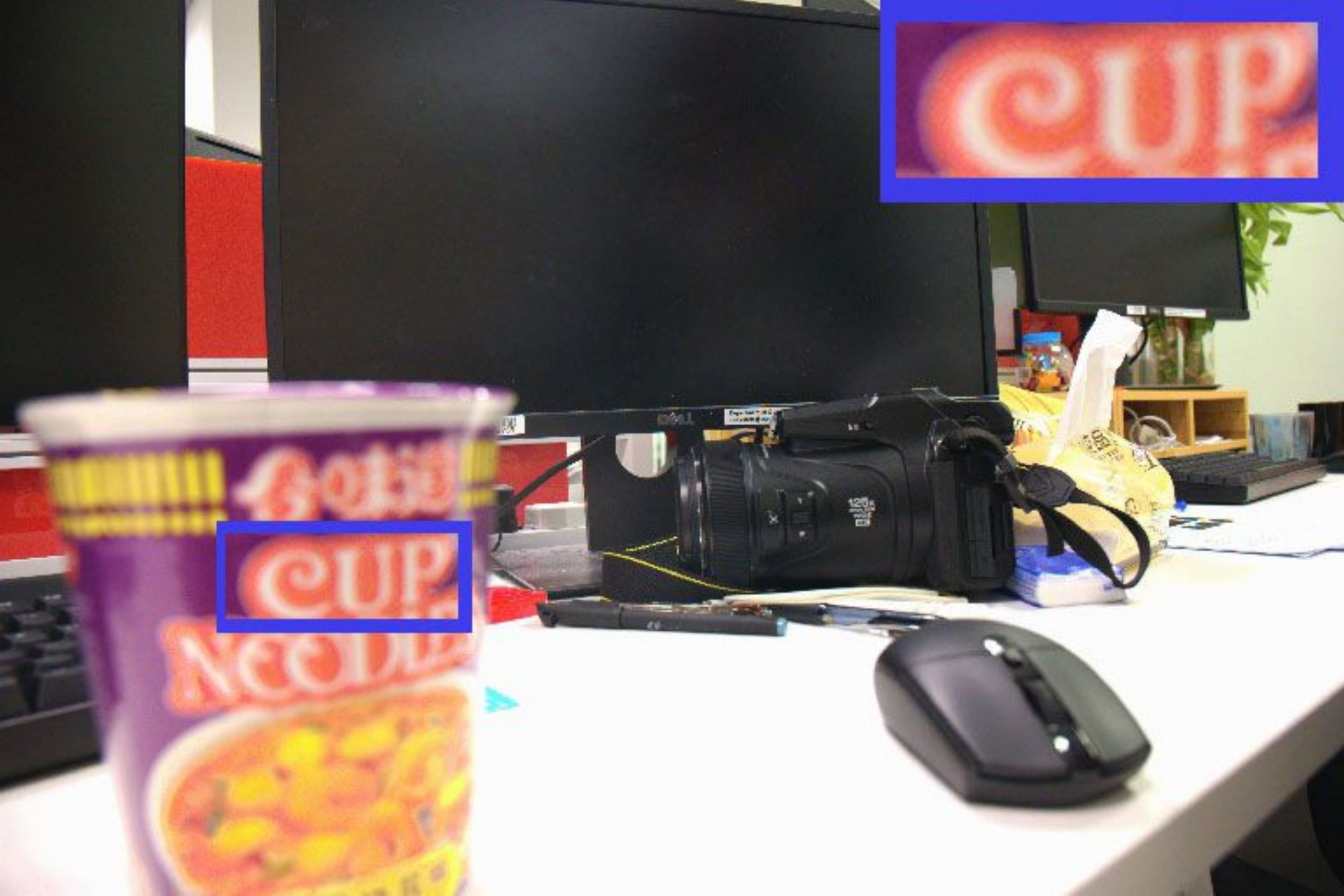}    &   \includegraphics[width=0.19\linewidth]{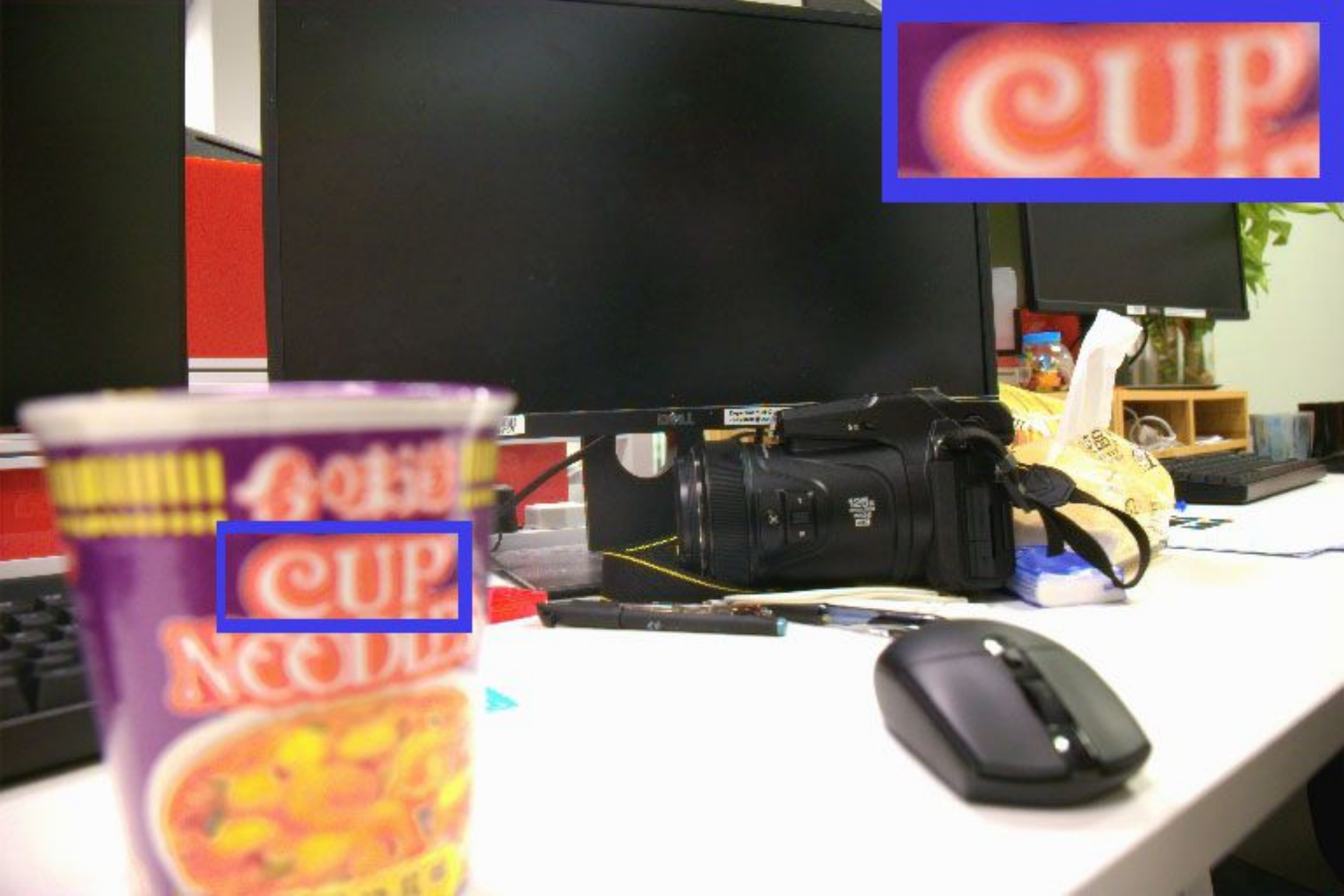}  & \includegraphics[width=0.19\linewidth]{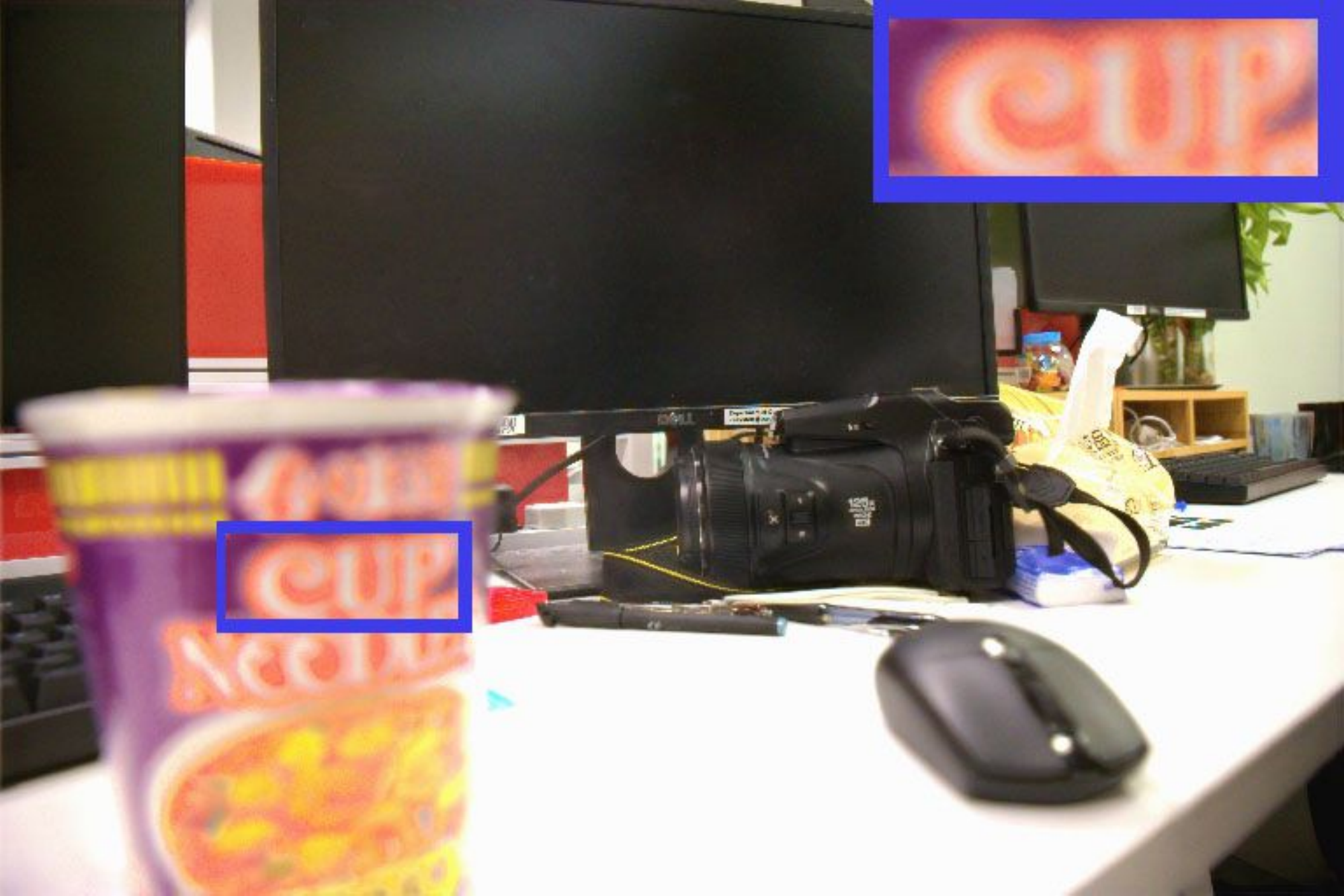} &    \includegraphics[width=0.19\linewidth]{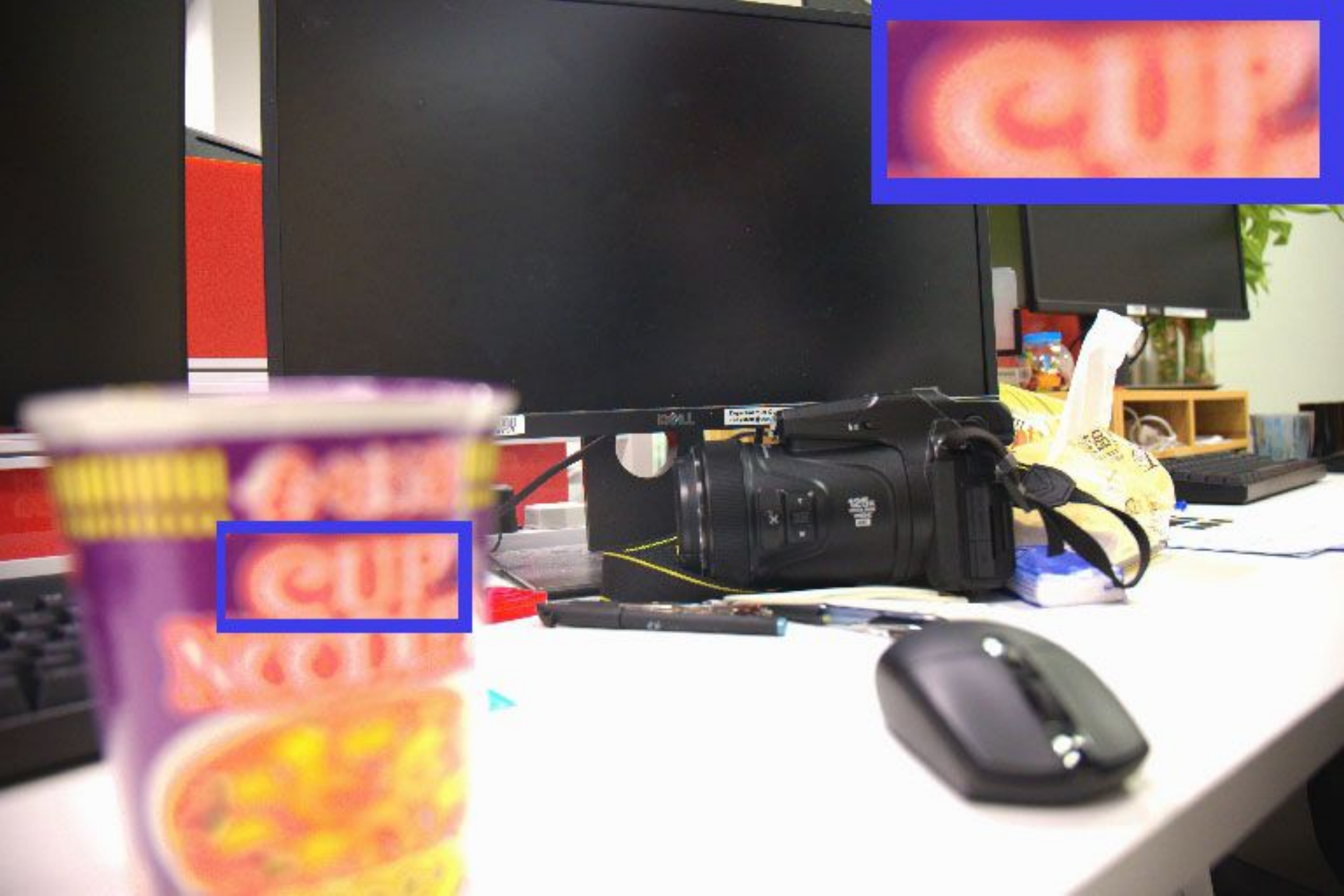}  \\

    \end{tabular}
    \caption{Images generated by each module of our simulator. Zoom-in to view details.}
    \label{ fig:pipeline}

\end{figure*}

\begin{table*}[t!]
\centering 
\begin{tabular}{@{}l p{0.5cm} c   p{0.3cm} c  p{0.3cm} c p{0.3cm} c p{0.3cm} c@{}} 
\toprule 
&& PC~\cite{chen2017fast} && DL~\cite{fan2018decouple} && Ours (EXP) && Ours (NS) &&  Ours (Full model)\\ 
\midrule 
\bf{PSNR (Nikon Z6)} && 18.78 && 34.34 && 35.31 && 35.36 && \bf{36.10}\\
\bf{SSIM (Nikon Z6)} && 0.542 && 0.879 && 0.903 && 0.911  && \bf{0.923}\\
\bf{PSNR (Canon 70D)} && 18.96 && 24.86 && 33.53 && 33.49  && \bf{34.28}\\
\bf{SSIM (Canon 70D)} && 0.413 && 0.429 && 0.831 && 0.837  && \bf{0.846}\\
\bottomrule \\
\end{tabular}
\caption{PSNR and SSIM of different methods and different stages of our model: PC~\cite{chen2017fast}, DL~\cite{fan2018decouple}, EXP (Ours after exposure module), NS (Ours after noise module), and Full (Our full model).}
\label{tab:ablation} 
\end{table*}
 
\section{Experiments}
\subsection{Dataset}
We collect a new dataset for simulating images with different settings. The dataset contains around 10,000 raw images, which are divided into 450 image sequences in different scenes.  In each sequence, we capture 20 to 25 images with different camera settings. ISO,  exposure time, and aperture f-number are sampled respectively from the discrete levels in the range $[100, 16000]$, $[1/8000s, 4s]$, and $[4.0, 22.0]$.

The collected dataset contains both indoor and outdoor images taken under different lighting conditions. The illuminance of these scenes varies from 1 lux to 20000 lux. We adapt the camera settings for each scene to capture more information. In brighter scenes, we use lower ISO, shorter exposure time, and smaller aperture size on average for capturing image sequences. In consideration of the diversity of the dataset, we use two different cameras, DSLR camera Canon 70D and mirrorless camera Nikon Z6, for data acquisition. The captured scene sequence has to be static. We mount cameras on sturdy tripods and utilize corresponding remote control mobile applications for photo shooting to avoid unintended camera movement. We manually filter out all images containing a moving object. 

\begin{figure*}
    \centering 
    \begin{tabular}{@{}c@{\hspace{0.5mm}}c@{\hspace{0.5mm}}c@{\hspace{0.5mm}}c@{\hspace{0.5mm}}c@{}}
    Input & PC & DL & Ours & Ground Truth \\
     \includegraphics[width=0.19\linewidth]{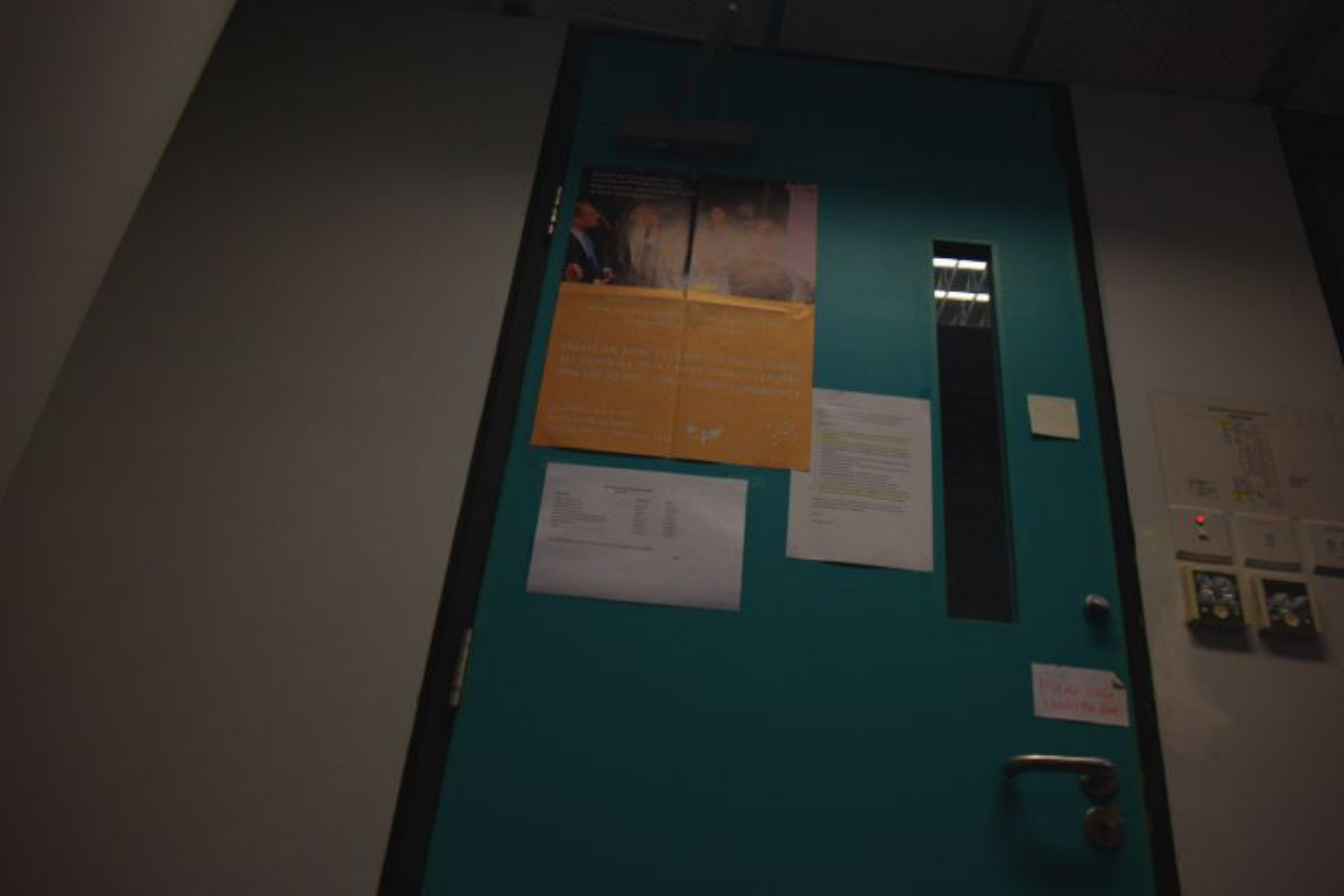} &
     \includegraphics[width=0.19\linewidth]{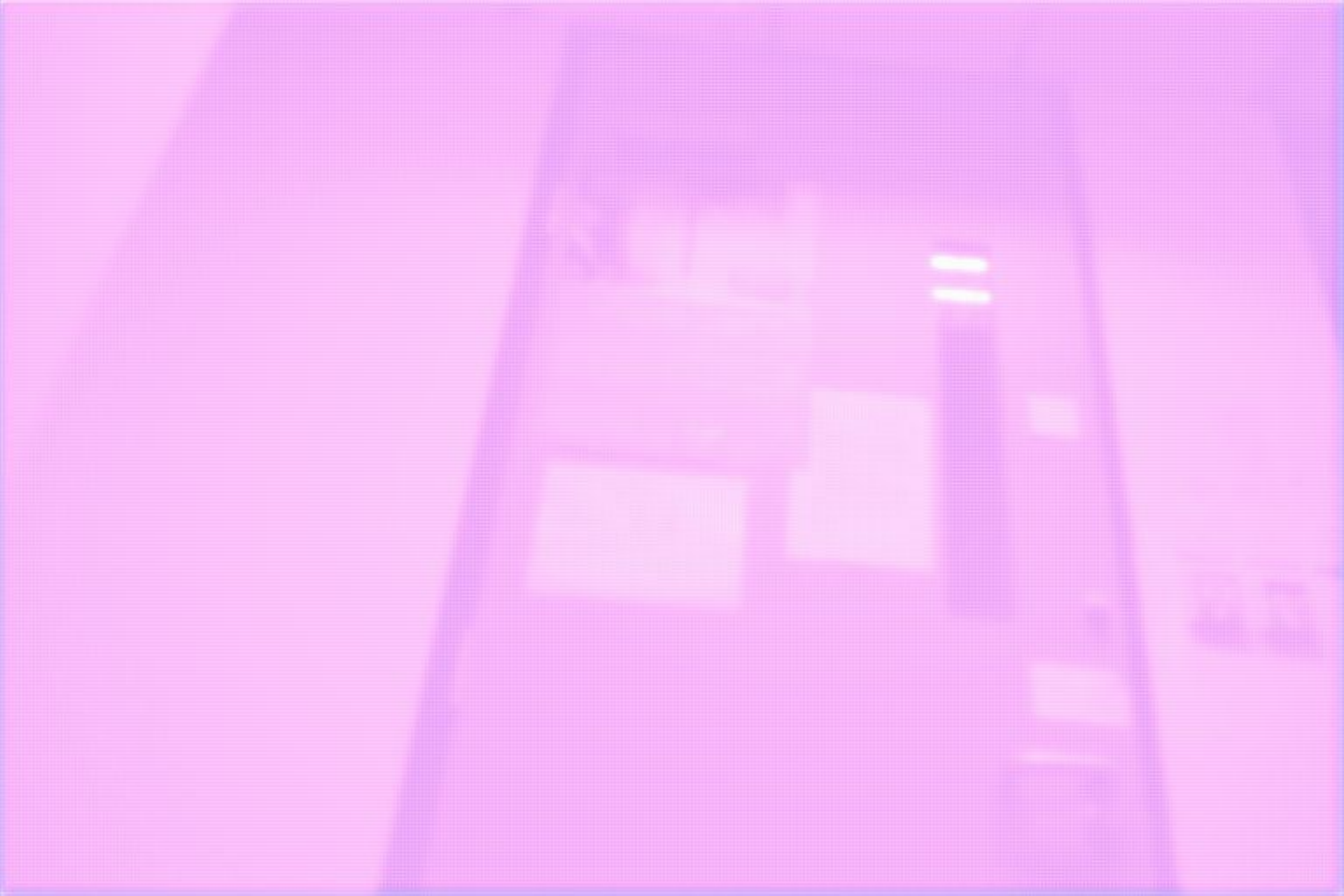} &    \includegraphics[width=0.19\linewidth]{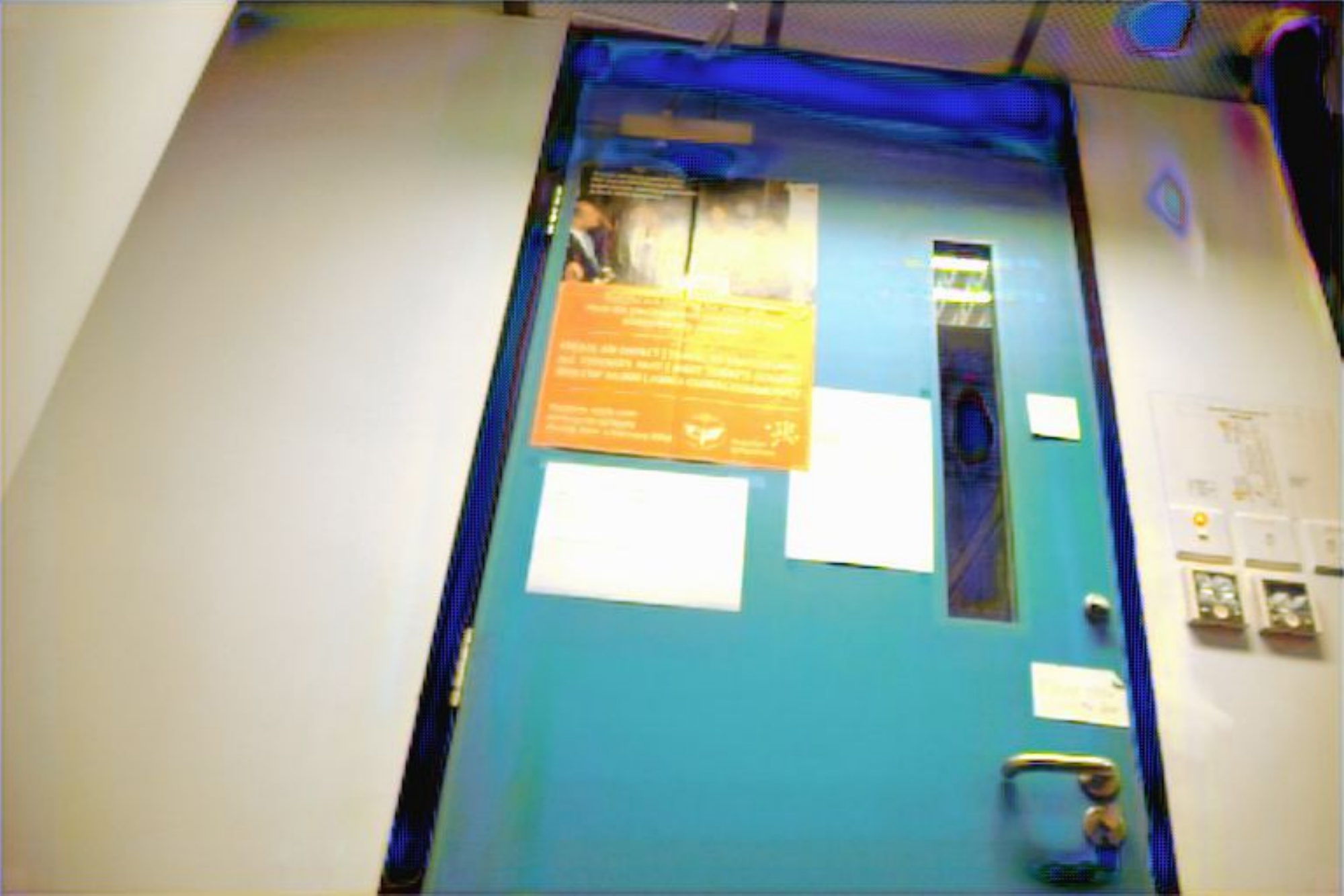}  & \includegraphics[width=0.19\linewidth]{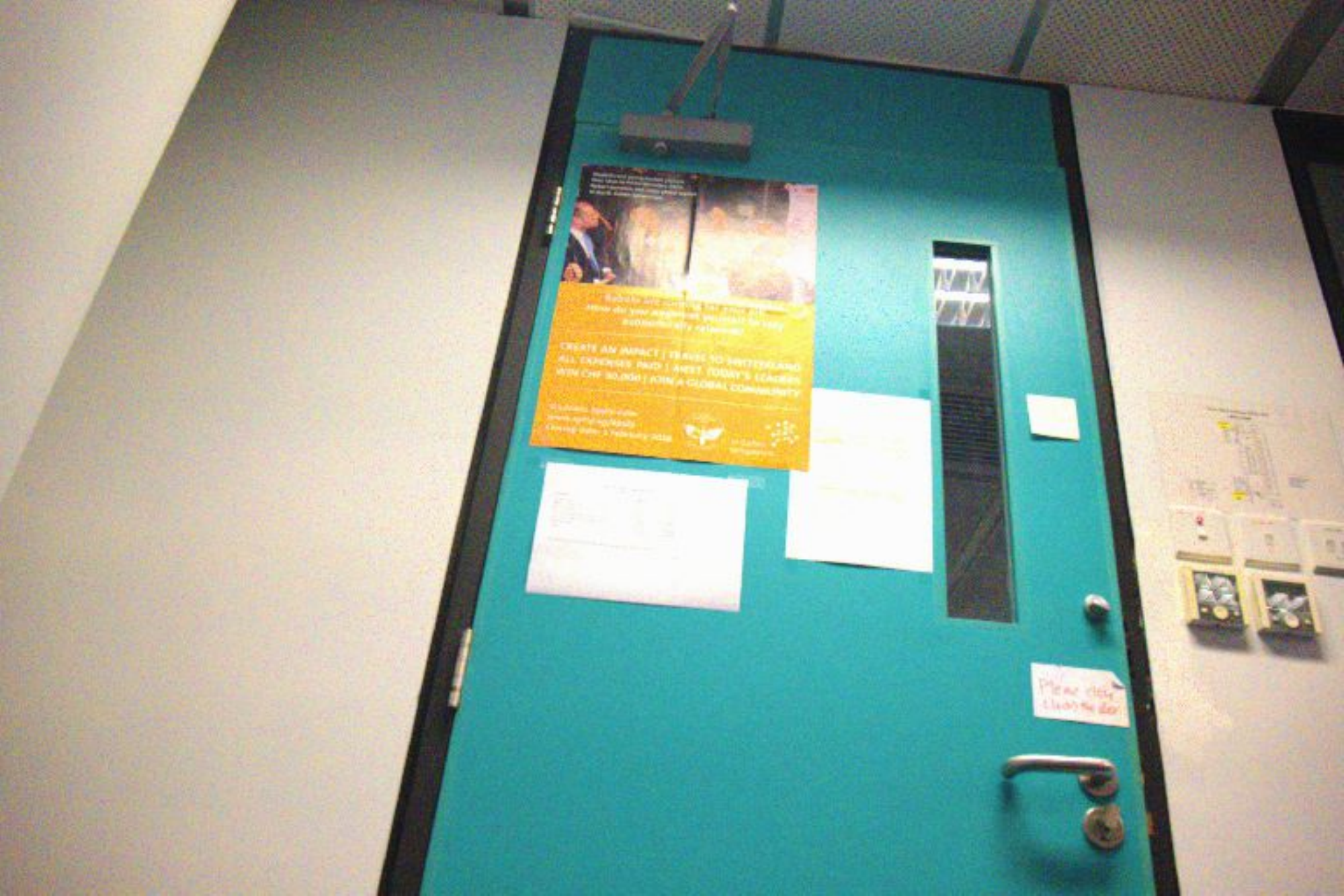} &    \includegraphics[width=0.19\linewidth]{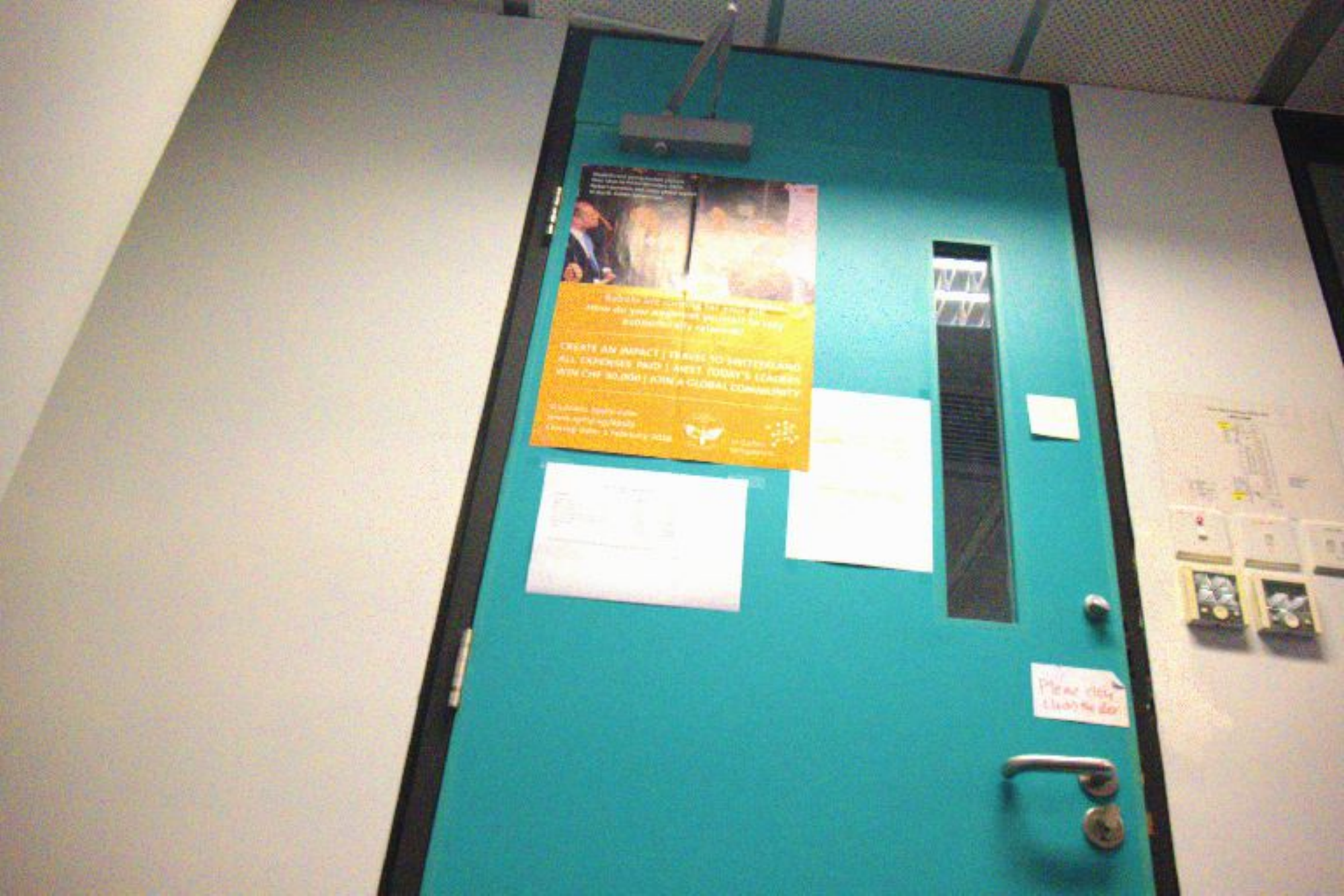}     \\

    \includegraphics[width=0.19\linewidth]{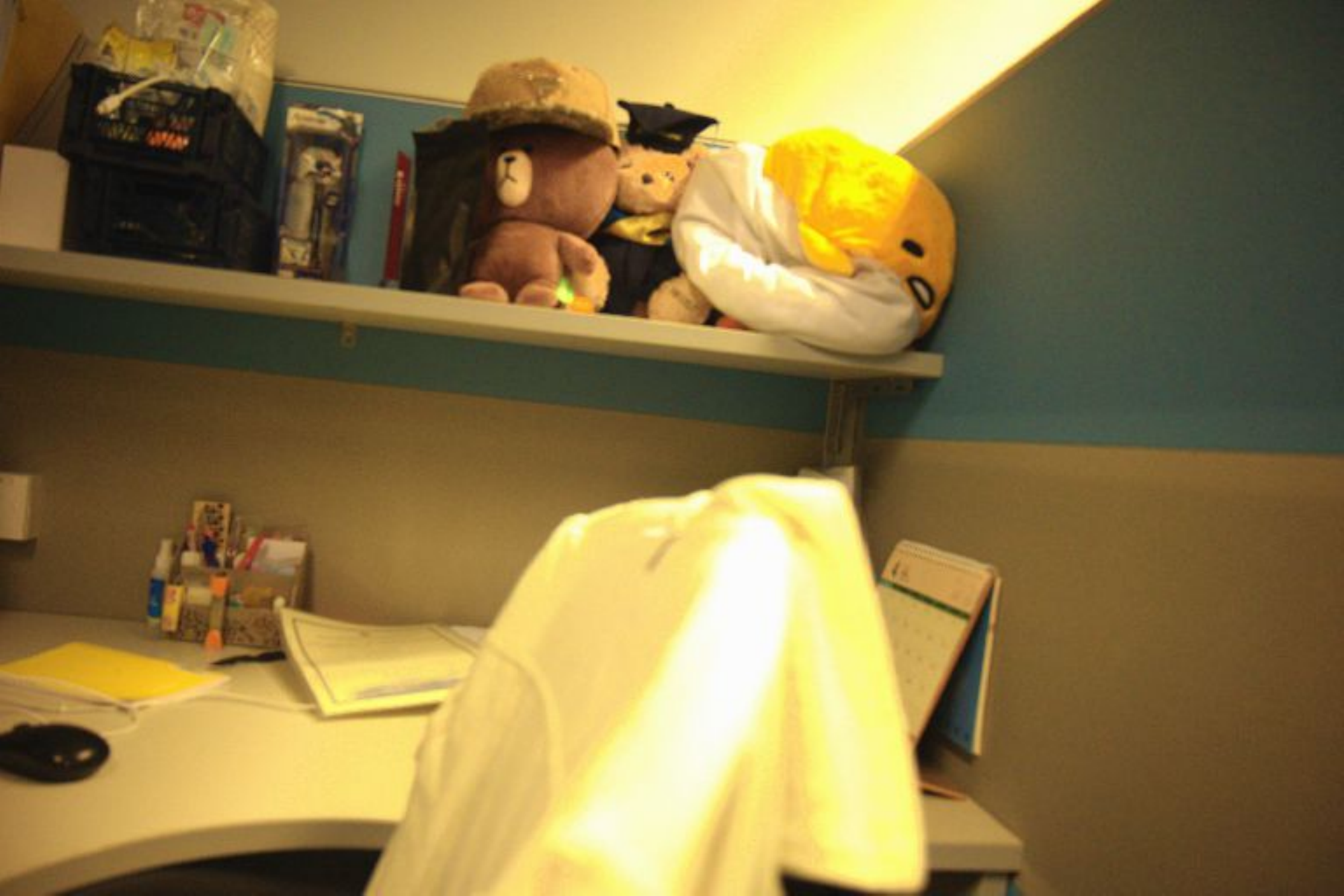} &
     \includegraphics[width=0.19\linewidth]{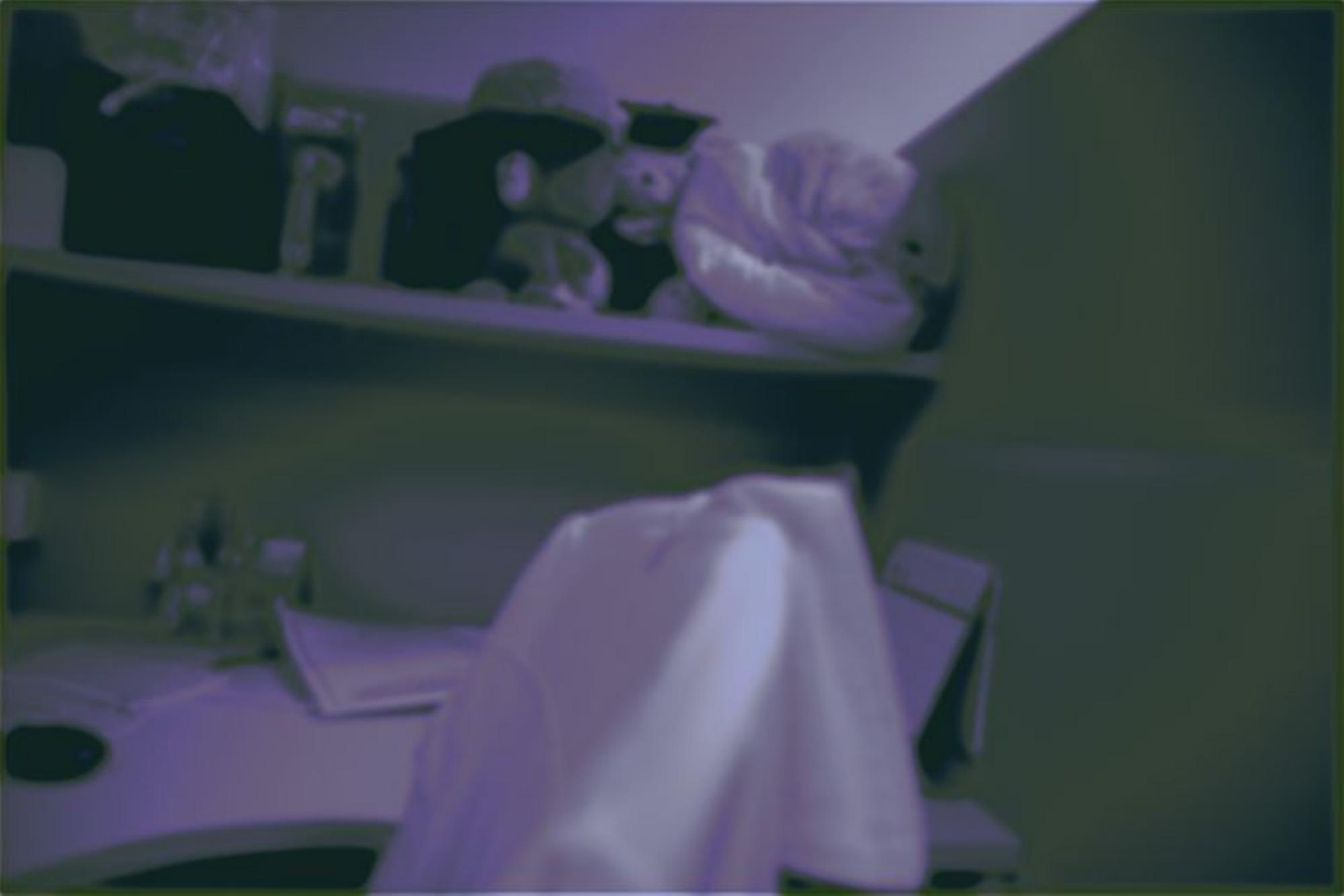} &    \includegraphics[width=0.19\linewidth]{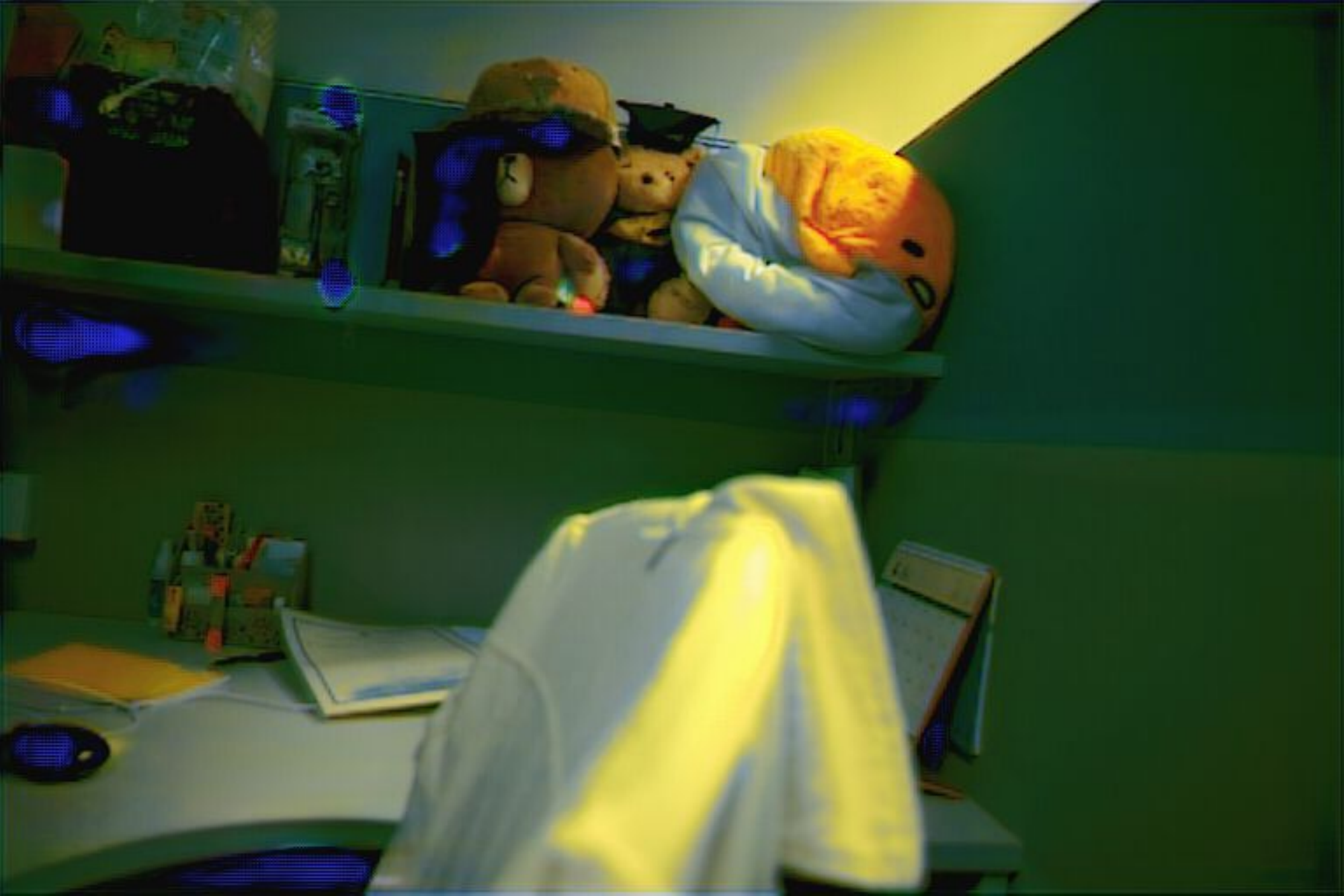}  & \includegraphics[width=0.19\linewidth]{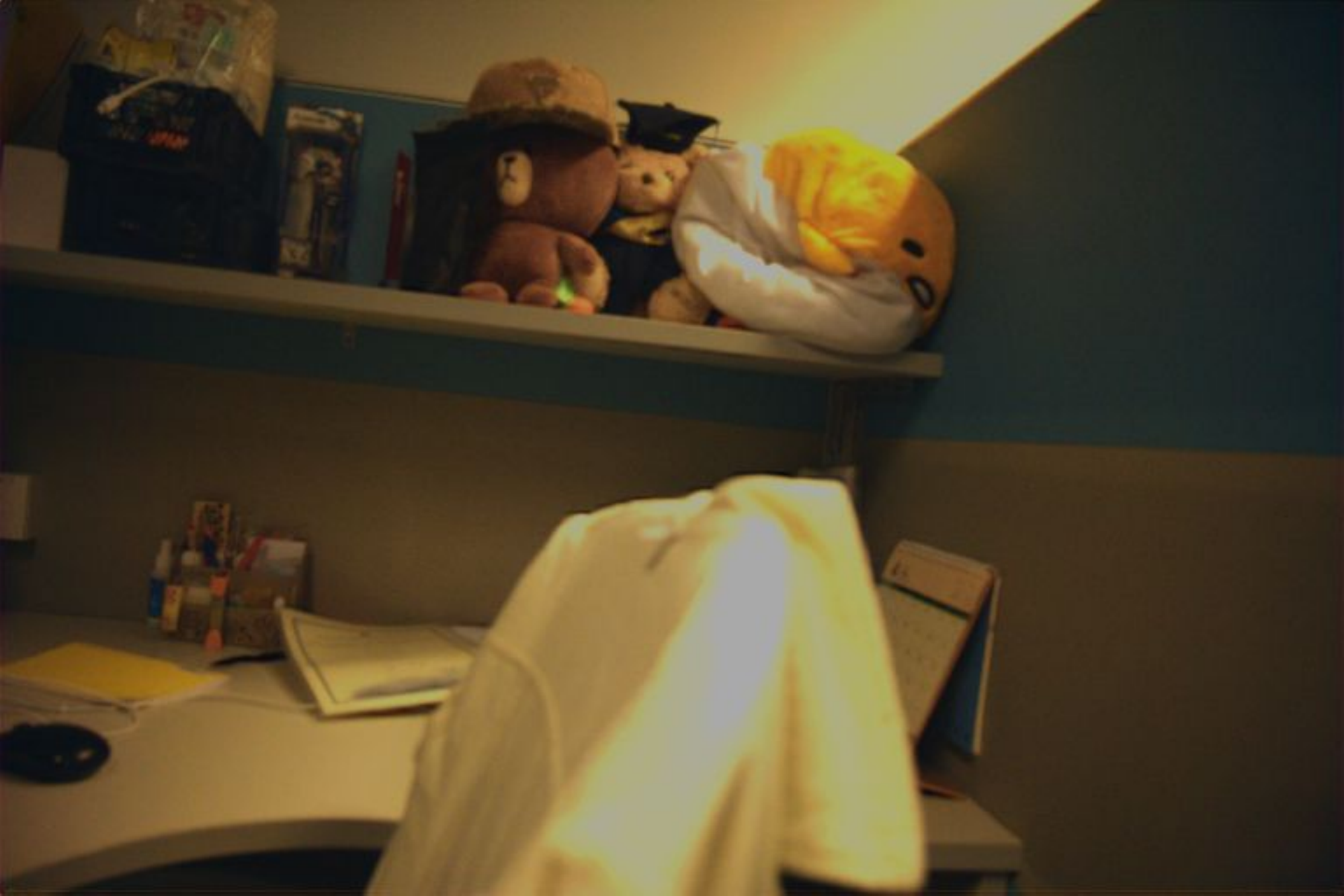} &    \includegraphics[width=0.19\linewidth]{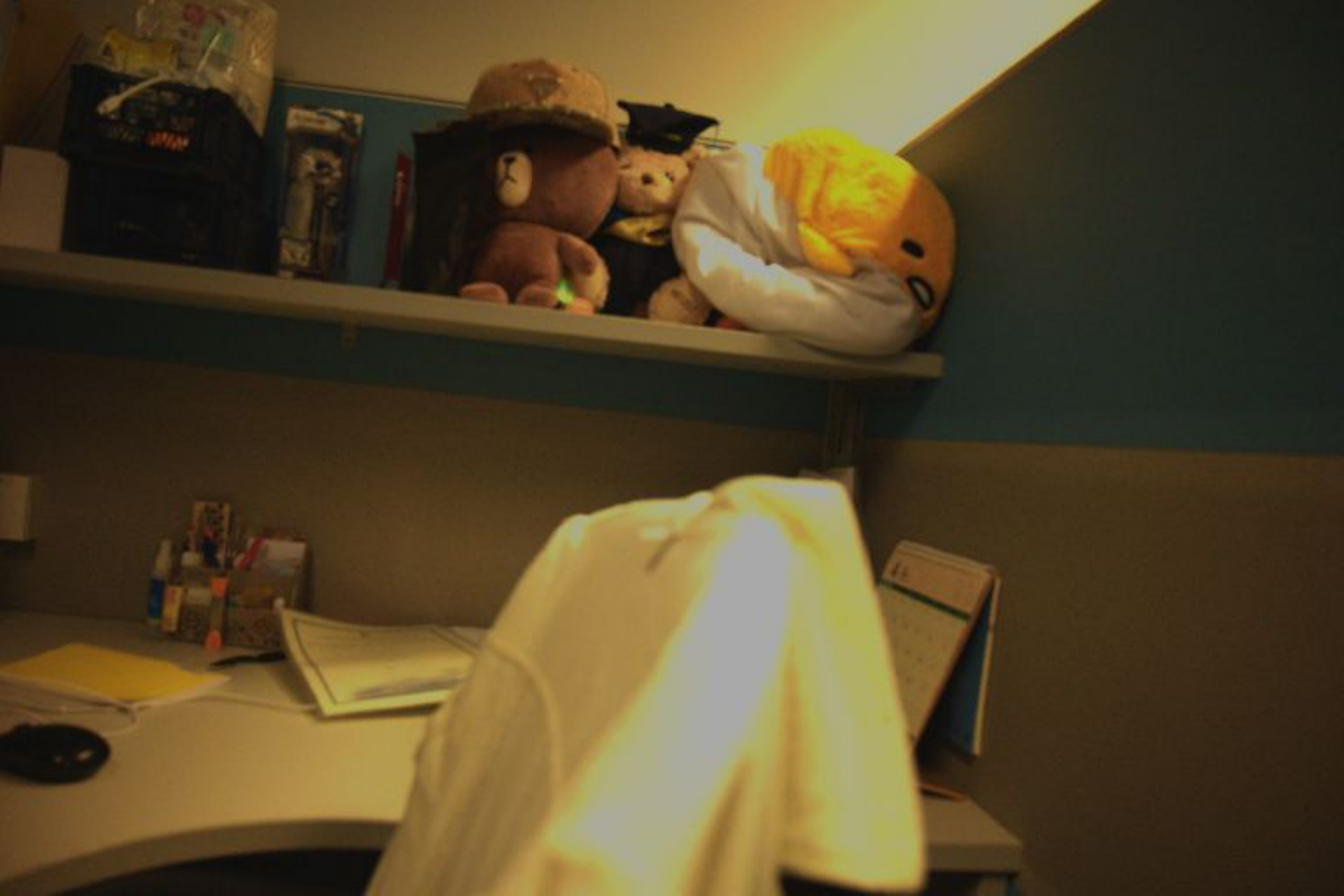}     \\
     
     \includegraphics[width=0.19\linewidth]{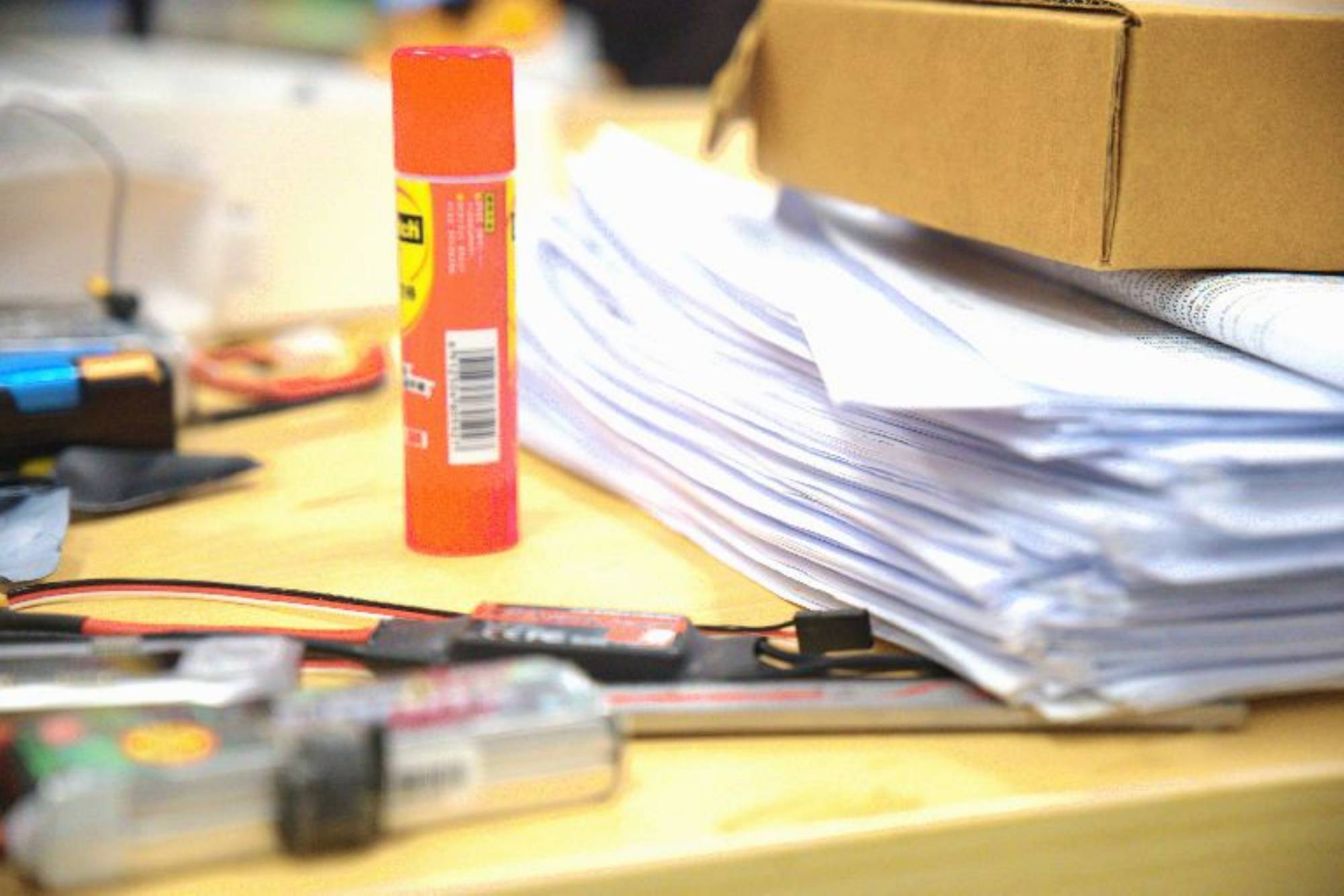} &
     \includegraphics[width=0.19\linewidth]{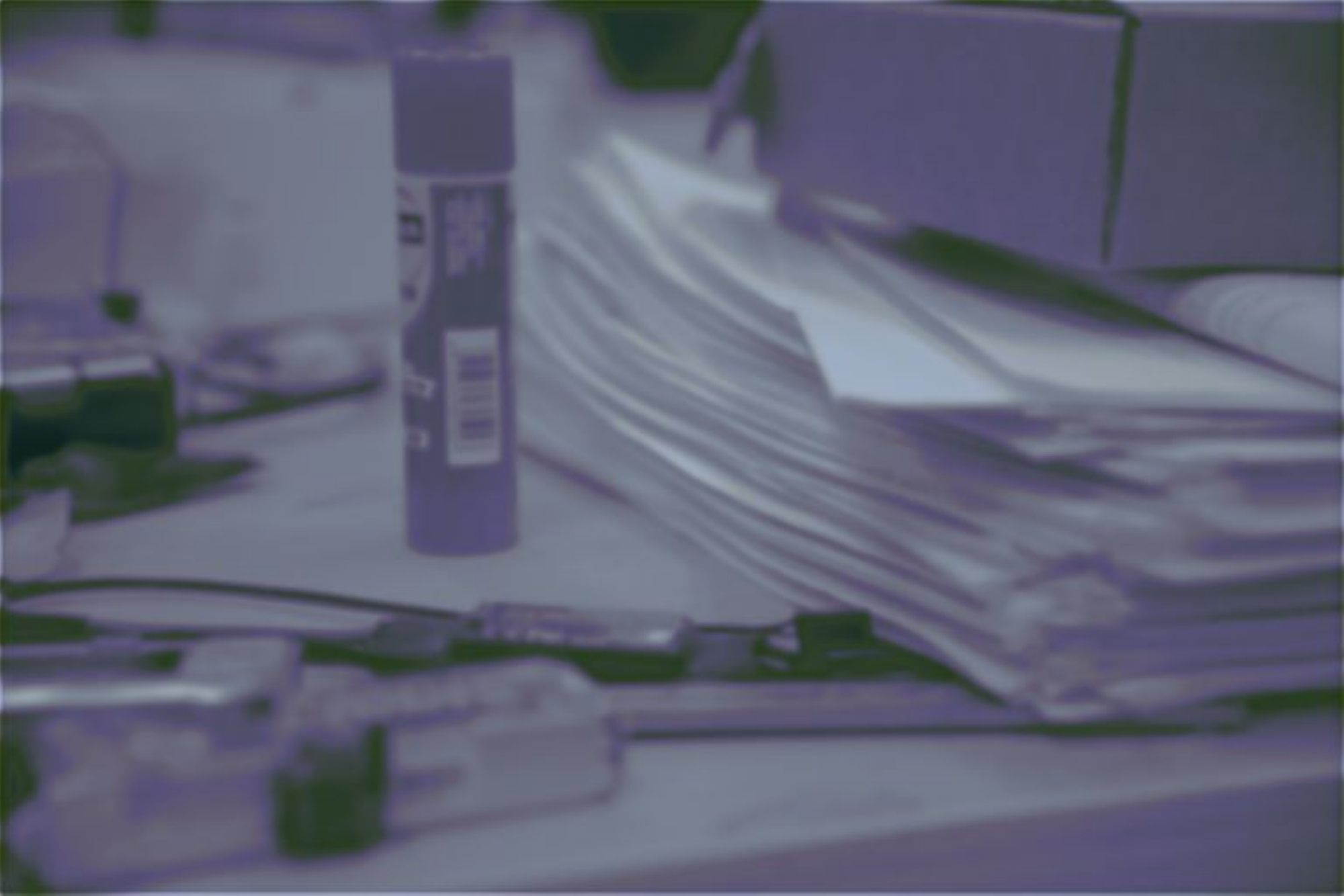} &    \includegraphics[width=0.19\linewidth]{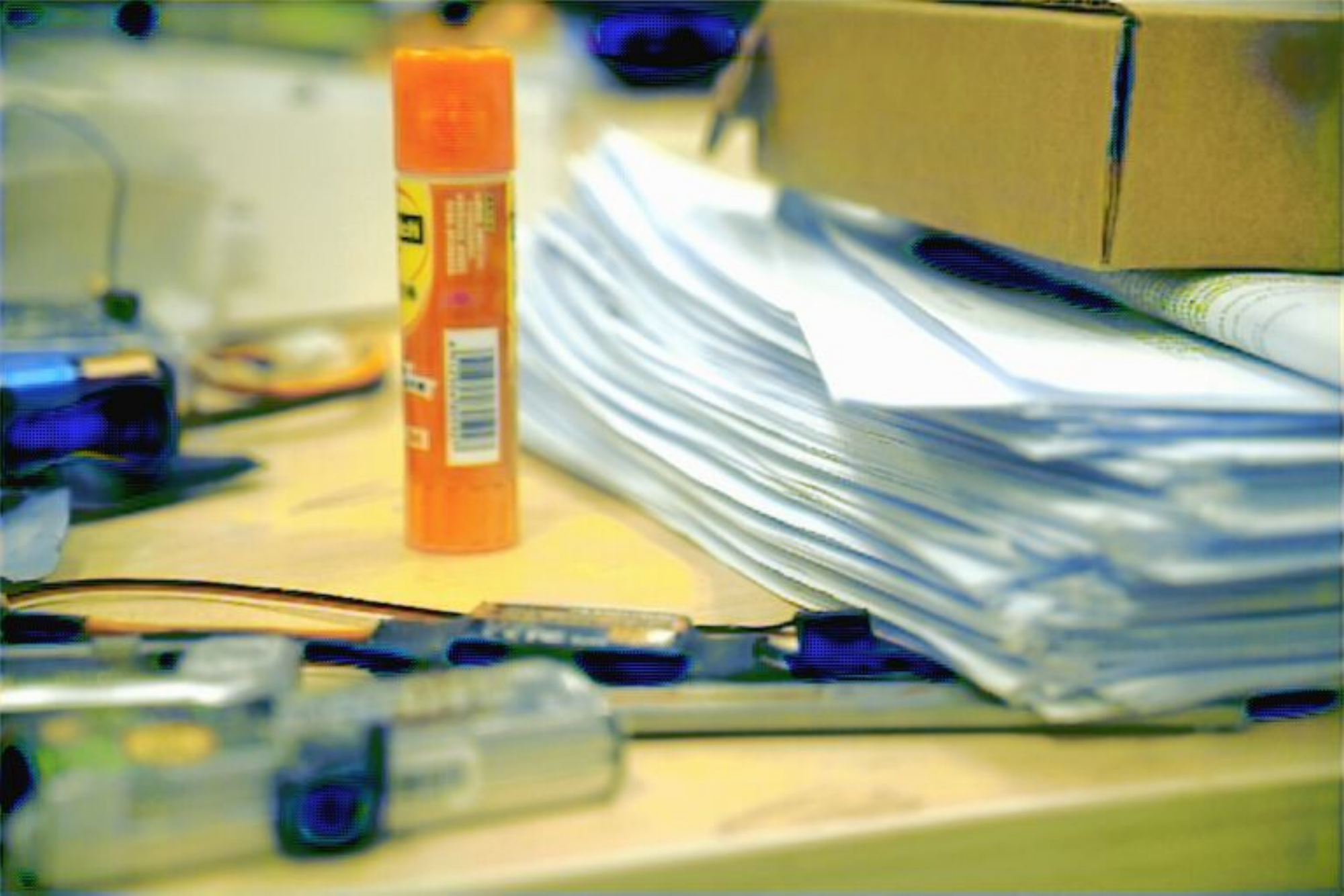}  & \includegraphics[width=0.19\linewidth]{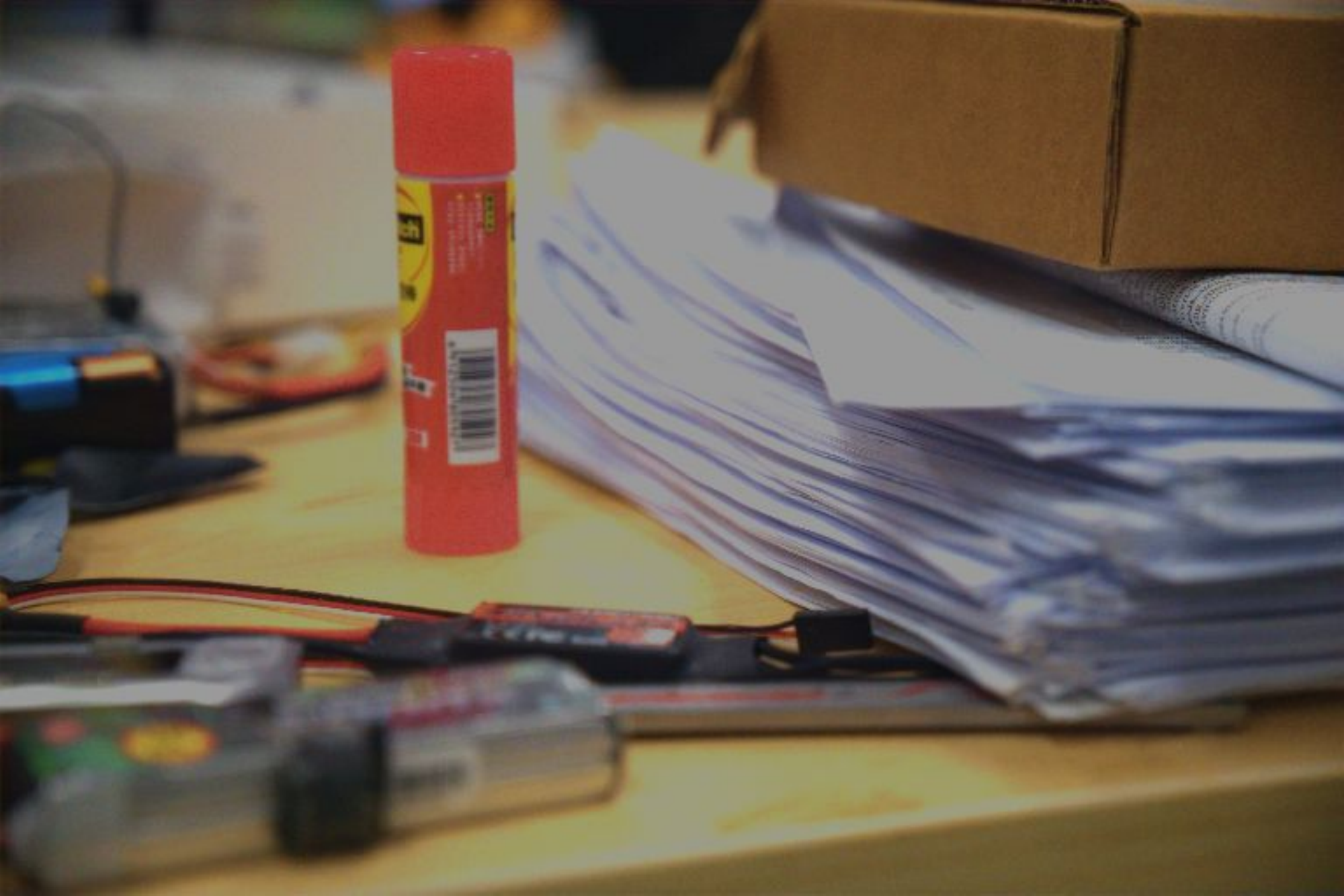} &    \includegraphics[width=0.19\linewidth]{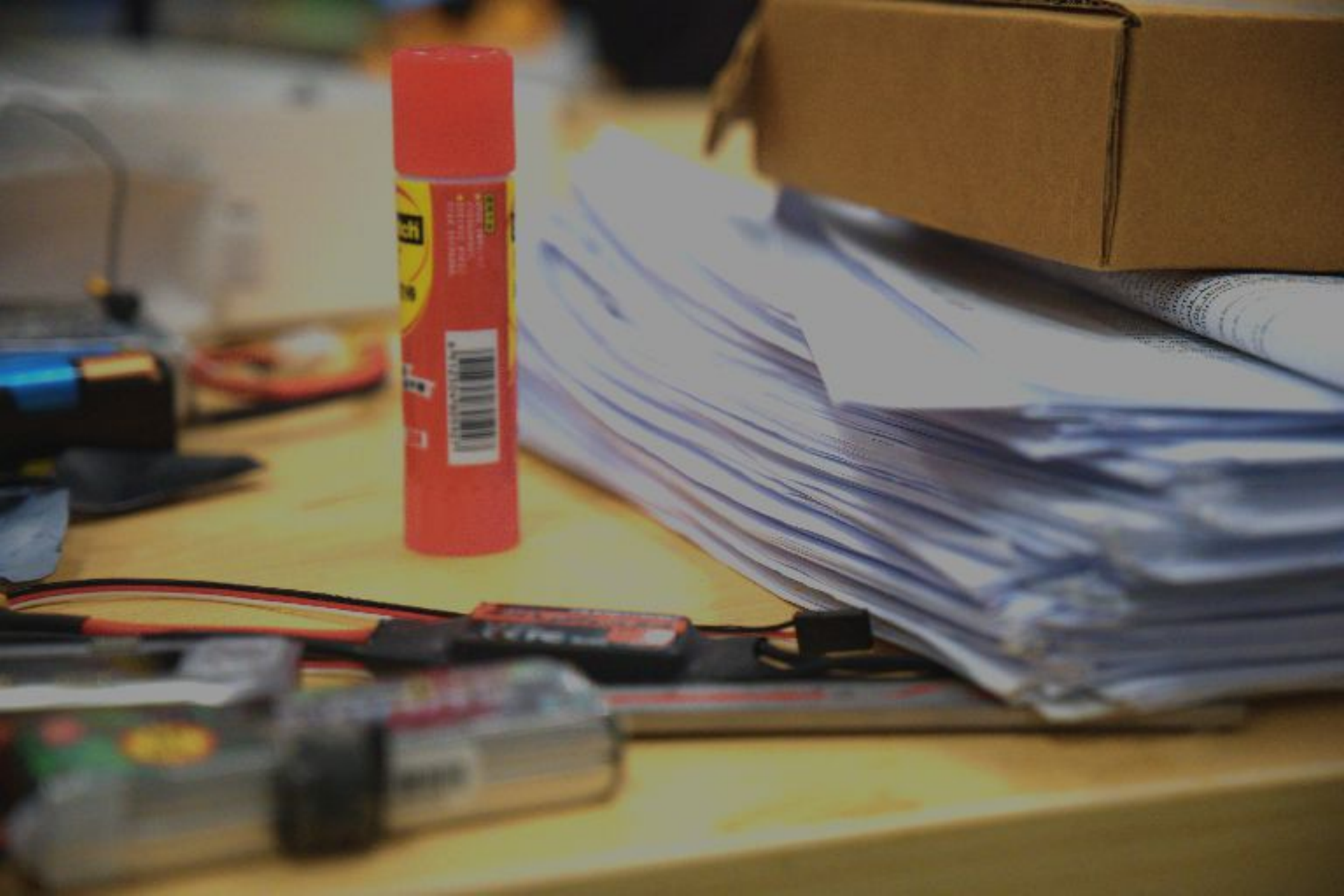}     \\
     
     \includegraphics[width=0.19\linewidth]{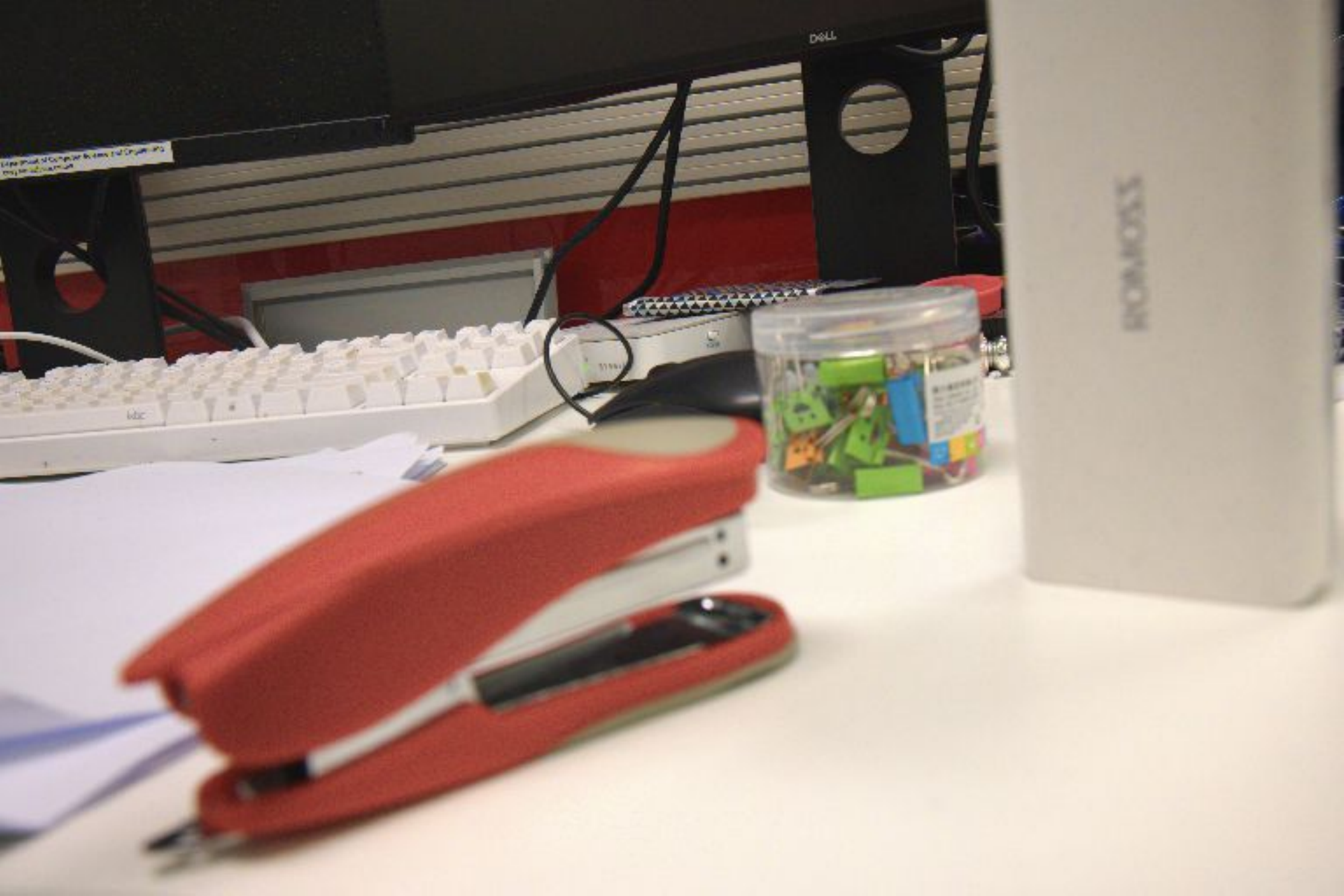} &
     \includegraphics[width=0.19\linewidth]{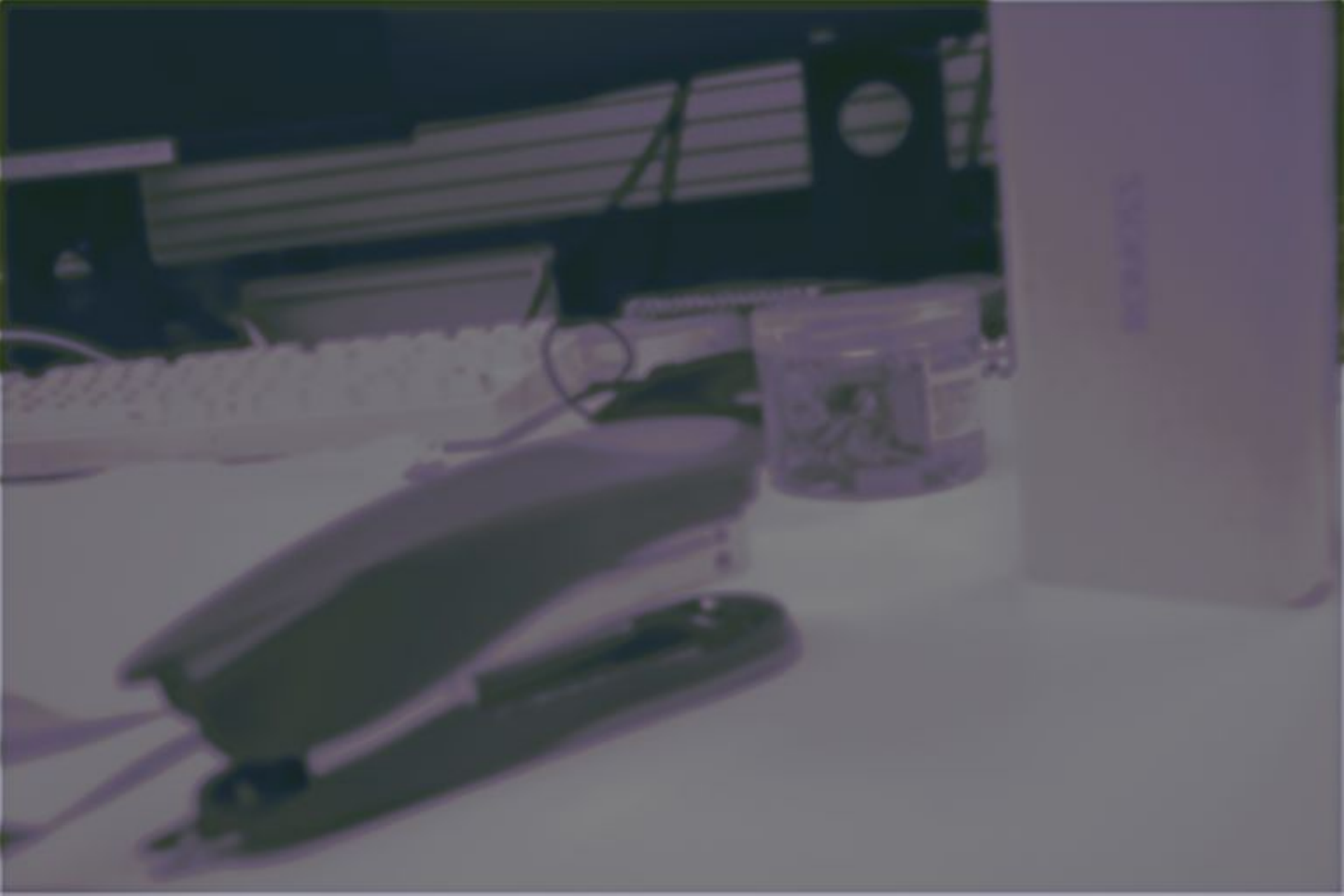} &    \includegraphics[width=0.19\linewidth]{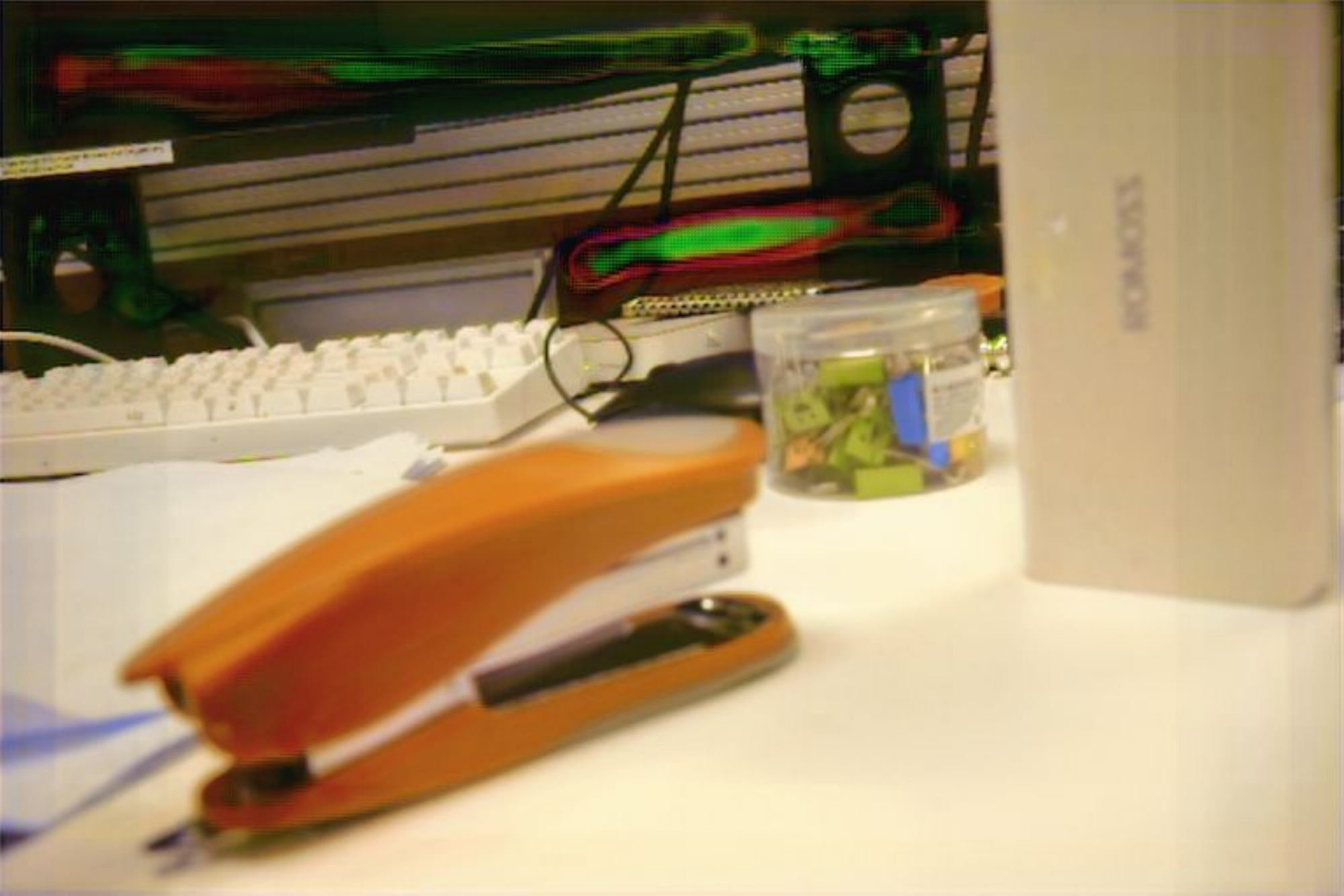}  & \includegraphics[width=0.19\linewidth]{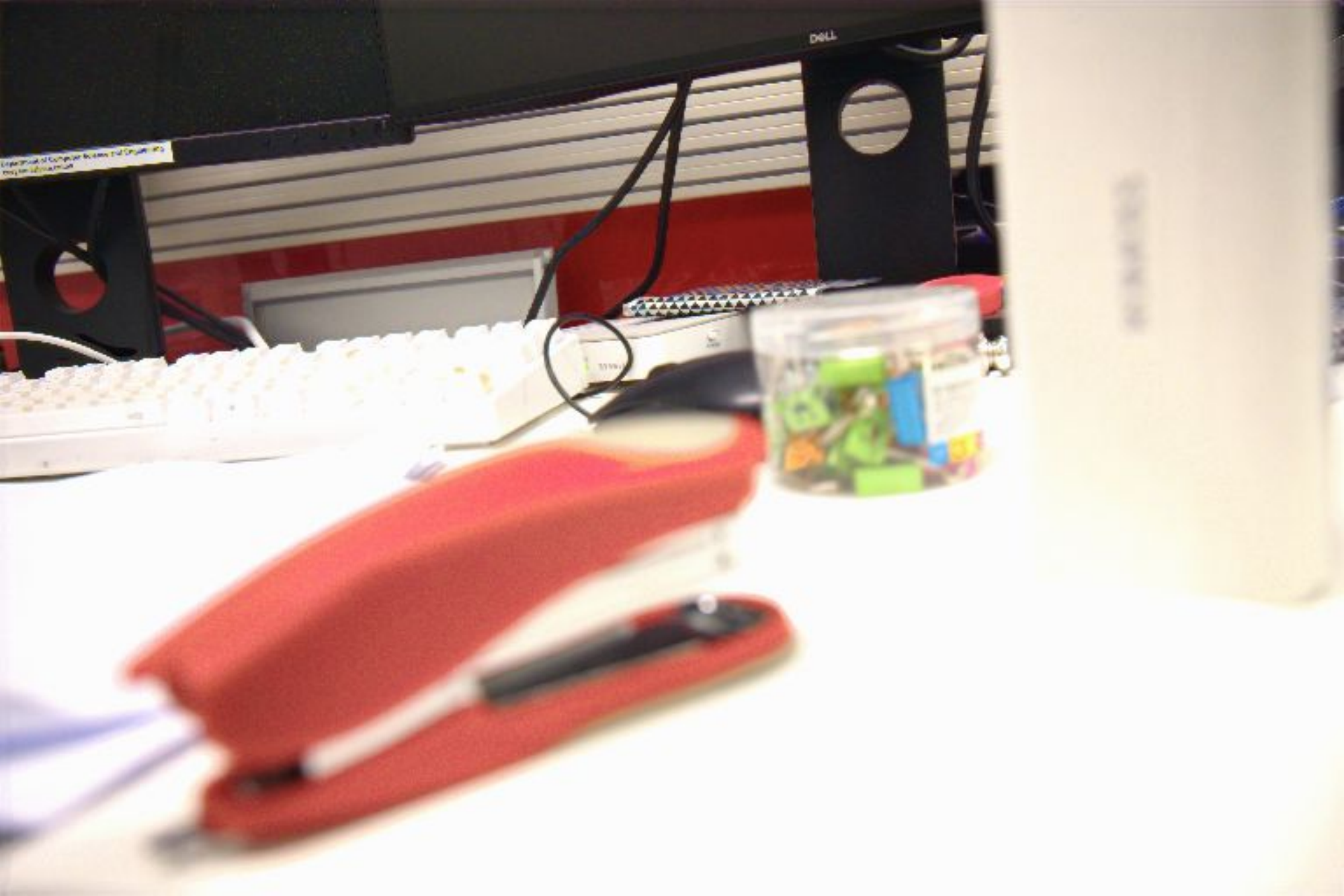} &    \includegraphics[width=0.19\linewidth]{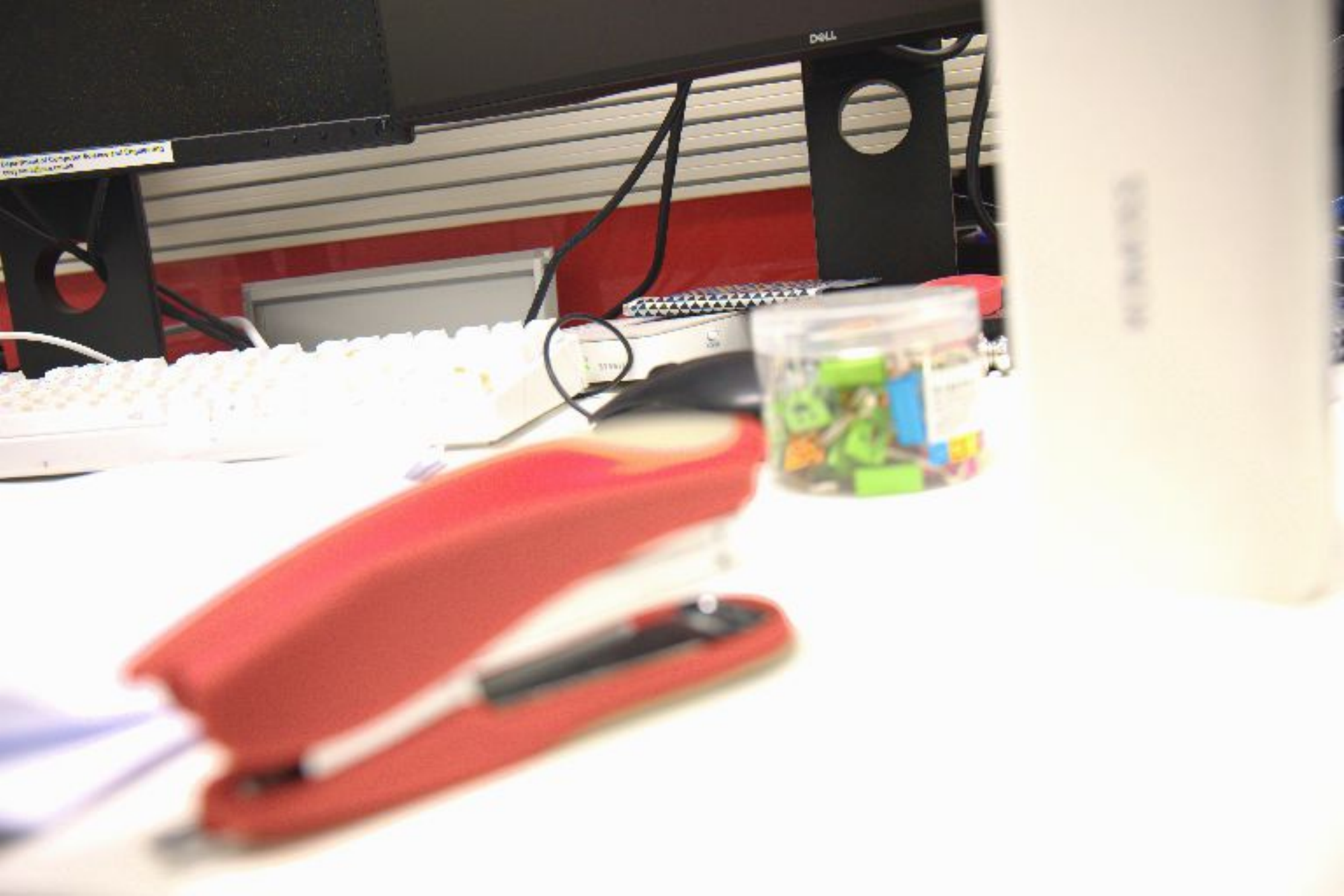}     \\
     
    \end{tabular}
    \caption{Comparison with different methods. Simple parameter concatenation~\cite{chen2017fast} fails to generate natural images. Decouple learning~\cite{fan2018decouple} can generate correct illuminance level but with incorrect color. Our method achieves the most similar result to the ground truth image. Zoom-in to view details.}
    \label{fig:simulation}
\end{figure*}

\begin{figure}[!ht]
    \centering 
    \begin{tabular}{@{}c@{\hspace{0.01mm}}c@{\hspace{0.5mm}}c@{\hspace{0.5mm}}c@{\hspace{0.5mm}}c@{}}
    \rotatebox[origin=c]{90}{\hspace{0.5mm}\scriptsize Input} &
     \raisebox{-.5\height}{\includegraphics[width=0.24\linewidth]{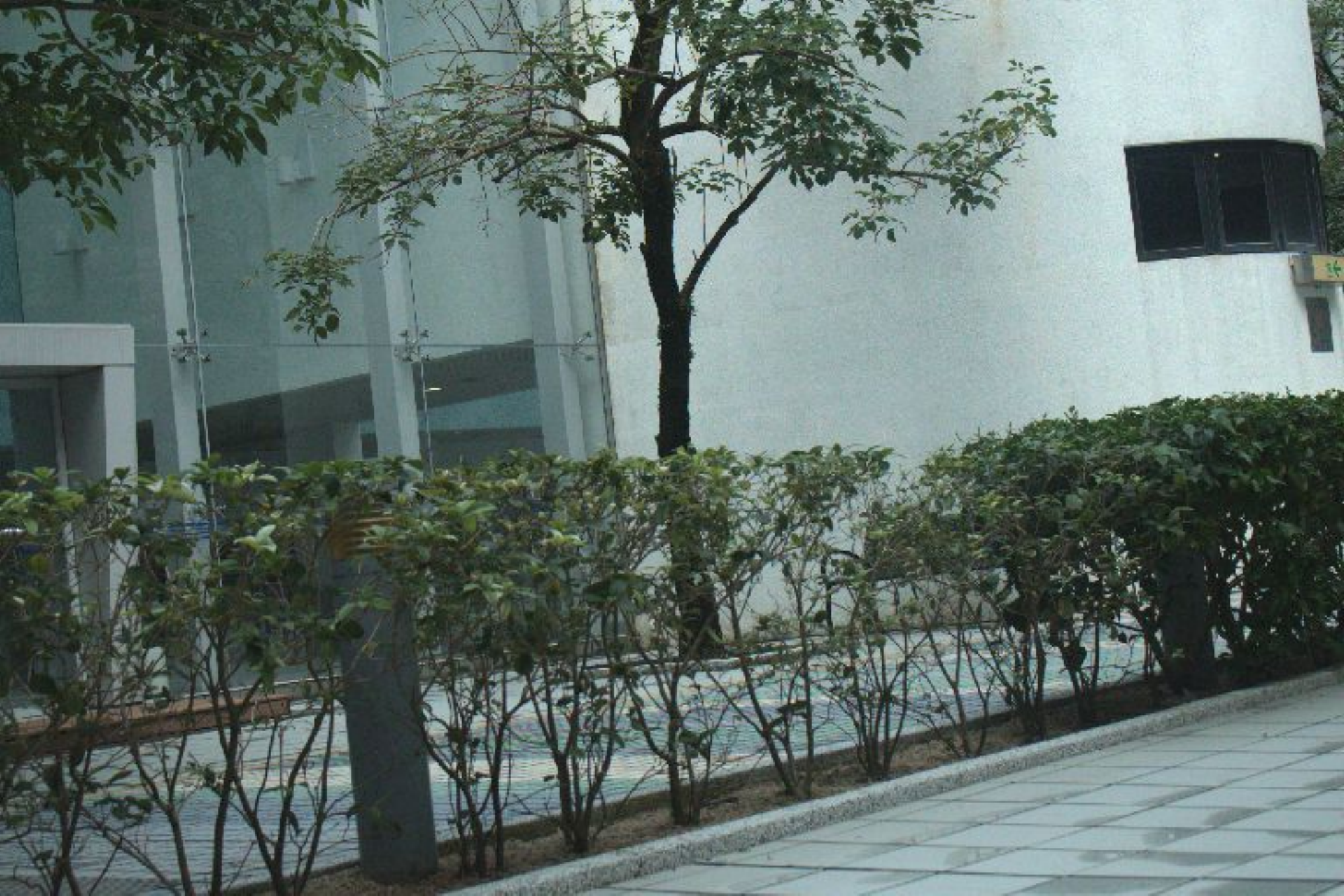}} &    \raisebox{-.5\height}{\includegraphics[width=0.24\linewidth]{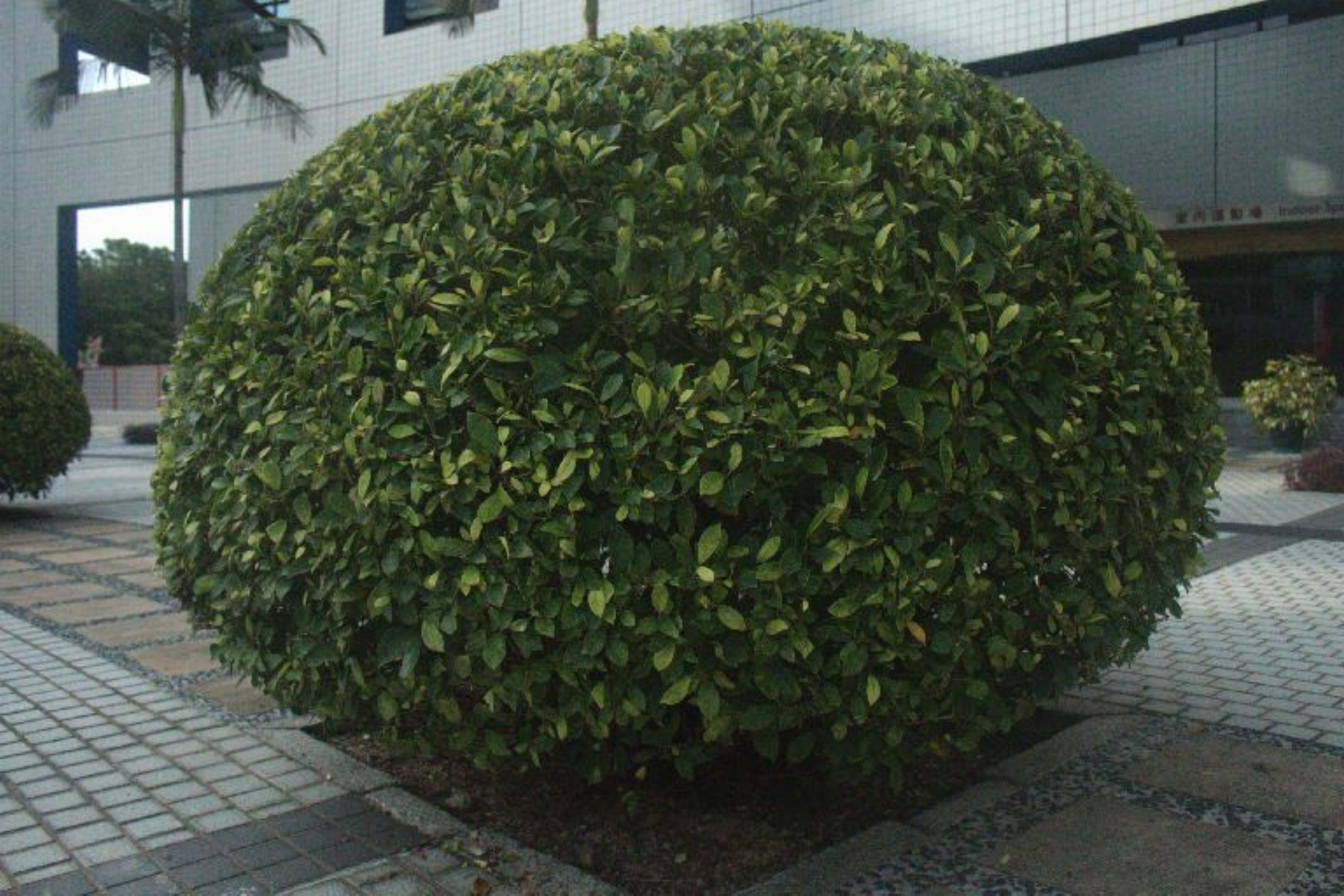}}  & \raisebox{-.5\height}{\includegraphics[width=0.24\linewidth]{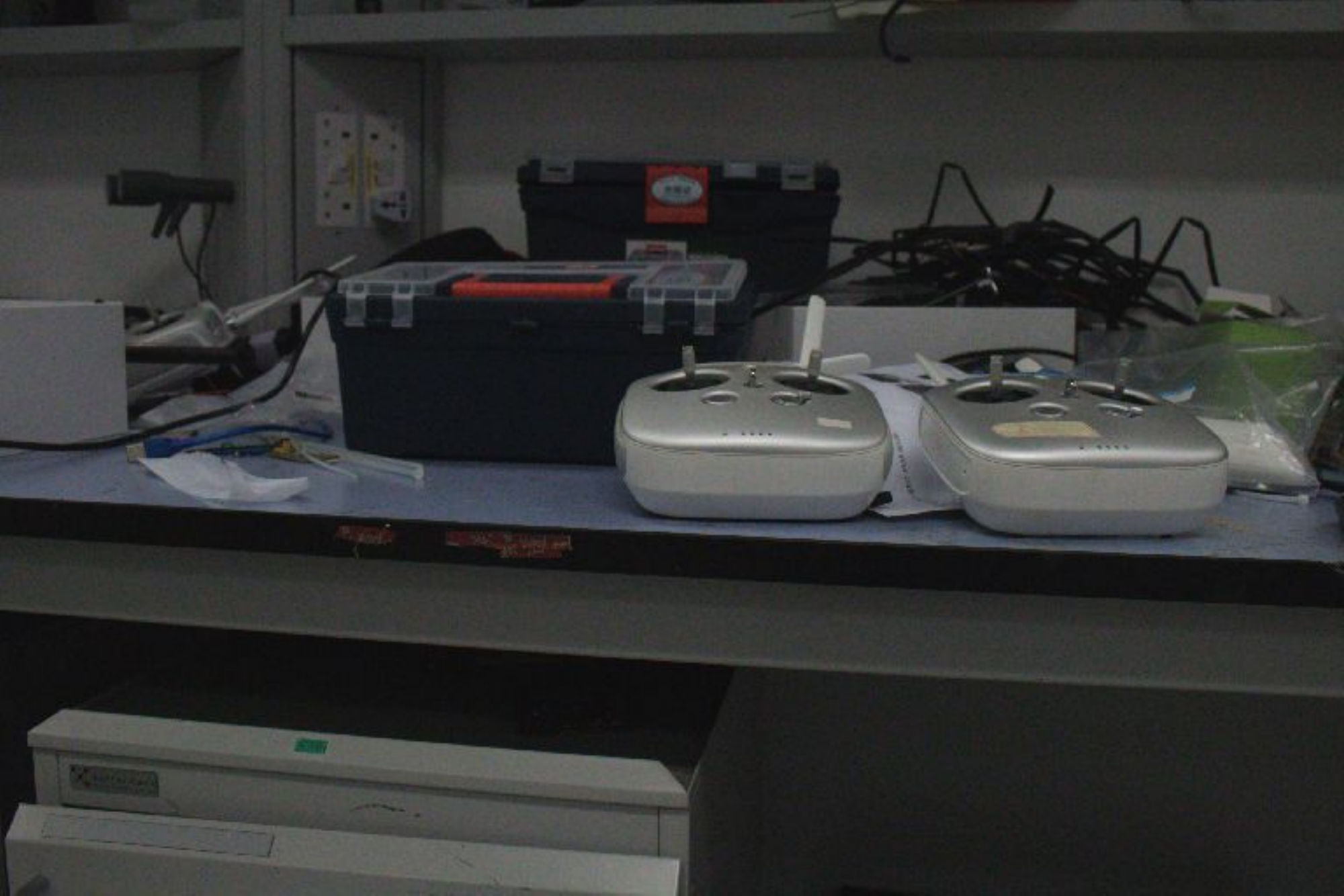}} &    \raisebox{-.5\height}{\includegraphics[width=0.24\linewidth]{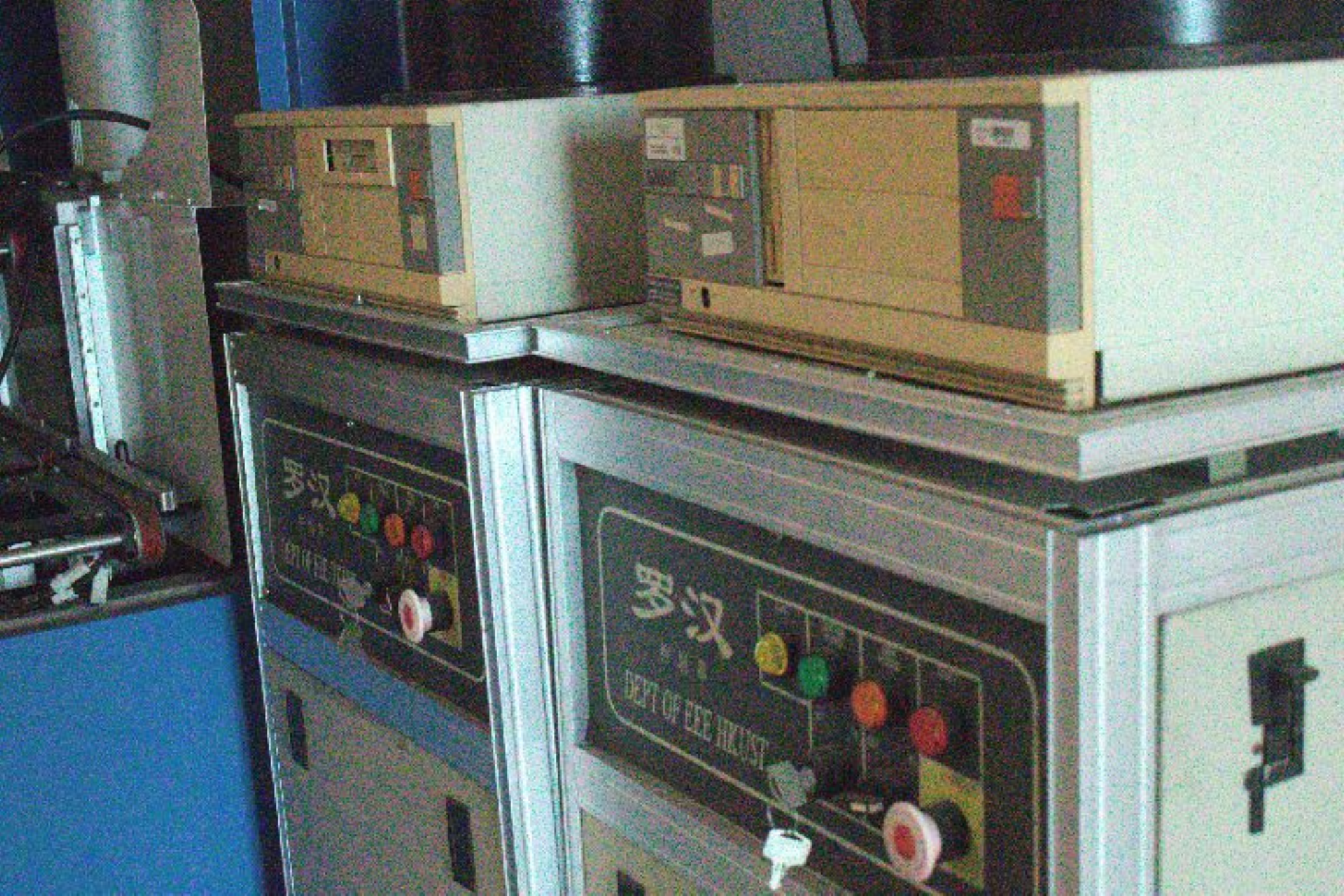}}     \\
    \rotatebox[origin=c]{90}{\hspace{0.5mm}\scriptsize Output} &
     \raisebox{-.5\height}{\includegraphics[width=0.24\linewidth]{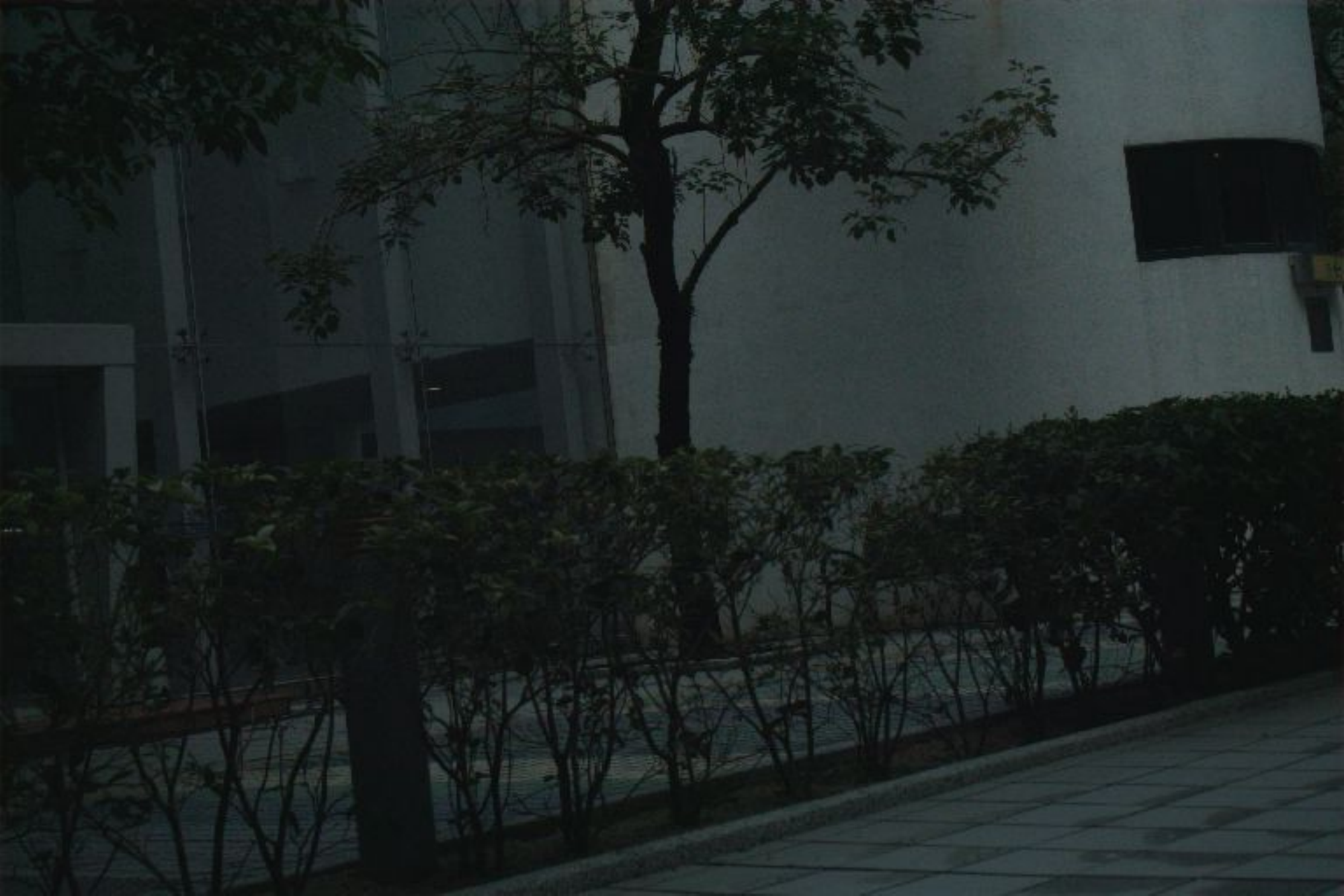}}       &    \raisebox{-.5\height}{\includegraphics[width=0.24\linewidth]{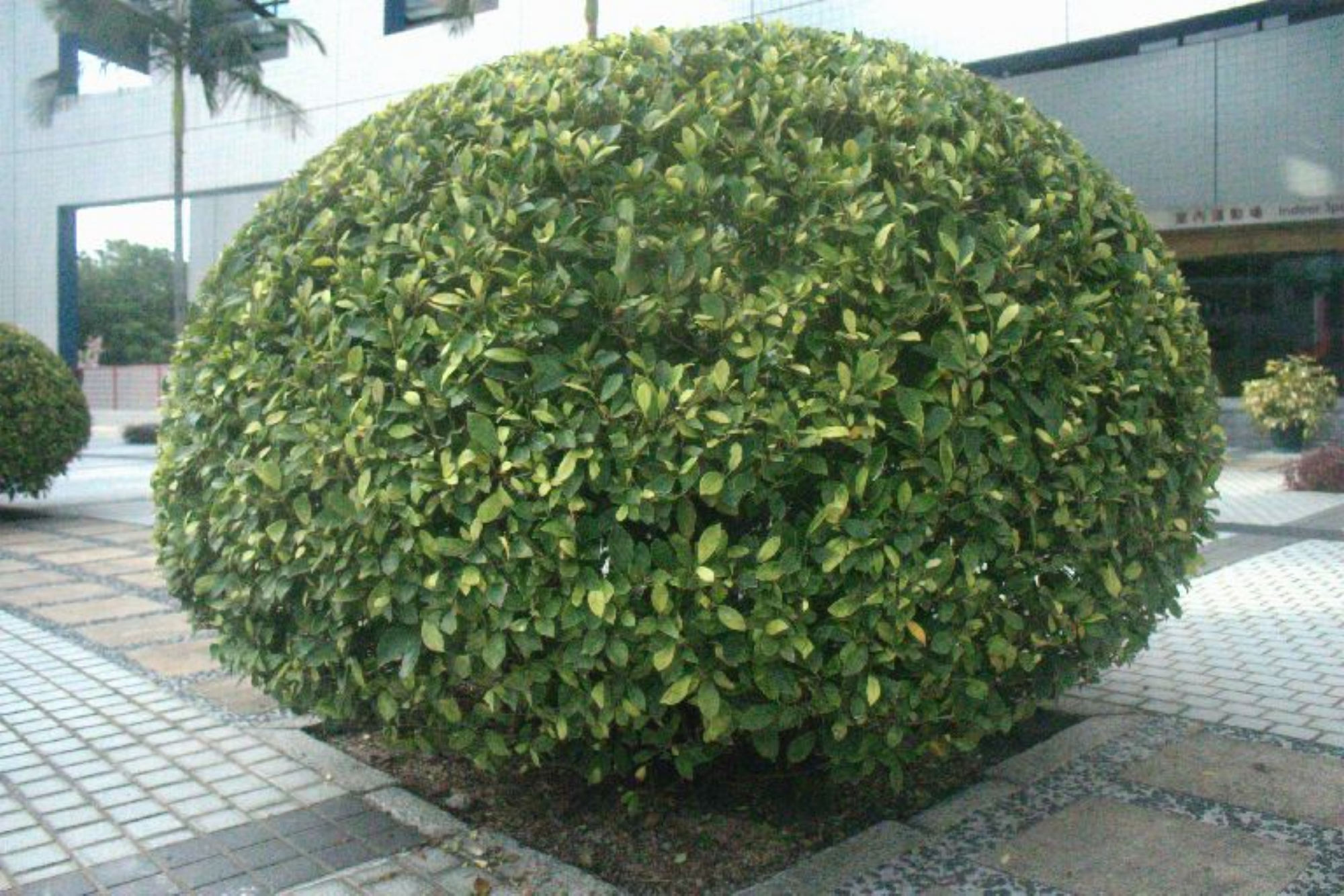}}  & \raisebox{-.5\height}{\includegraphics[width=0.24\linewidth]{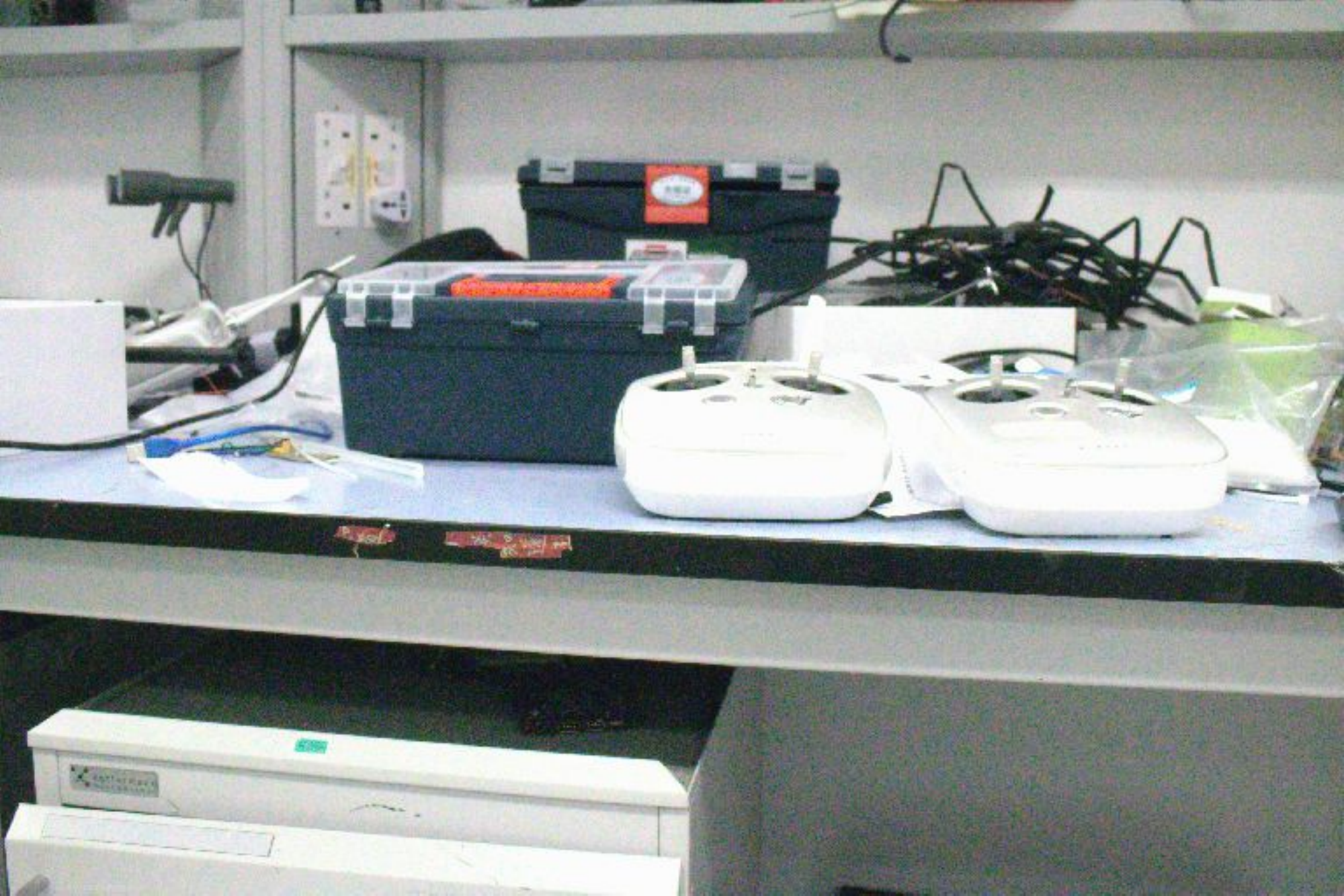}} &    \raisebox{-.5\height}{\includegraphics[width=0.24\linewidth]{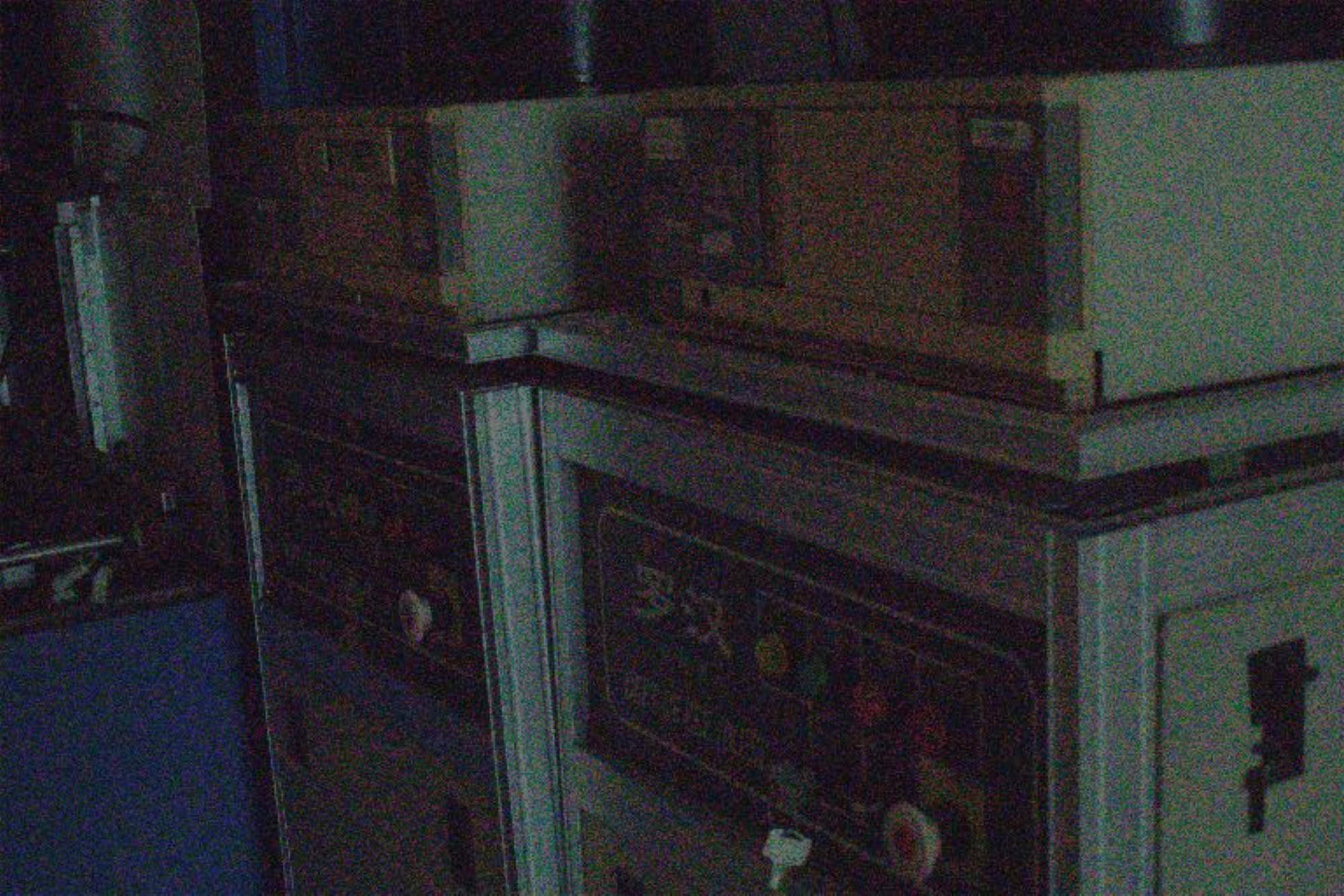}}        \\
     \rotatebox[origin=c]{90}{\hspace{0.5mm}\scriptsize  GT }&
     \raisebox{-.5\height}{\includegraphics[width=0.24\linewidth]{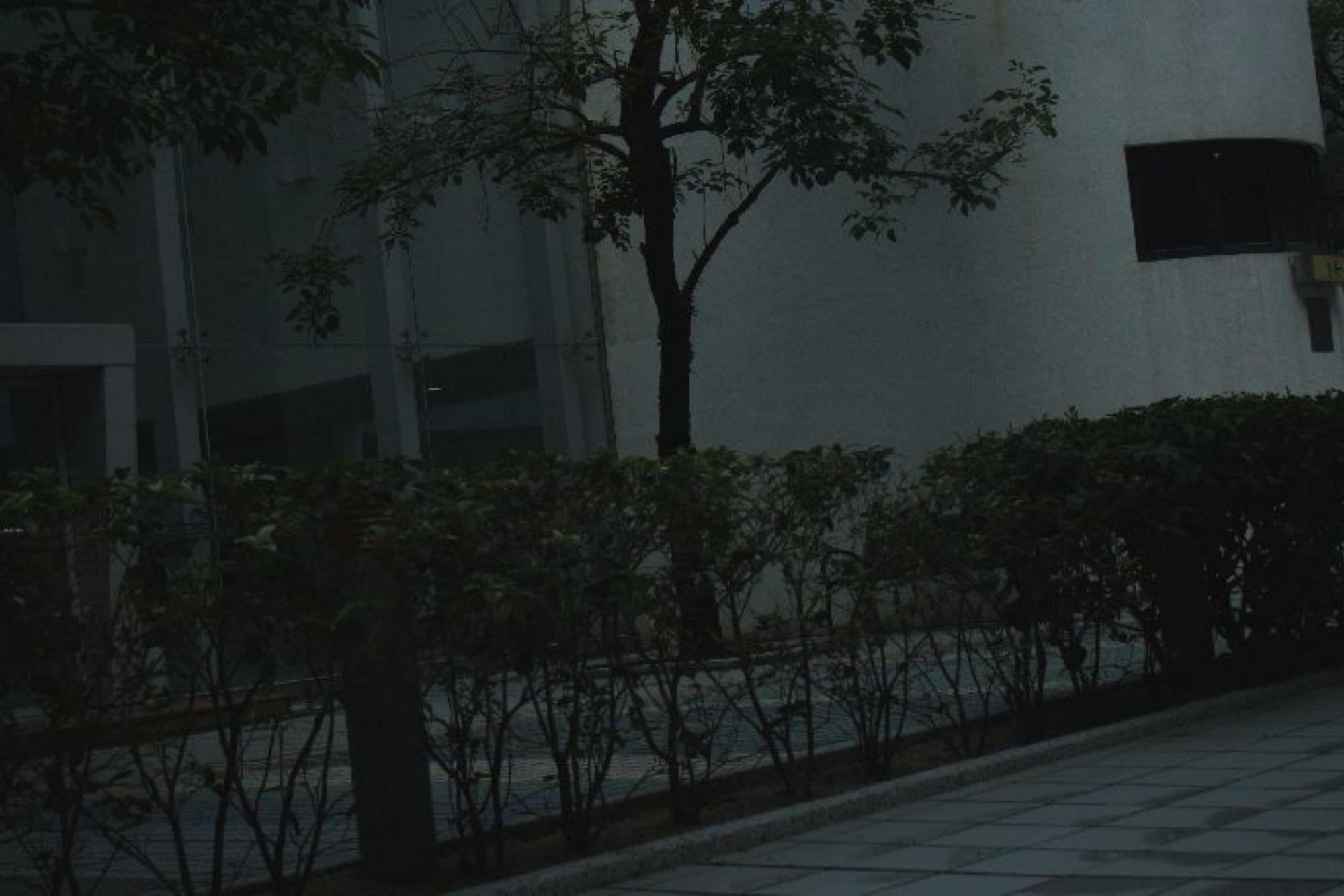}}       &    \raisebox{-.5\height}{\includegraphics[width=0.24\linewidth]{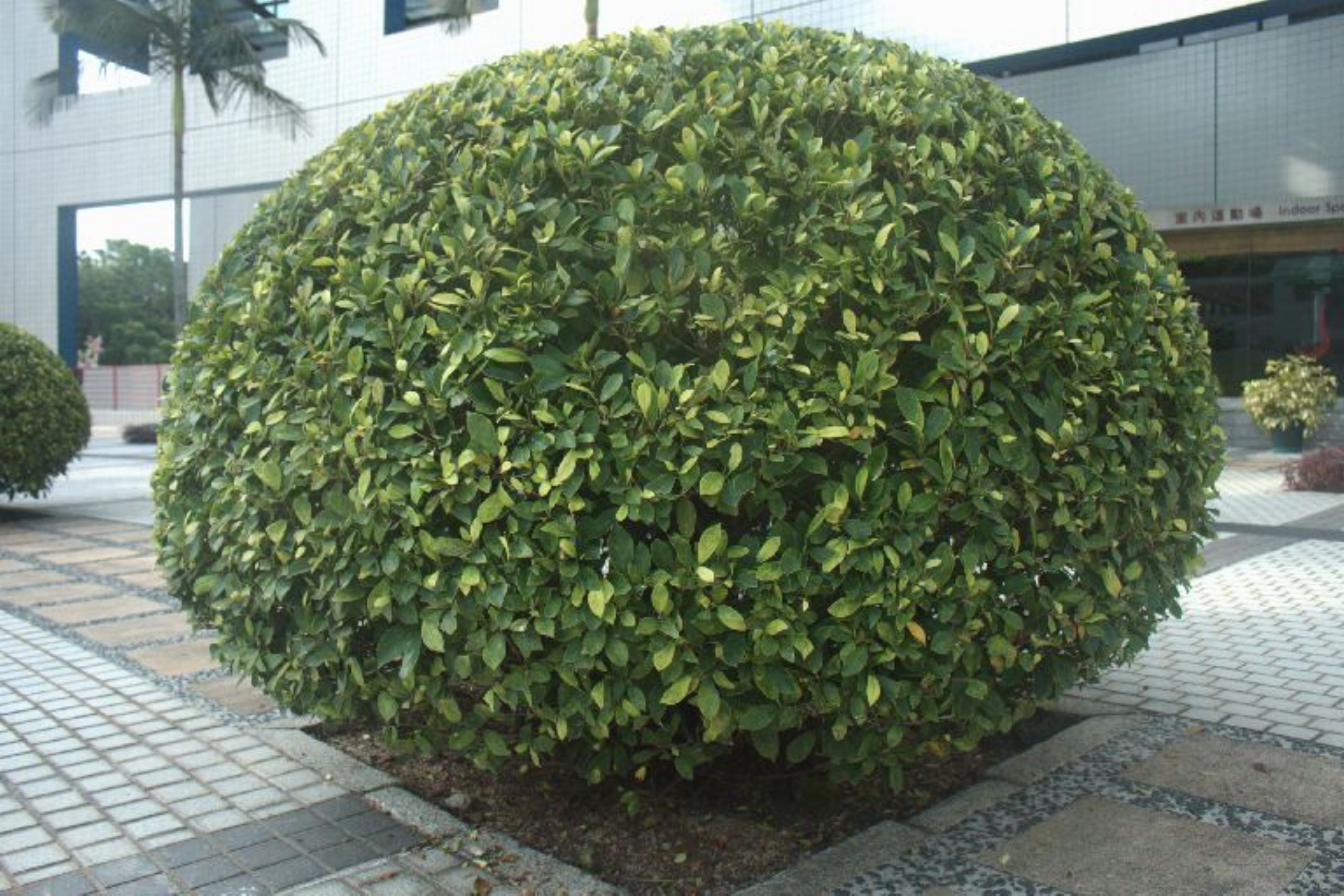}}  & \raisebox{-.5\height}{\includegraphics[width=0.24\linewidth]{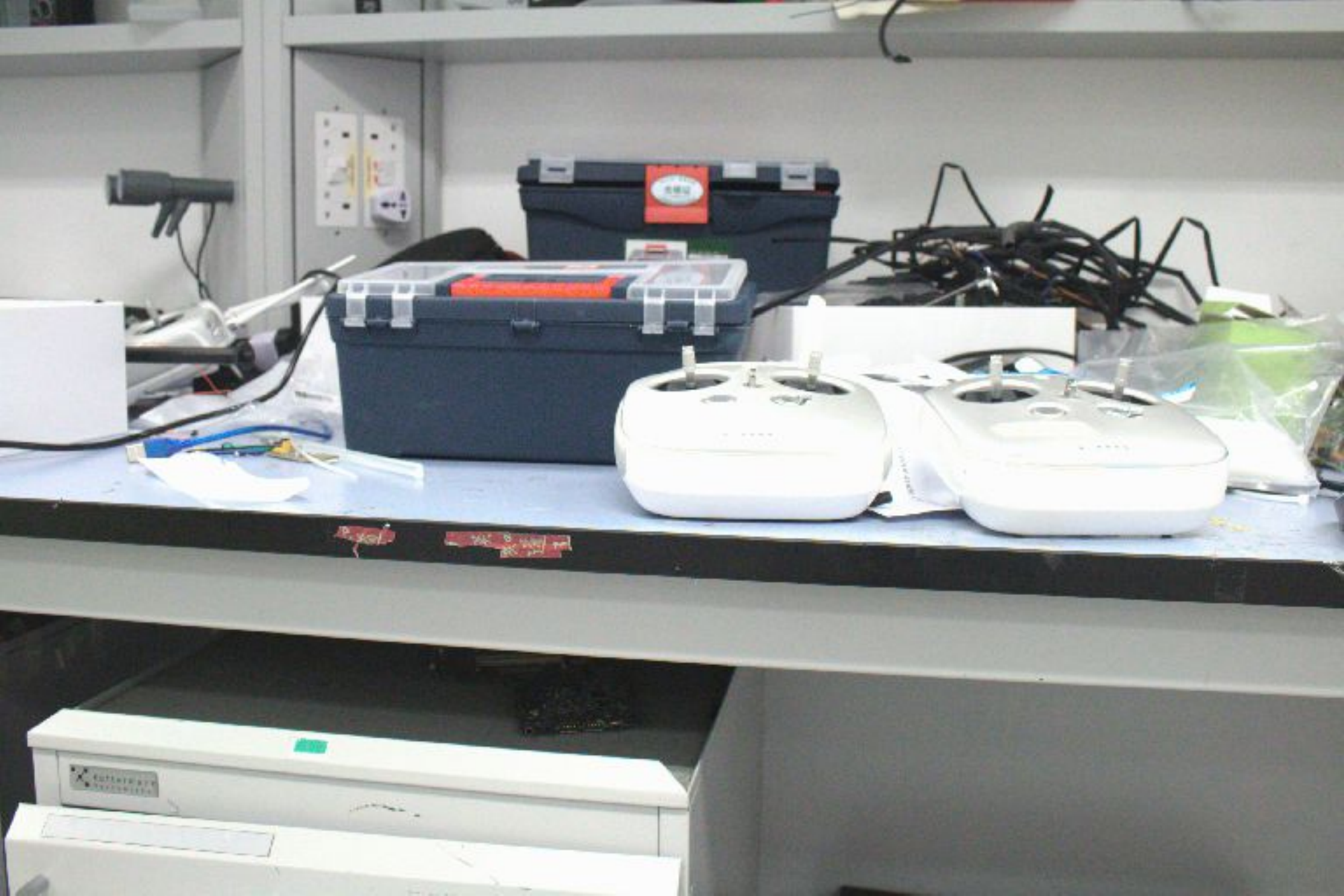}} &    \raisebox{-.5\height}{\includegraphics[width=0.24\linewidth]{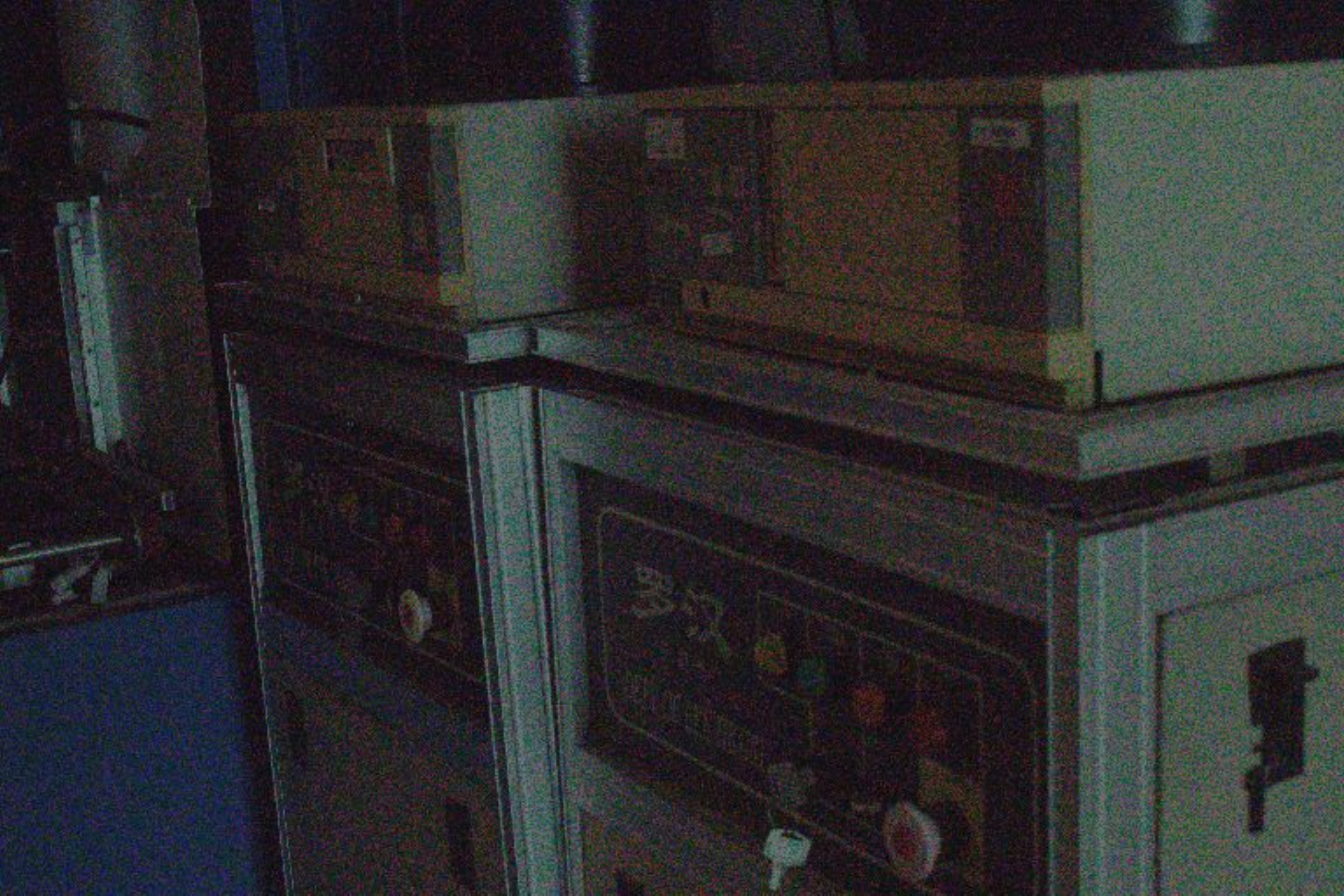}}        \\
     
    \end{tabular}
    \caption{Cross-dataset validation on the Canon 70D dataset with model trained on the Nikon Z6 dataset. Zoom-in to view details.}
    \label{fig:cross}
\end{figure}


\subsection{Experimental setup}

\mypara{Data preparation.} We randomly split the dataset into a training set of $90\%$ of image sequences and a testing set of $10\%$ of image sequences. As pre-processing, we also unpack the Bayer raw pattern to four RGGB channels and deduct the recorded black level. The unpacked raw data is then normalized between 0 and 1. We obtain the camera parameters (ISO, exposure time, and aperture) and pre-calibrated noise level function from the camera meta-data. We also retrieve the color processing pipeline parameters (white balance, tone curve, and color matrix), which we utilize for rendering from raw-RGB to sRGB color space for visualization. 

\mypara{Pair selection.}
We adopt different pair selection strategies to facilitate the training for each module. In the exposure training stage, any random pairs from the same scene sequence can be selected. In the noise manipulation stage, we only choose pairs in which the output has low ISO settings (100, 200, and 400). Similarly, we only choose pairs from small to large aperture sizes to magnify the defocus blur. 

\mypara{Training details.}
We train the exposure module for 5 epochs, the noise module for 30 epochs, and the aperture module for 30 epochs separately. The model is then jointly finetuned for another 10 epochs. The batch size is set to 2. We use Adam Optimizer with a learning rate of 1e-3 for all three modules and reduce the learning rate by a factor of 10 every 20 epochs. The entire training process takes about 48 hours on two NVIDIA RTX 2080 Ti GPUs.
 
 \subsection{Baselines}
 Our designed baseline models aim to learn a direct mapping to generate raw images given different camera settings.
 
\mypara{Parameter concatenation (PC).}
Image to image translation has shown effectiveness in multiple image generation tasks~\cite{chen2017fast,isola2017image}. In our case, we can easily modify these approaches to adapt to a raw-to-raw translation task. We change the input and output of the model proposed in~\cite{chen2017fast} to unpacked raw images. Other settings, including loss types and network structures, remain the same. The model takes the camera input parameters and output parameters settings as additional input channels. The baselines with and without noise level maps generate similar results.

\mypara{Decouple learning (DL).}
Another structure applicable to our case is DecoupleNet~\cite{fan2018decouple}. The model learns parameterized image operators by directly associating the operator parameters with the weights of convolutional layers. Similarly, we need to modify the input and output to raw data. Operator parameters are replaced by input and output camera parameters in our case.

\subsection{Image quality evaluation}
Fig.~\ref{ fig:pipeline} visualizes the output of each module of our method, and Fig.~\ref{fig:simulation} provides a qualitative comparison of the visual quality of raw data simulated by different methods. Our simulated raw has higher visual fidelity and fewer artifacts than the baselines. Without prior knowledge of lens design, baseline models struggle to figure out the relationship between camera settings and output raw images. As a result, they generate images with incorrect illuminance and biased color. 

Table~\ref{tab:ablation} shows the quantitative analysis of our model and baselines trained on the Nikon Dataset. We also evaluate the average PSNR and SSIM of the generated raw data after each module. Note that the PSNR and SSIM in step NS do not show significant improvement to step EXP because they highly depend on the target settings of the simulator. For instance, if the input raw is noisy and the target is to add noise, then PSNR and SSIM in step NS can be worse than those of step EXP due to the denoising function of the noise module.  More analysis on specific simulation direction (e.g., from low to high ISO) can be found in the \textbf{supplement}.

\subsection{Cross-dataset validation}
We conduct a cross-dataset evaluation to validate the generalization ability of the proposed model. We simulate Canon 70D raw data using the model trained on the Nikon Z6 dataset. Fig.~\ref{fig:cross} shows that our model generates stable simulation results. Table~\ref{tab:ablation} shows the quantitative results, which implies that baseline methods suffer from severe degradation when testing on different sensors.


\section{Applications}
\begin{figure}[!t]
    \centering 
    \begin{tabular}{@{\hspace{1mm}}c@{\hspace{1mm}}c@{\hspace{1mm}}c@{\hspace{1mm}}c@{}}
    
    \scriptsize Input & \scriptsize Simulated Image 1 & \scriptsize Simulated Image 2 & \scriptsize Simulated Image 3\\
     \raisebox{-.5\height}{\includegraphics[width=0.24\linewidth]{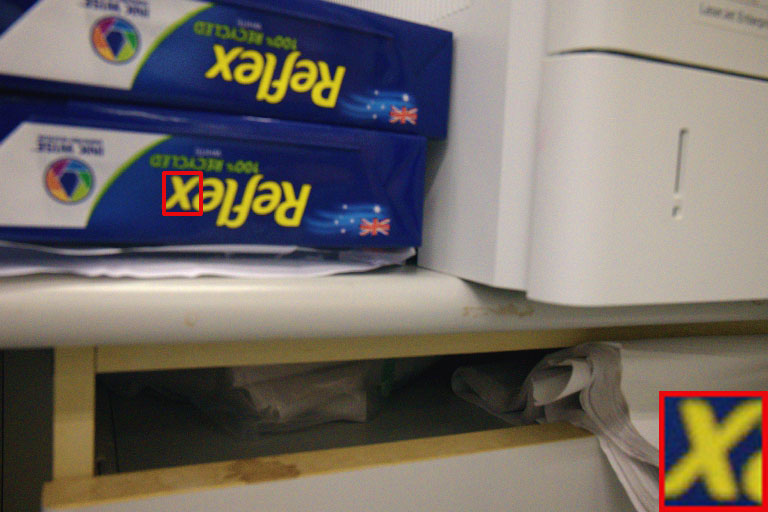}} &    \raisebox{-.5\height}{\includegraphics[width=0.24\linewidth]{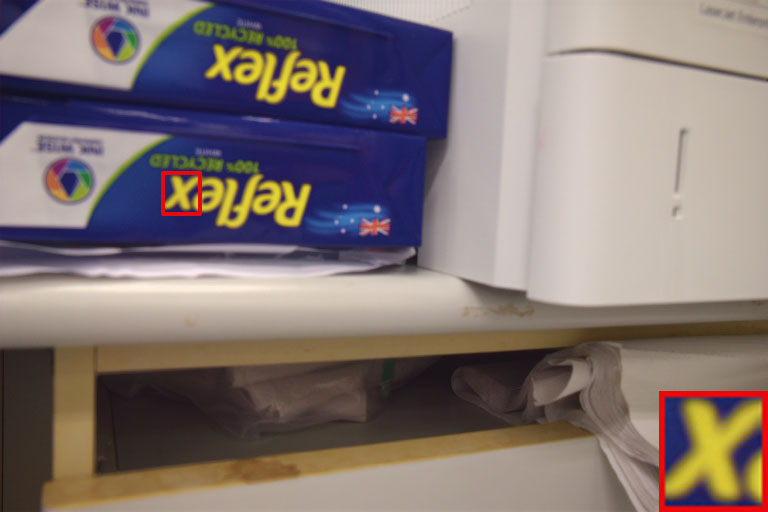}}  & \raisebox{-.5\height}{\includegraphics[width=0.24\linewidth]{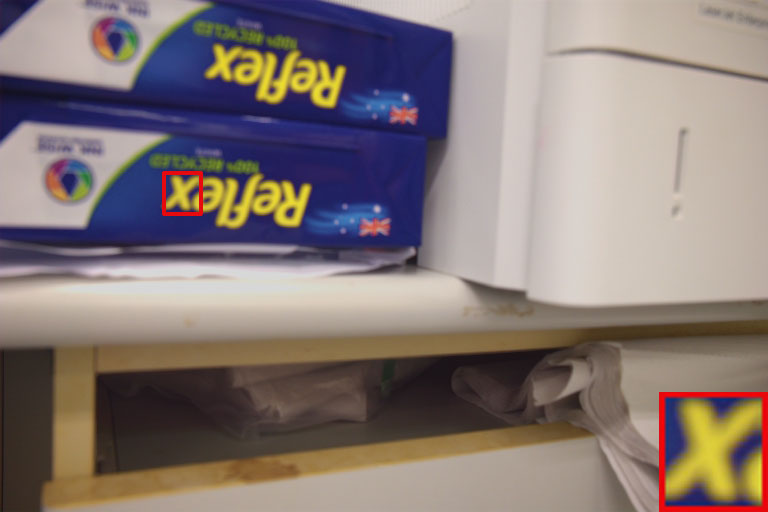}} &    \raisebox{-.5\height}{\includegraphics[width=0.24\linewidth]{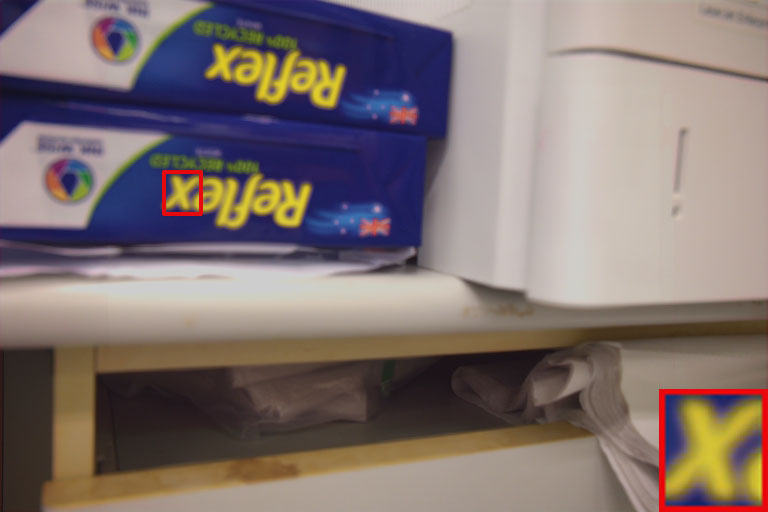}}  \\
      \scriptsize f = 9.0 & \scriptsize f = 6.3 & \scriptsize f = 5.0  & \scriptsize f = 4.0 \\
      &
     \raisebox{-.5\height}{\includegraphics[width=0.24\linewidth]{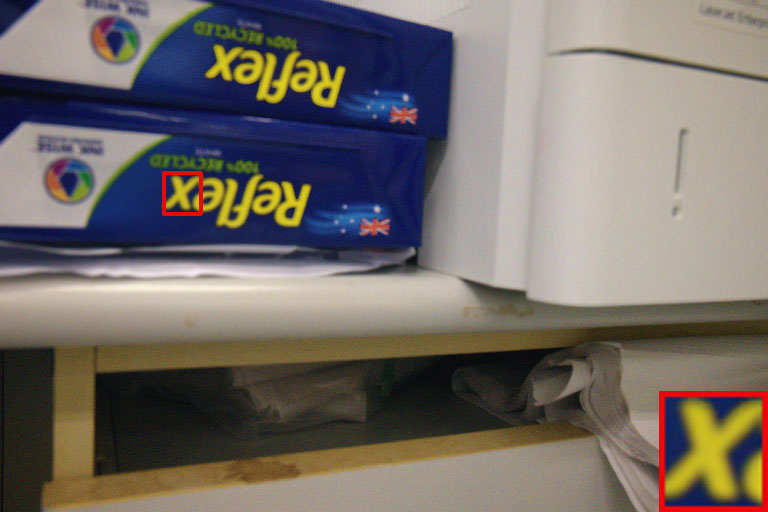}}  & \raisebox{-.5\height}{\includegraphics[width=0.24\linewidth]{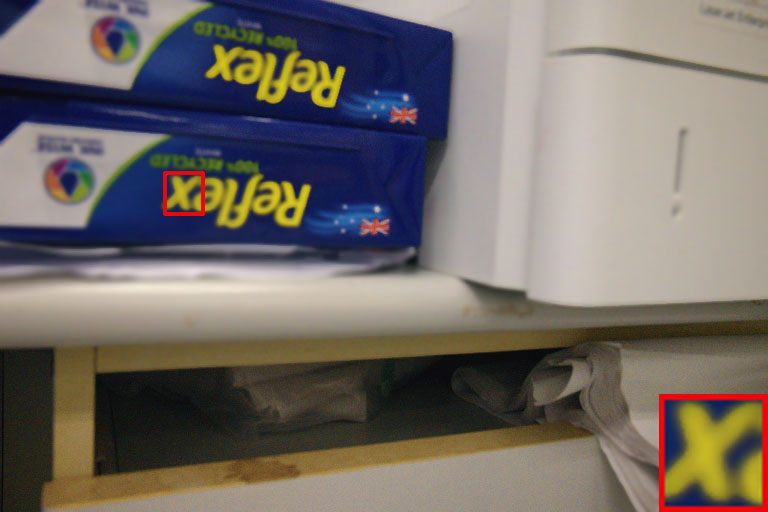}} &    \raisebox{-.5\height}{\includegraphics[width=0.24\linewidth]{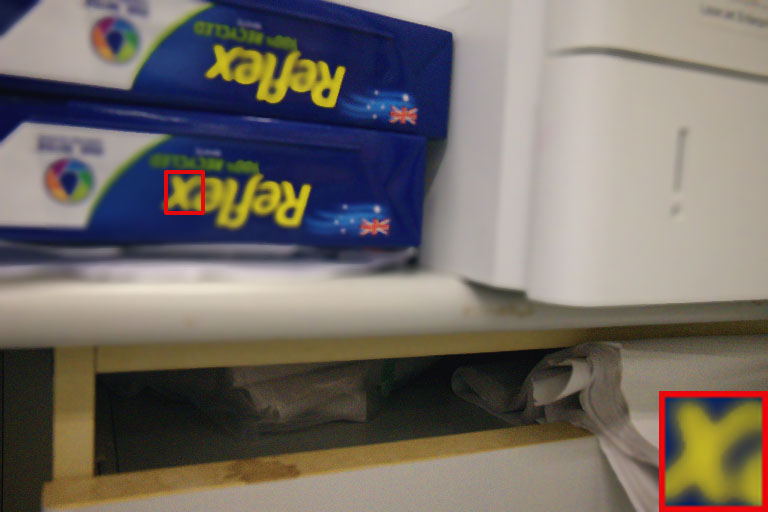}}\\
  & \scriptsize $\sigma$ = 5 & \scriptsize $\sigma$ = 10  & \scriptsize $\sigma$ = 15 \\
     \raisebox{-.5\height}{\includegraphics[width=0.24\linewidth]{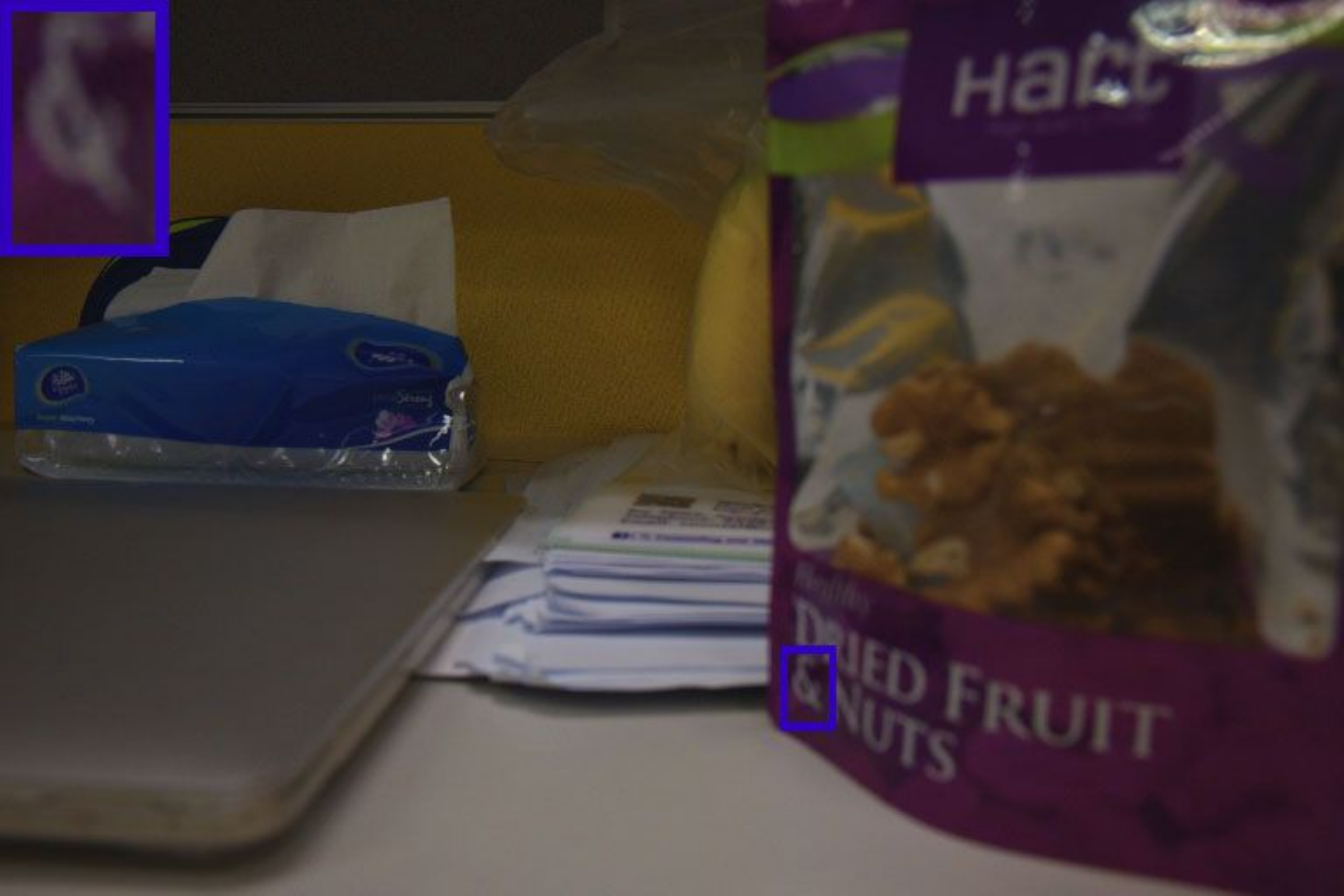}}       &    \raisebox{-.5\height}{\includegraphics[width=0.24\linewidth]{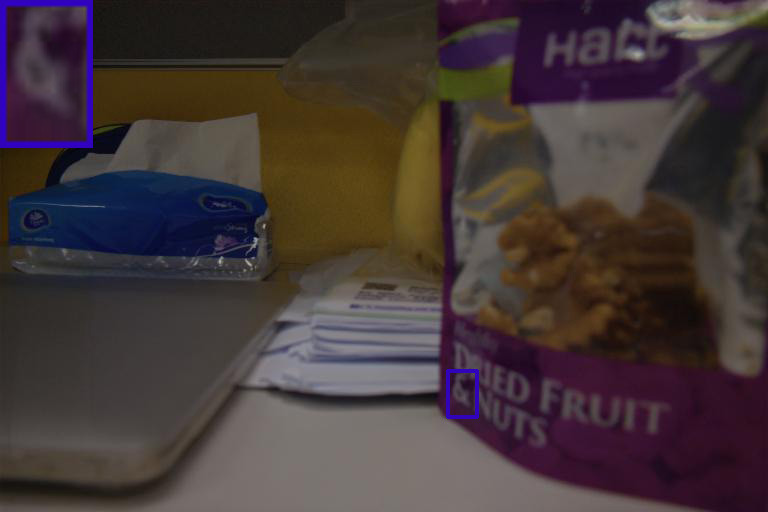}}  & \raisebox{-.5\height}{\includegraphics[width=0.24\linewidth]{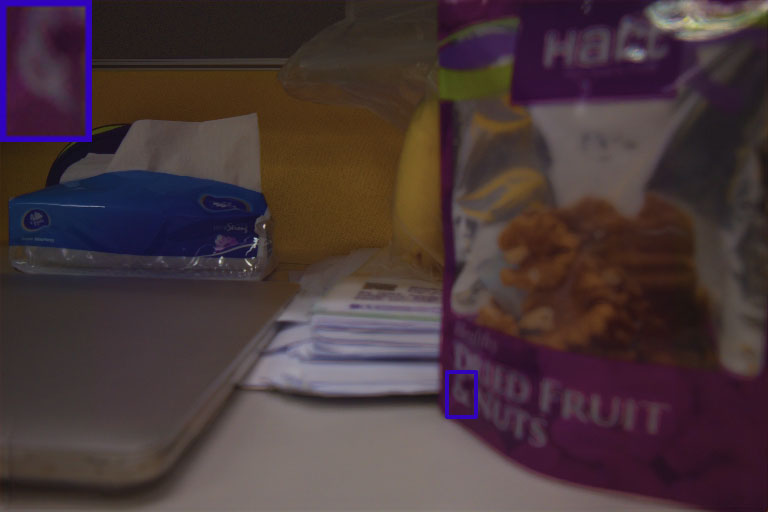}} &    \raisebox{-.5\height}{\includegraphics[width=0.24\linewidth]{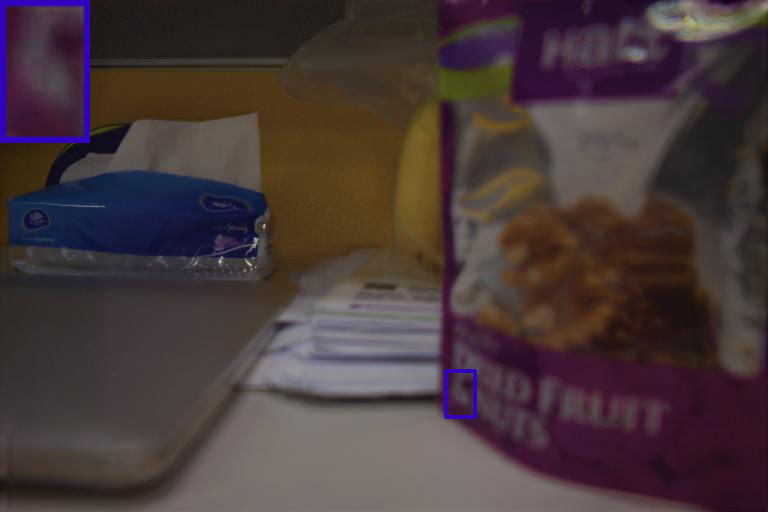}} \\
     \scriptsize f = 9.0 & \scriptsize f = 6.3 & \scriptsize f = 5.0  & \scriptsize f = 4.0 \\
     &
     \raisebox{-.5\height}{\includegraphics[width=0.24\linewidth]{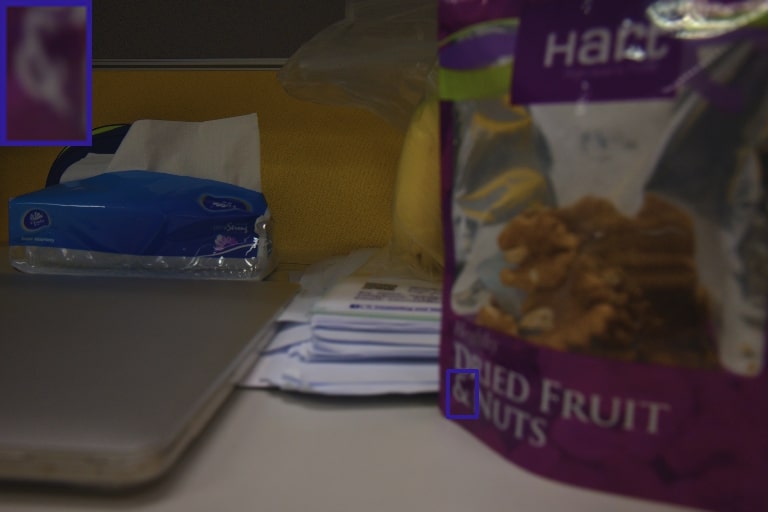}}  & \raisebox{-.5\height}{\includegraphics[width=0.24\linewidth]{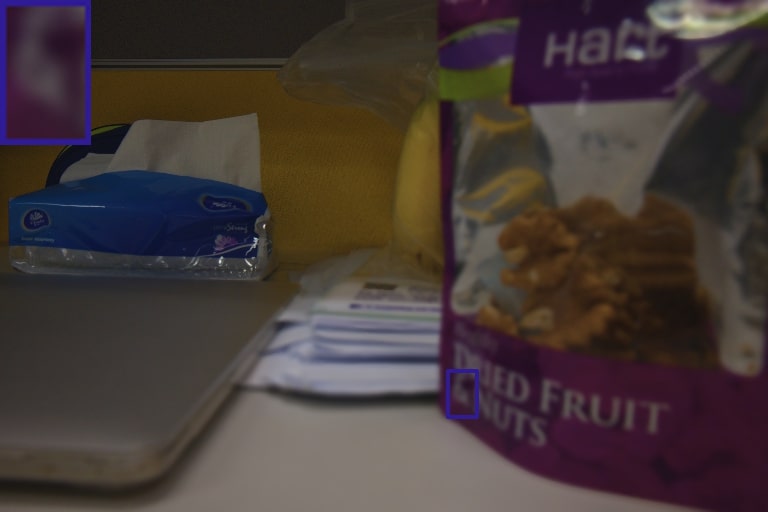}} &    \raisebox{-.5\height}{\includegraphics[width=0.24\linewidth]{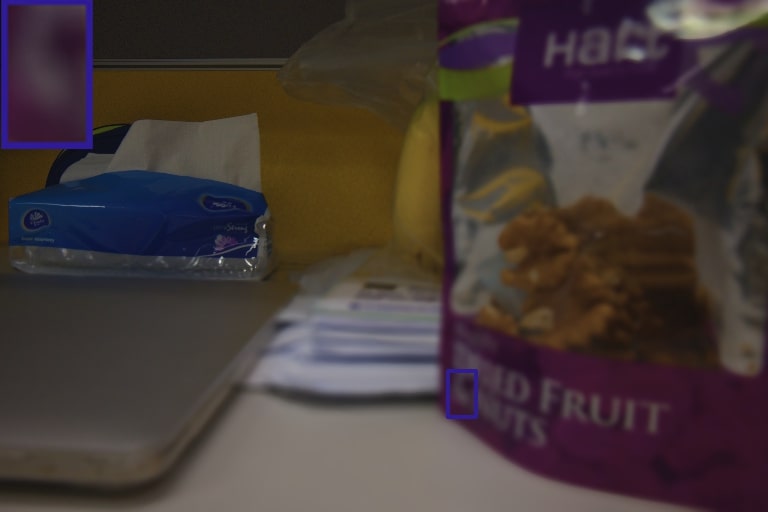}} \\
      & \scriptsize $\sigma$ = 5 & \scriptsize $\sigma$ = 10  & \scriptsize $\sigma$ = 15
     
    \end{tabular}
    \caption{Results of defocus magnification. The first and third rows are generated by our proposed simulator under different aperture settings. The second and fourth rows are generated by \cite{park2017unified} with different Gaussian kernels $\sigma$. Zoom-in to view details.}
    \vspace{-2mm}
    \label{fig:aperture}
\end{figure}

\mypara{Large-aperture enhancement.} The most straightforward application of our proposed structure is to magnify the defocus blur. As shown in Fig.~\ref{fig:aperture}, our model can correctly enhance the blurry region given a new aperture setting. Exposure is adjusted to keep the brightness unchanged. Compared with results from~\cite{park2017unified}, our model generates blurry content that is more consistent with the input image.




\begin{figure}[!t]
    \centering 
    \begin{tabular}{@{}c@{\hspace{0.01mm}}c@{\hspace{0.5mm}}c@{\hspace{0.5mm}}c@{\hspace{0.5mm}}c@{}}
    \rotatebox[origin=c]{90}{\scriptsize Input} &
     \raisebox{-.5\height}{\includegraphics[width=0.22\linewidth]{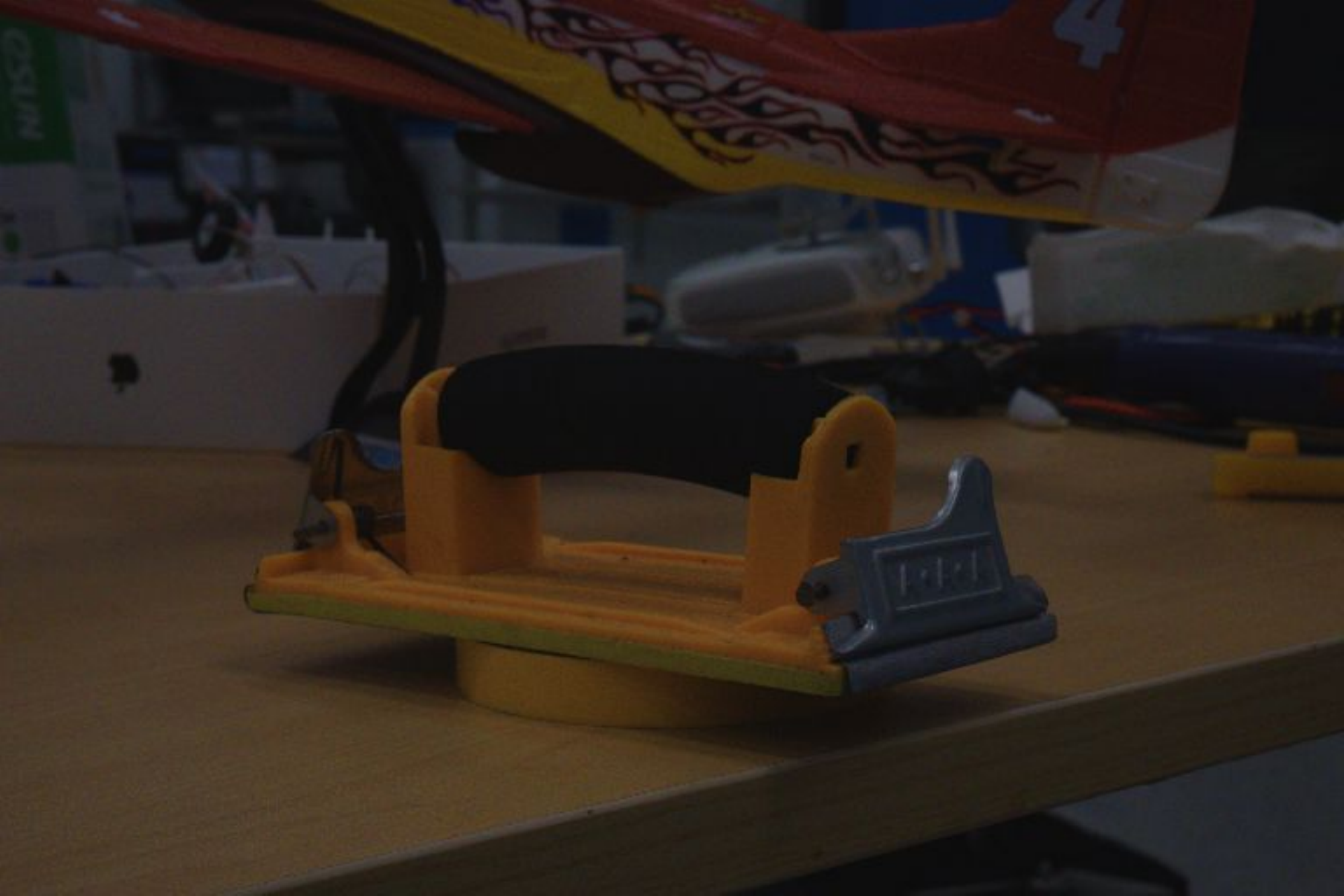}} &    \raisebox{-.5\height}{\includegraphics[width=0.22\linewidth]{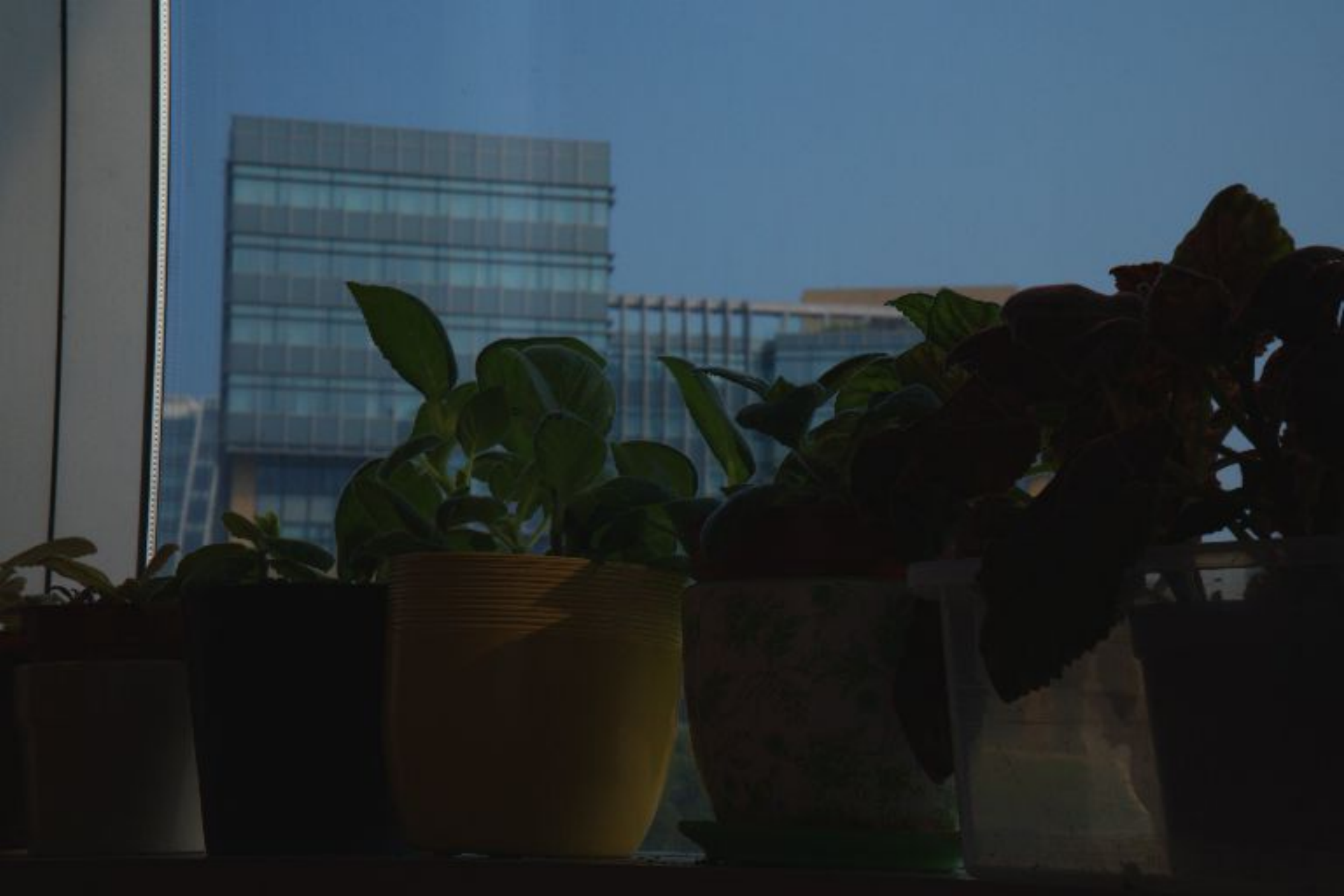}}  & \raisebox{-.5\height}{\includegraphics[width=0.22\linewidth]{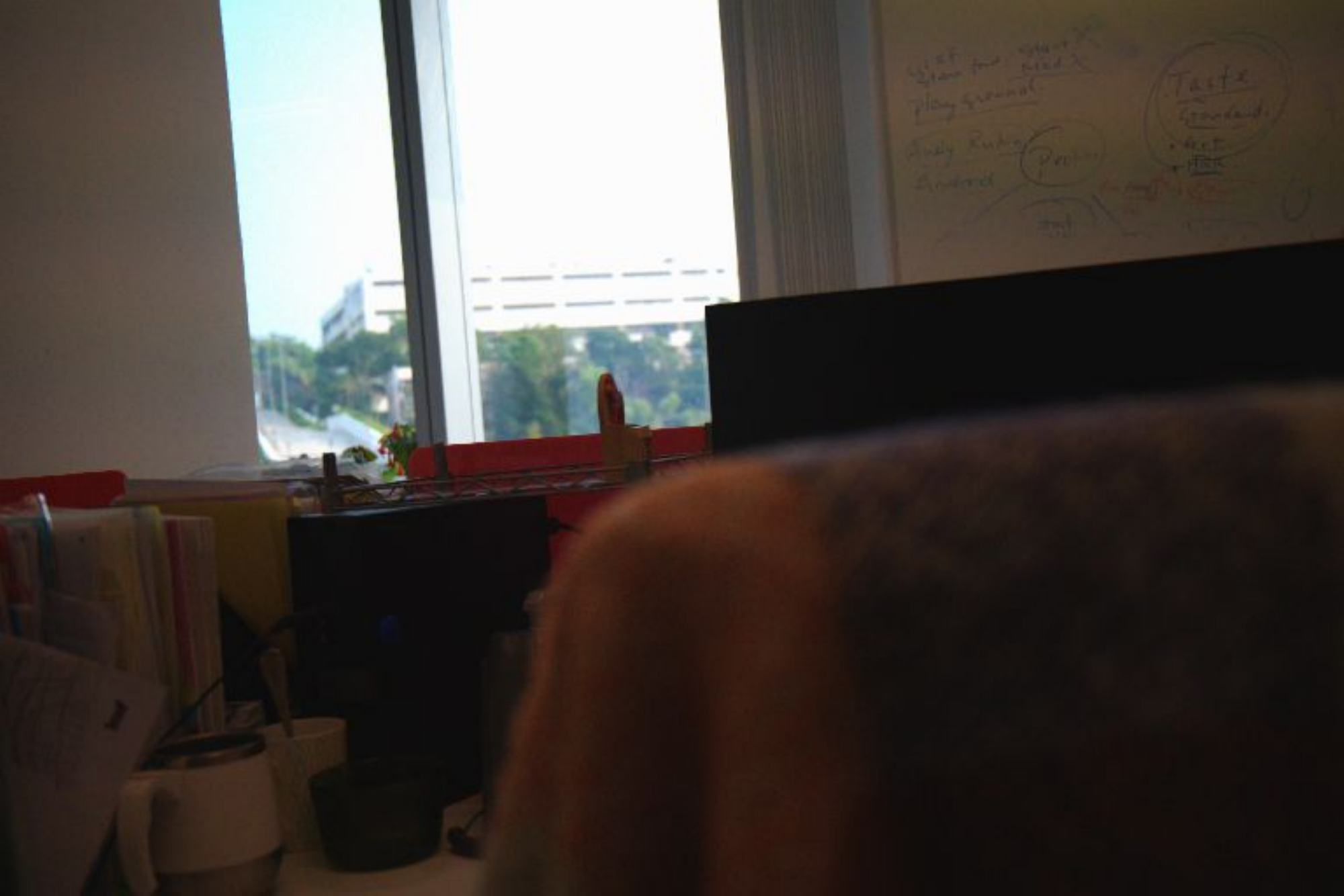}} &    \raisebox{-.5\height}{\includegraphics[width=0.22\linewidth]{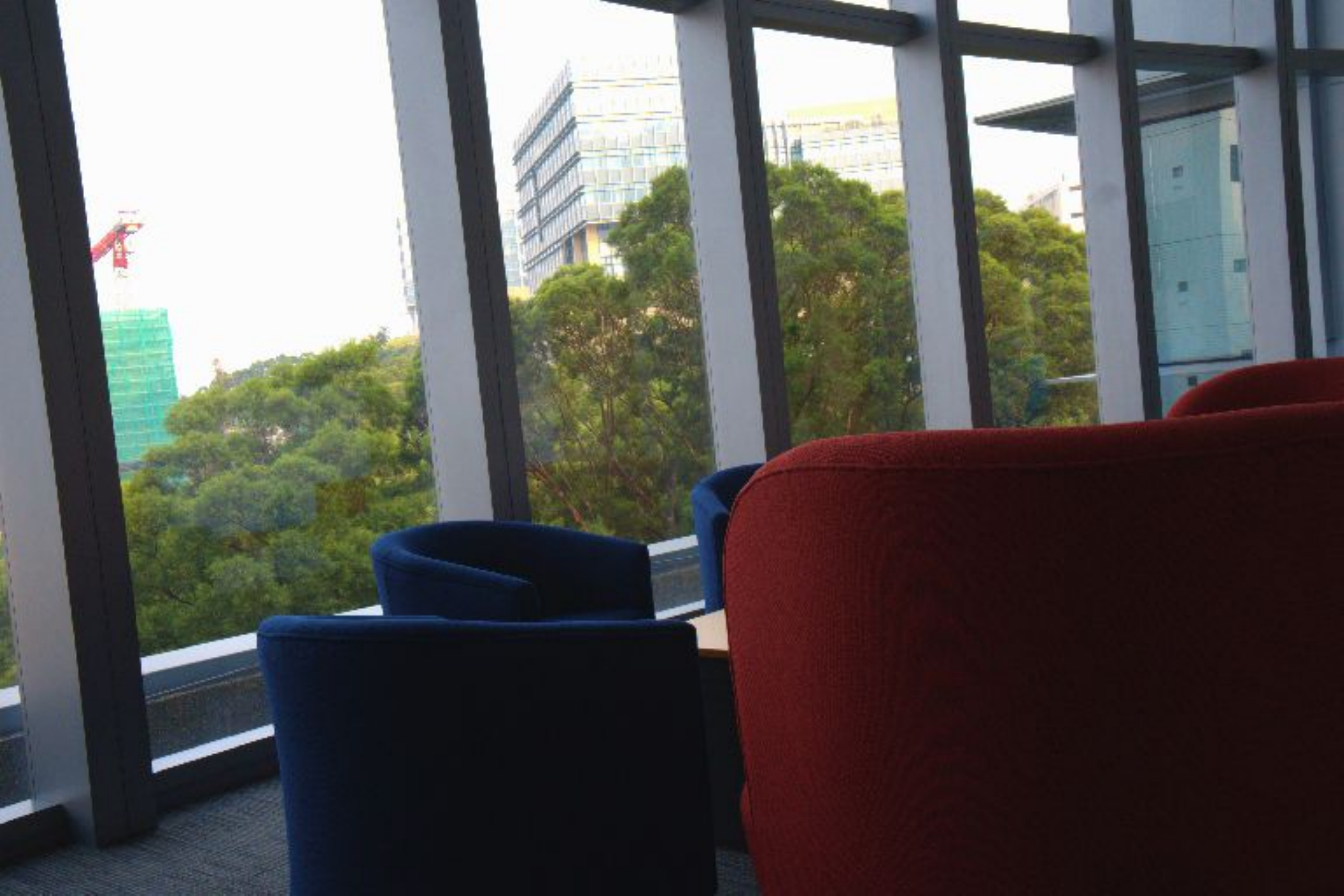}}  \vspace{0.5mm}    \\
    \rotatebox[origin=c]{90}{\scriptsize \makecell*[c]{Simulated\\ Image 1}} &
     \raisebox{-.5\height}{\includegraphics[width=0.22\linewidth]{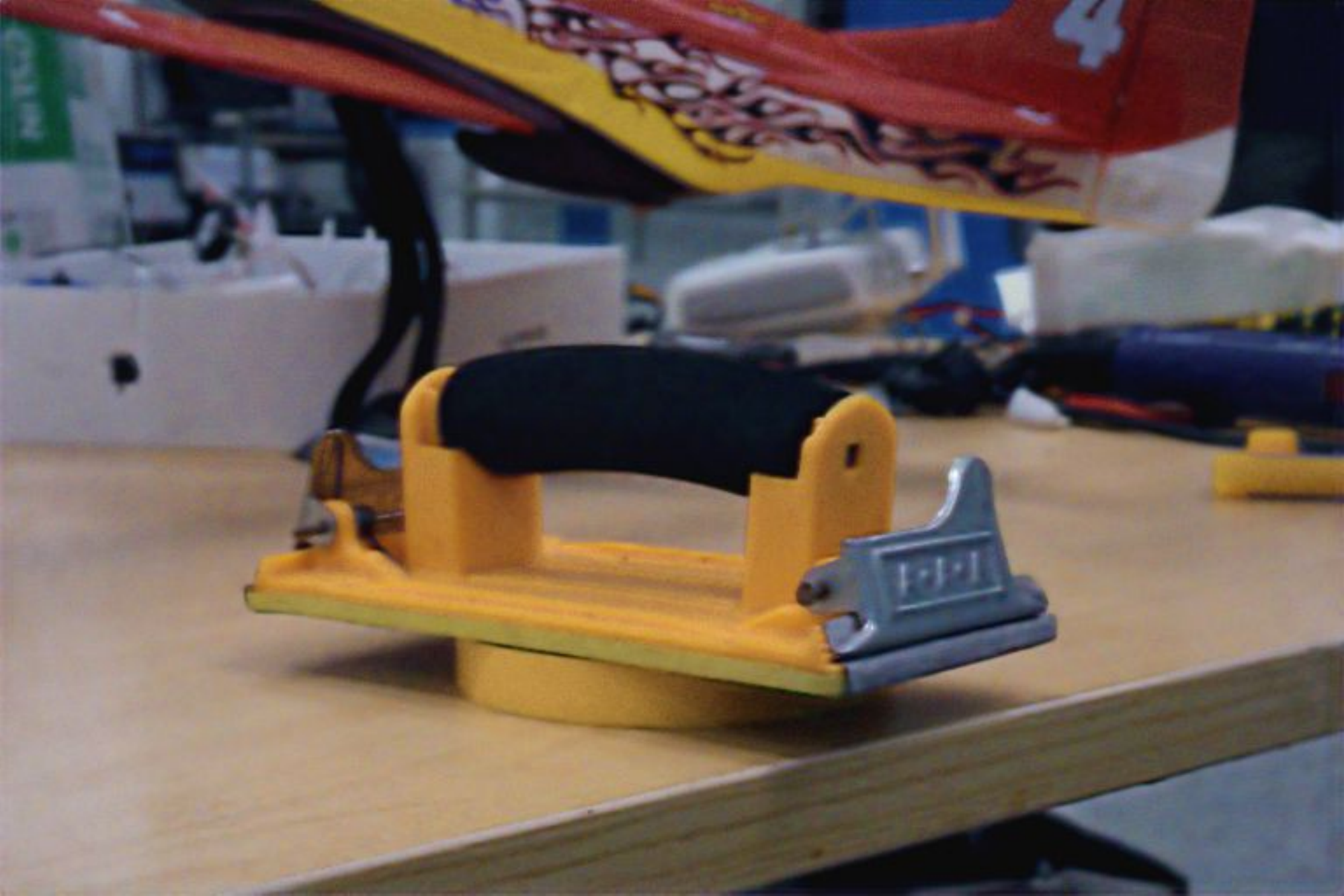}}       &    \raisebox{-.5\height}{\includegraphics[width=0.22\linewidth]{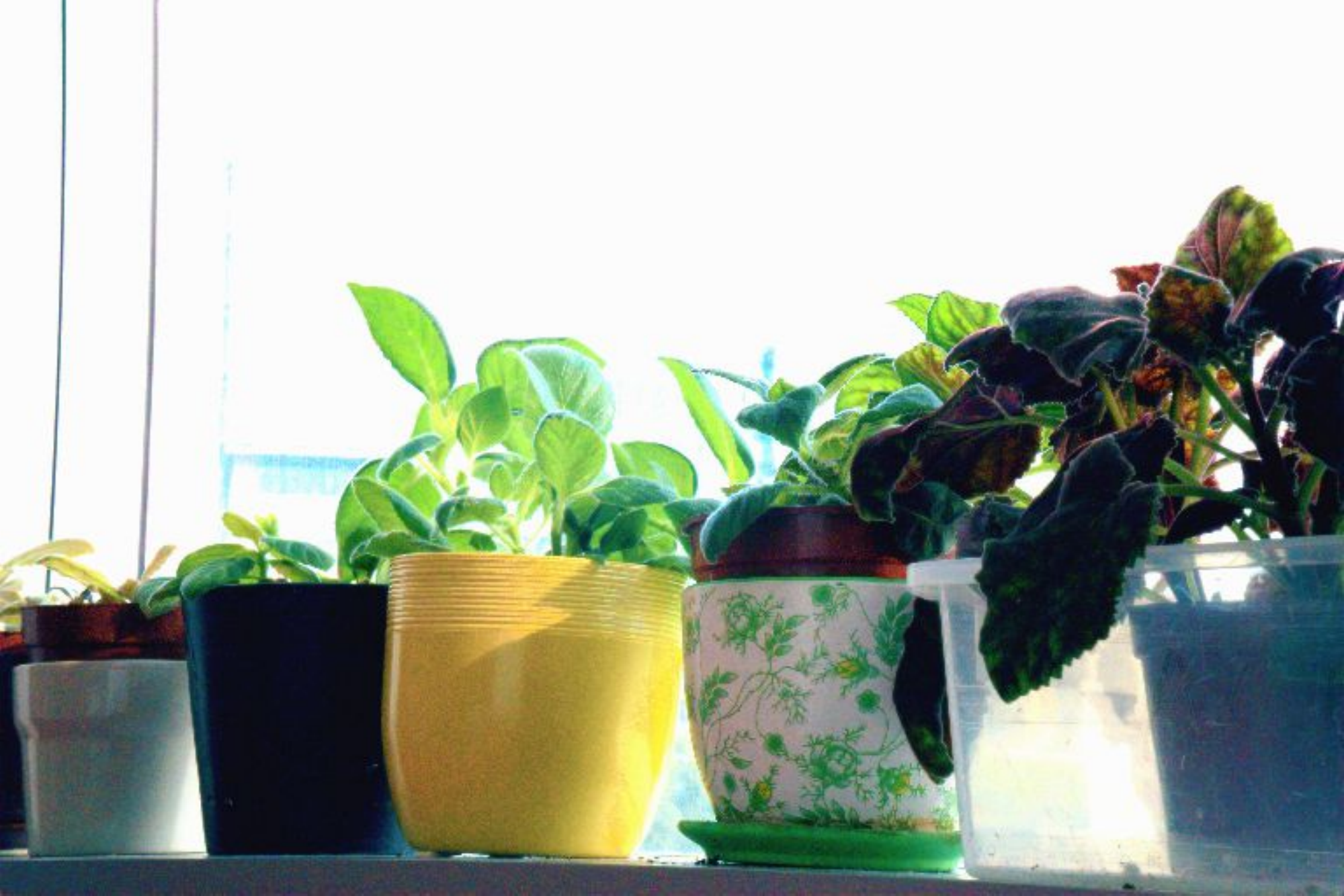}}  & \raisebox{-.5\height}{\includegraphics[width=0.22\linewidth]{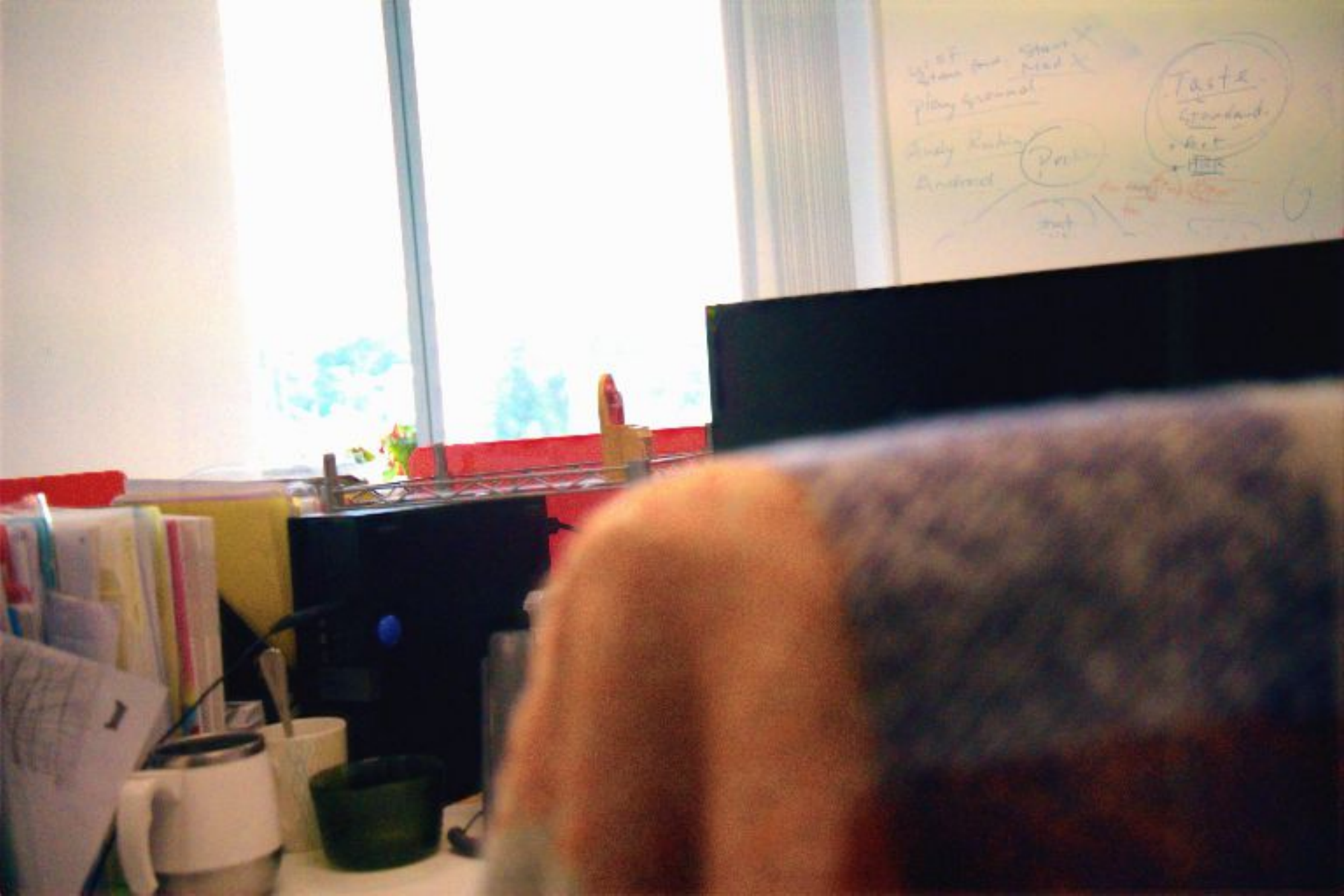}} &    \raisebox{-.5\height}{\includegraphics[width=0.22\linewidth]{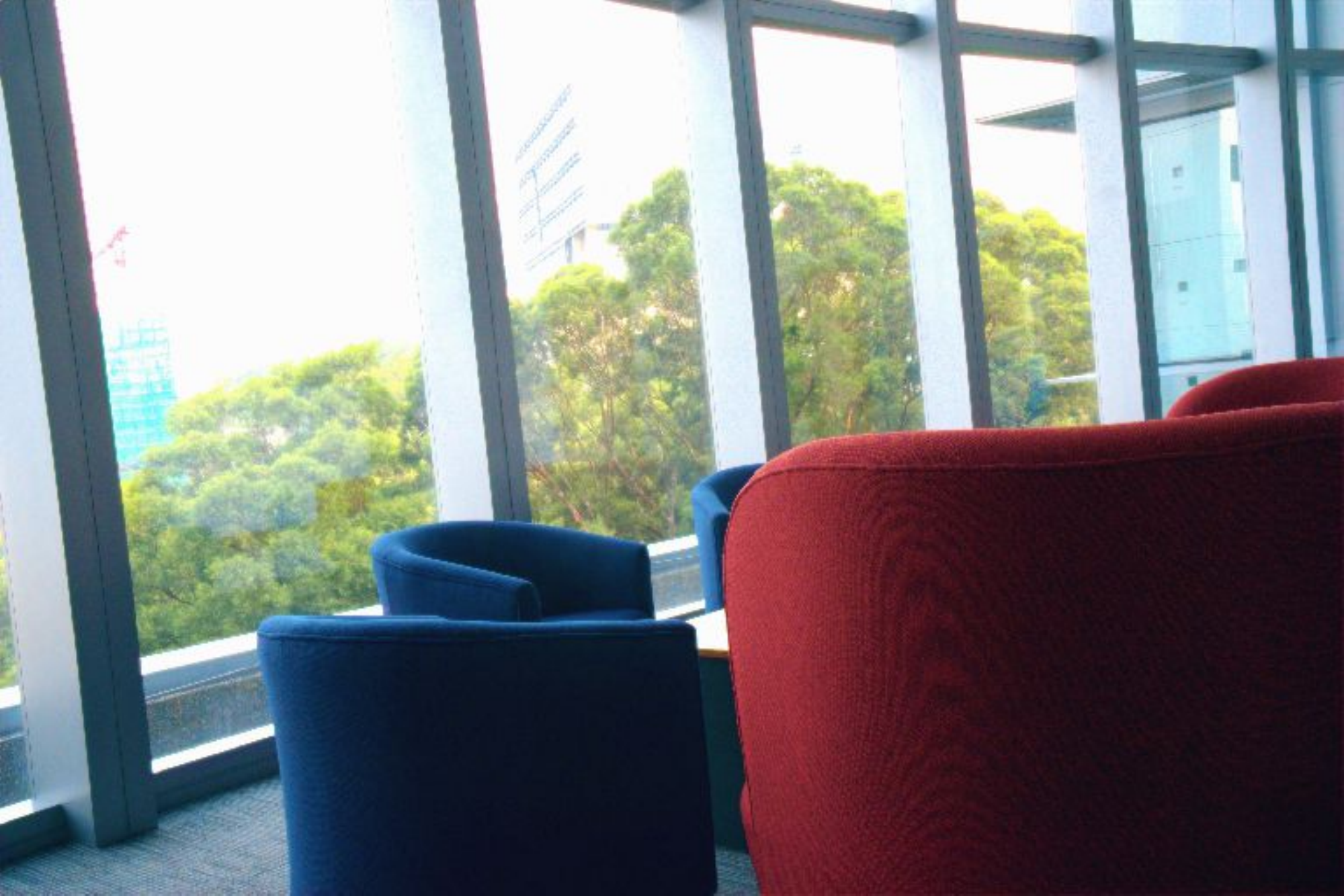}}   \vspace{0.5mm}      \\
     \rotatebox[origin=c]{90}{\scriptsize  \makecell*[c]{Simulated\\ Image 2}} &
     \raisebox{-.5\height}{\includegraphics[width=0.22\linewidth]{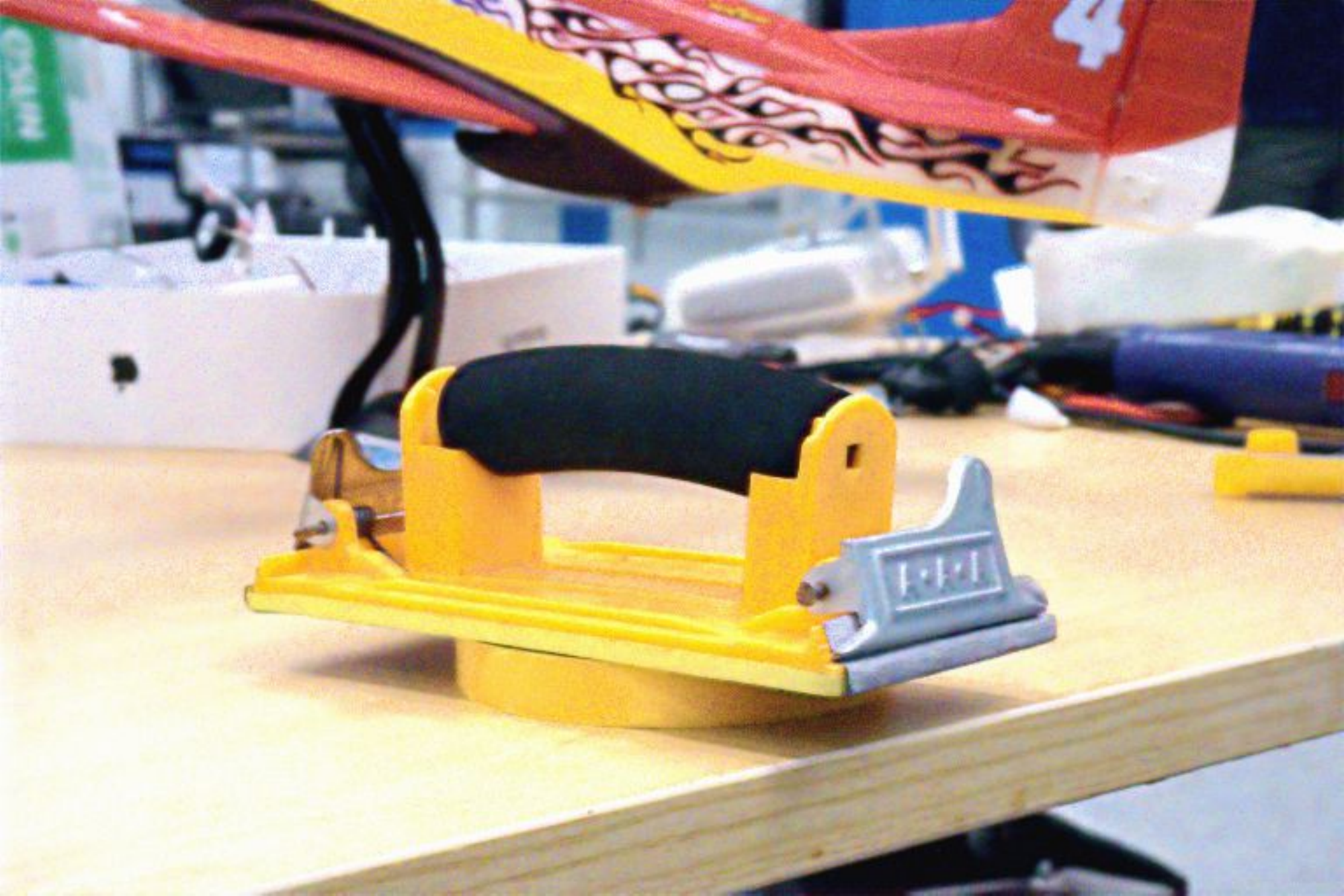}}       &    \raisebox{-.5\height}{\includegraphics[width=0.22\linewidth]{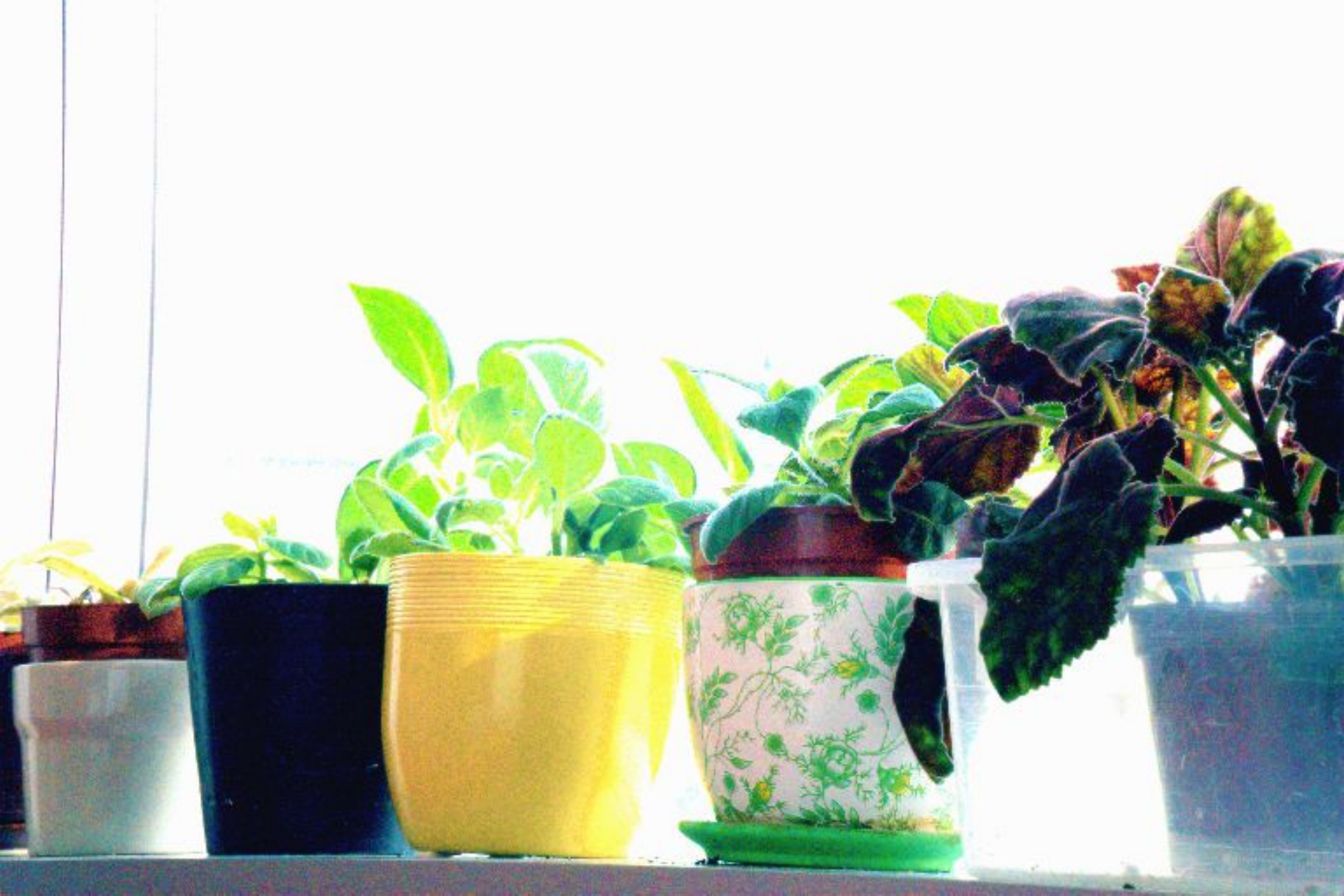}}  & \raisebox{-.5\height}{\includegraphics[width=0.22\linewidth]{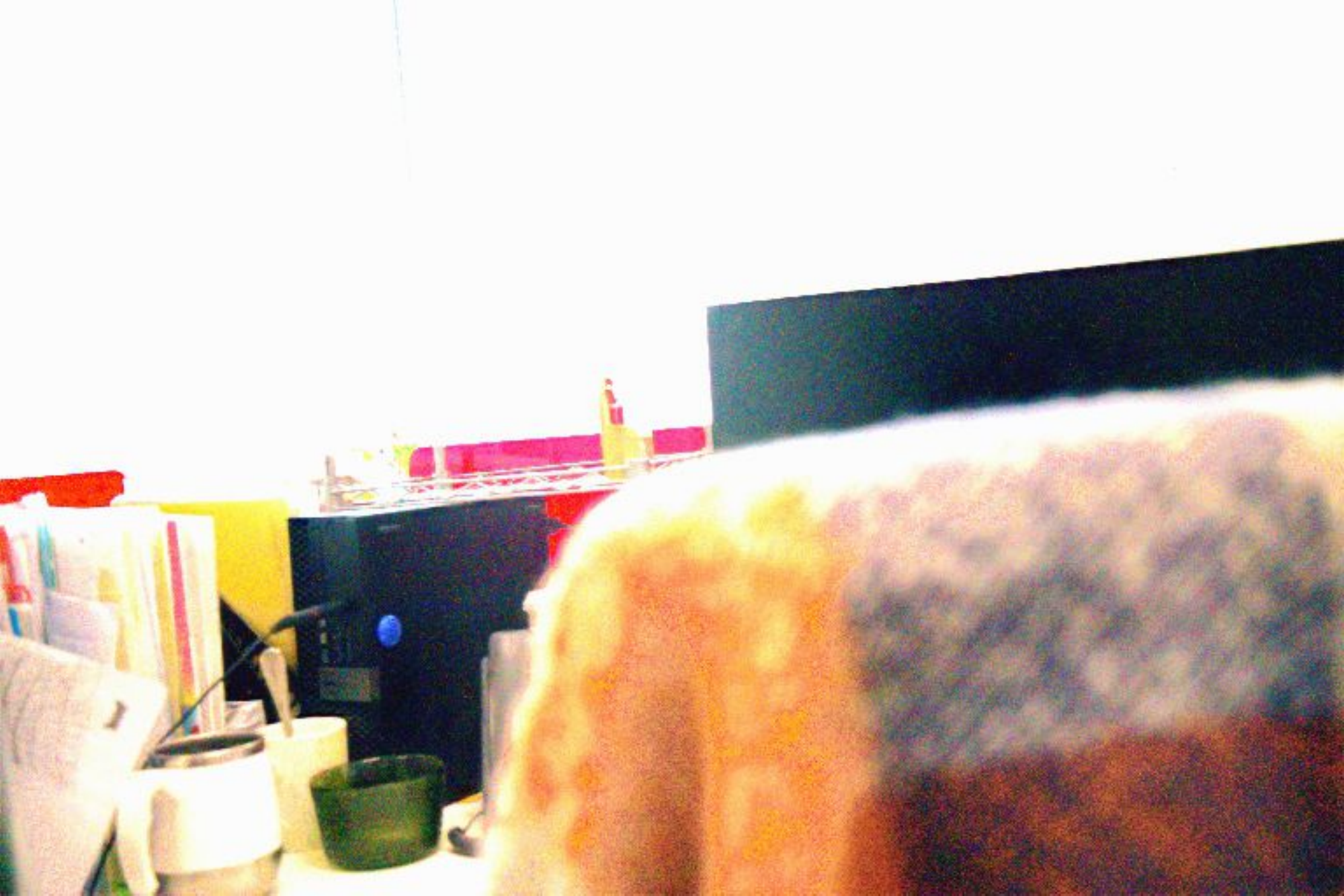}} &    \raisebox{-.5\height}{\includegraphics[width=0.22\linewidth]{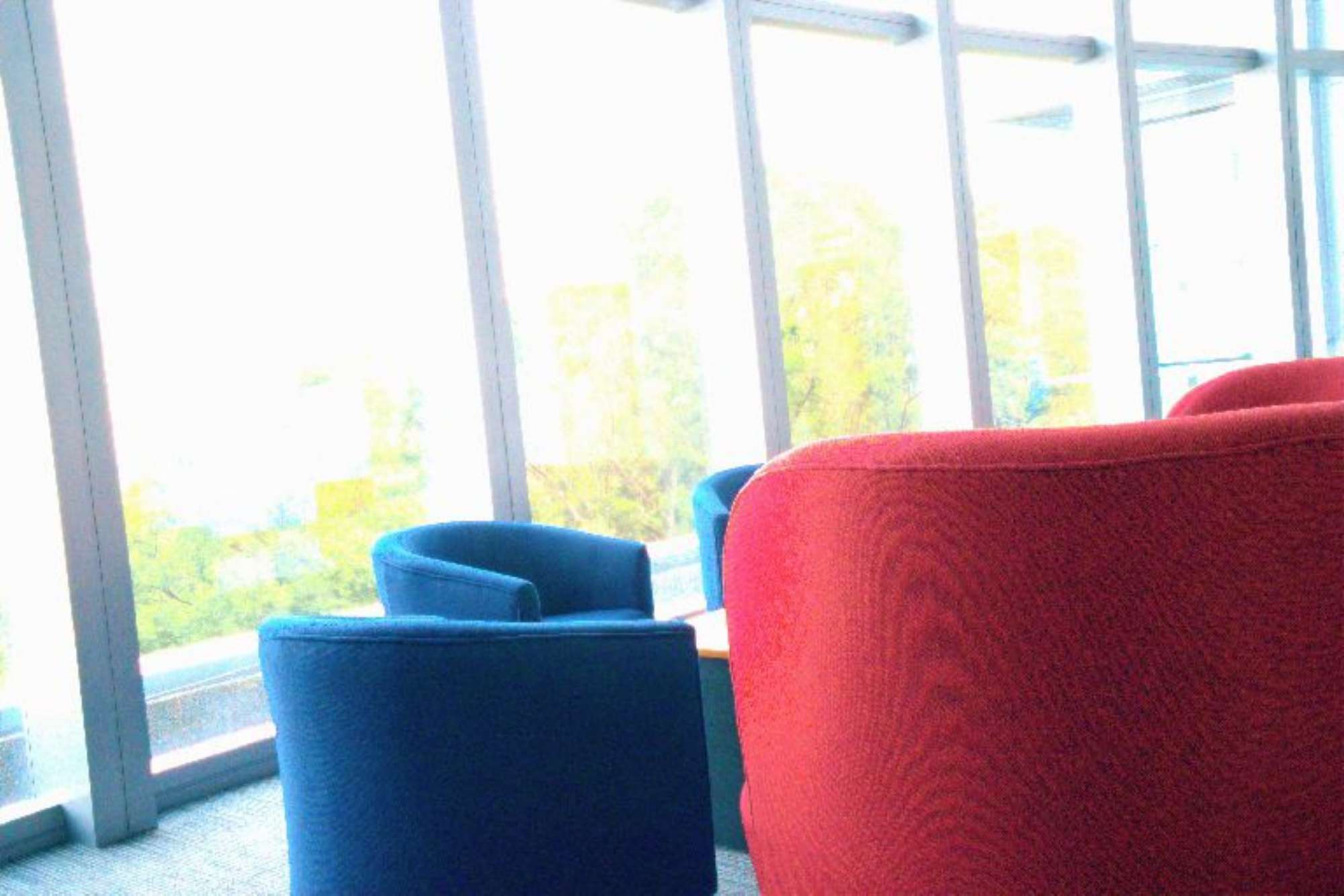}}    \vspace{0.5mm}     \\
     \rotatebox[origin=c]{90}{\scriptsize  \makecell*[c]{HDR\\ (Ours)}} &
     \raisebox{-.5\height}{\includegraphics[width=0.22\linewidth]{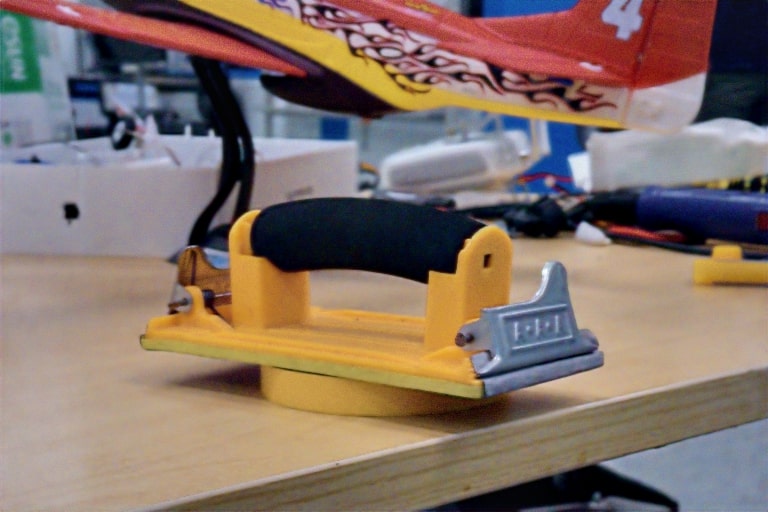}}       &    \raisebox{-.5\height}{\includegraphics[width=0.22\linewidth]{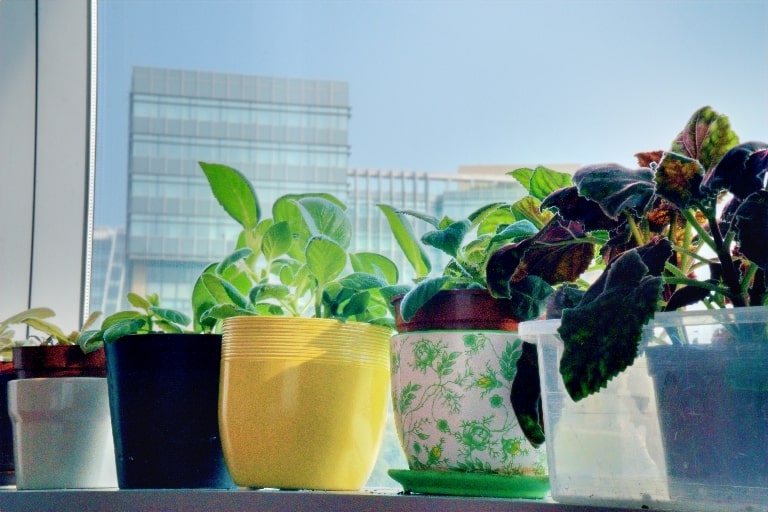}}  & \raisebox{-.5\height}{\includegraphics[width=0.22\linewidth]{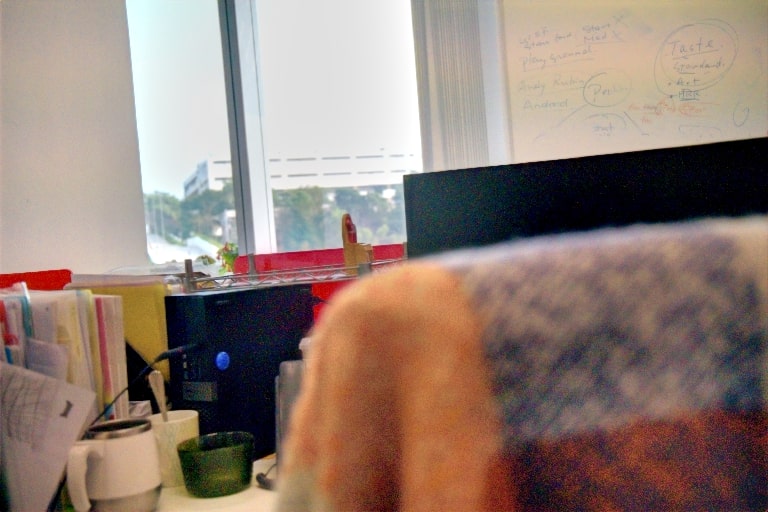}} &    \raisebox{-.5\height}{\includegraphics[width=0.22\linewidth]{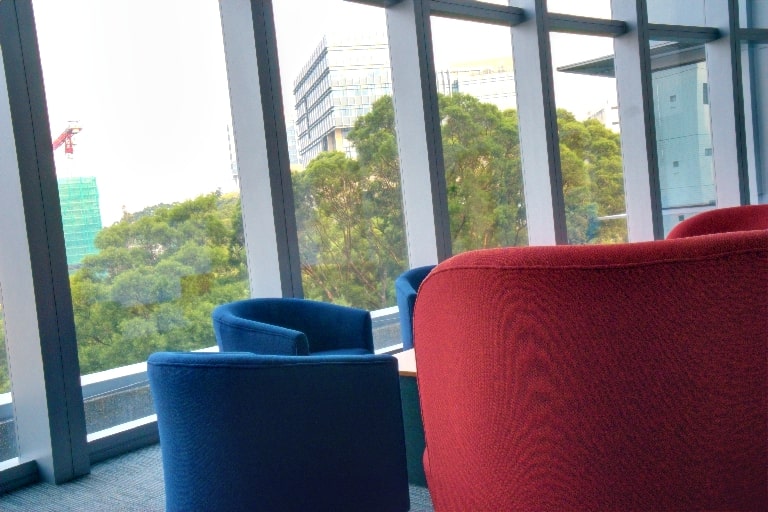}}   \vspace{0.5mm}      \\
     \rotatebox[origin=c]{90}{\scriptsize  \makecell*[c]{Enlighten-\\GAN}} &
     \raisebox{-.5\height}{\includegraphics[width=0.22\linewidth]{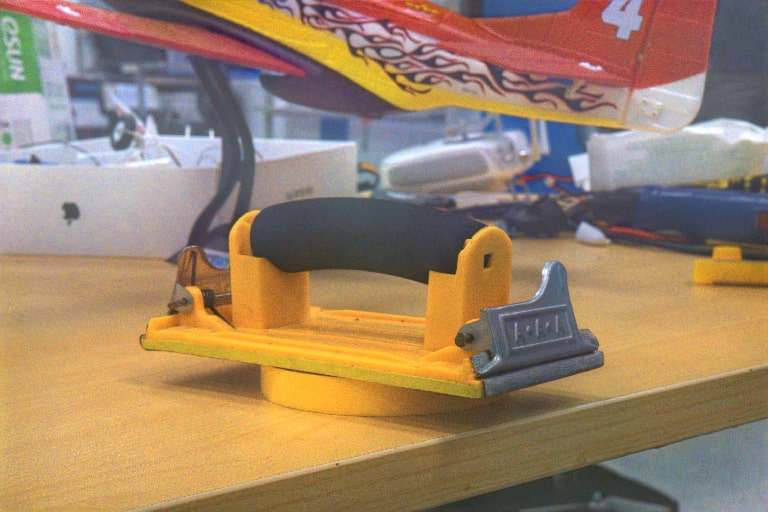}}       &    \raisebox{-.5\height}{\includegraphics[width=0.22\linewidth]{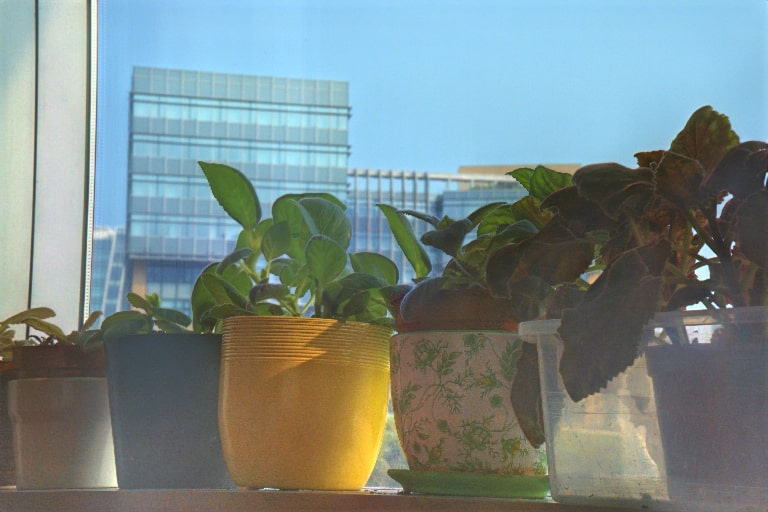}}  & \raisebox{-.5\height}{\includegraphics[width=0.22\linewidth]{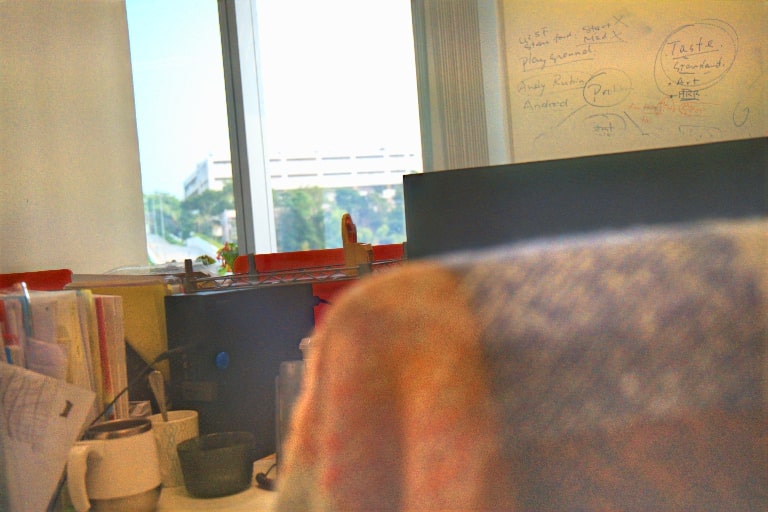}} &    \raisebox{-.5\height}{\includegraphics[width=0.22\linewidth]{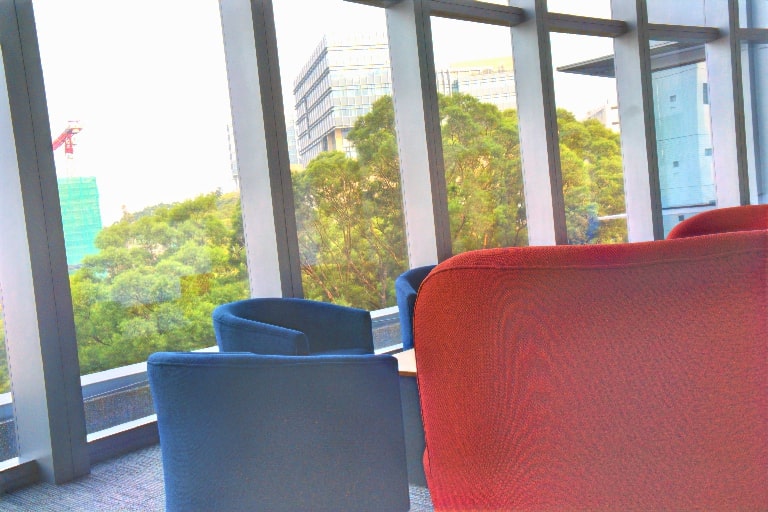}}   \vspace{0.5mm}     \\ 
     \rotatebox[origin=c]{90}{\scriptsize  \makecell*[c]{Zero-DCE}} &
     \raisebox{-.5\height}{\includegraphics[width=0.22\linewidth]{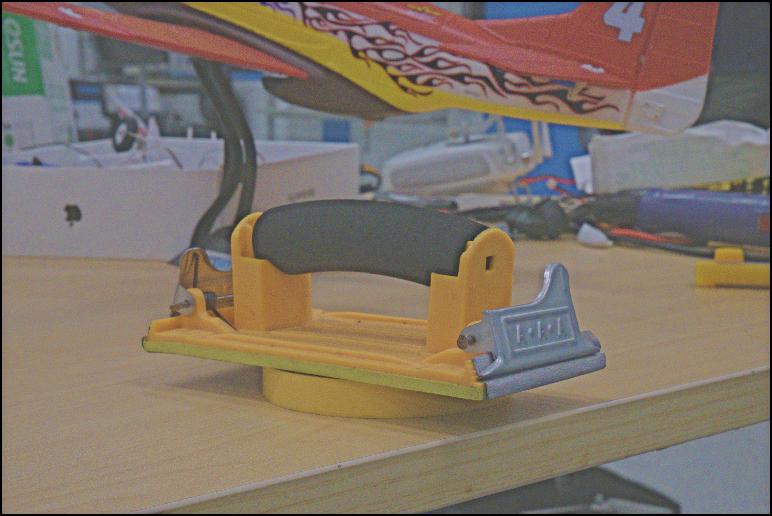}}       &    \raisebox{-.5\height}{\includegraphics[width=0.22\linewidth]{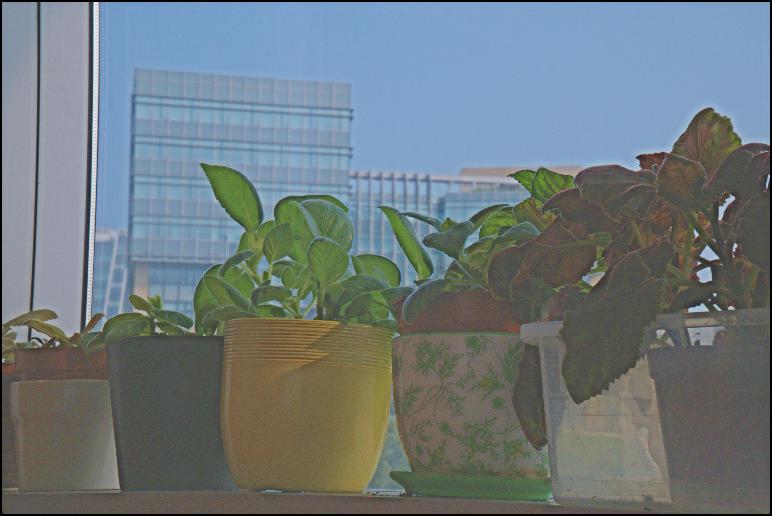}}  & \raisebox{-.5\height}{\includegraphics[width=0.22\linewidth]{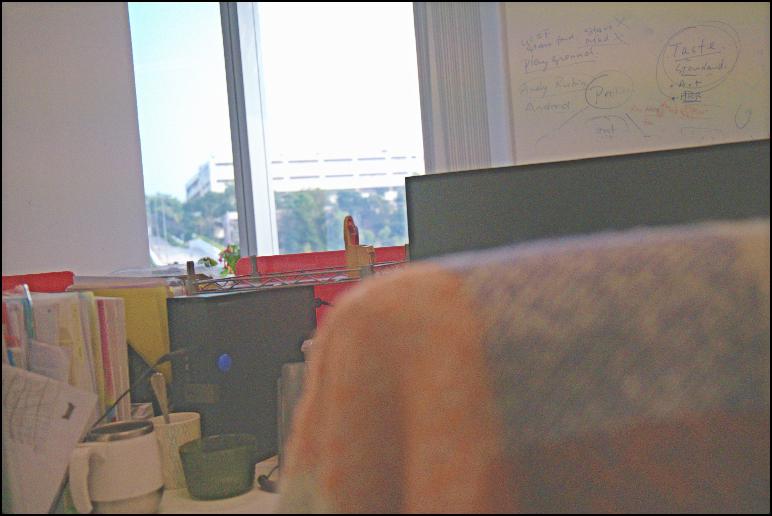}} &    \raisebox{-.5\height}{\includegraphics[width=0.22\linewidth]{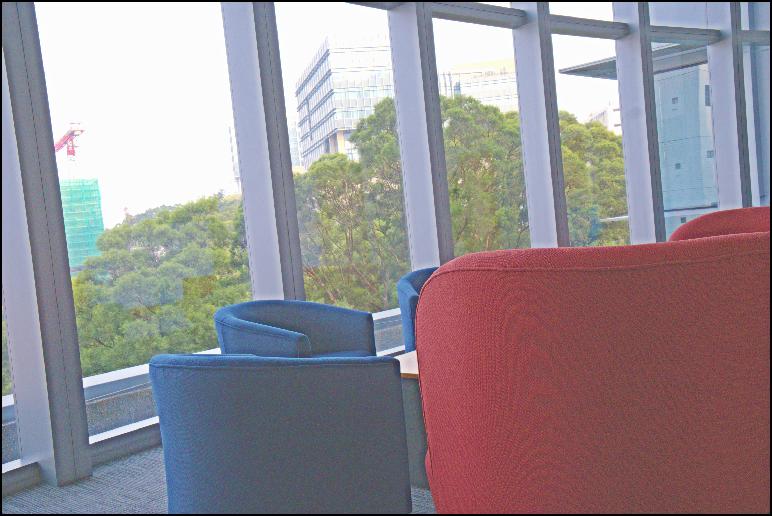}}     \vspace{0.5mm}   \\
     \vspace{-2mm}
     
    \end{tabular}
    \caption{Results of HDR image generation. We generate HDR images by~\cite{lee2018hdr}. We pick two representative simulated images for display. Results from EnlightenGAN~\cite{jiang2019enlightengan} and Zero-DCE~\cite{ZeroDCE} are listed in the last two rows. Zoom-in to view details.}
    \label{fig:hdr}
\end{figure}

\mypara{HDR images.} As our simulator can generate images with other exposure settings, a very natural application is to generate HDR images. Our HDR pipeline is straightforward: we first simulate multiple output raw images from the input raw using a group of camera exposure settings.  We set ISO to be as small as possible for reducing the image noise and calculate the exposure time by shifting original exposure in every 0.5$Ev$ stop in the range $[-4Ev, 4Ev]$. The output aperture keeps unchanged. We then convert raw data to sRGB-space images and fuse them using the approach proposed by~\cite{lee2018hdr} to generate HDR output. In this pipeline, we can fully exploit the information at each exposure level and reduce the noise introduced when brightening dark regions. As Fig.~\ref{fig:hdr} shows, our fused output brings back the missing details and appears more attractive than those generated by recent works~\cite{jiang2019enlightengan, ZeroDCE}.

\begin{figure}[!t]
    \centering 
    \begin{tabular}{@{}c@{\hspace{0.2mm}}c@{\hspace{0.2mm}}c@{\hspace{0.2mm}}c@{\hspace{0.2mm}}c@{}}
     \includegraphics[width=0.198\linewidth]{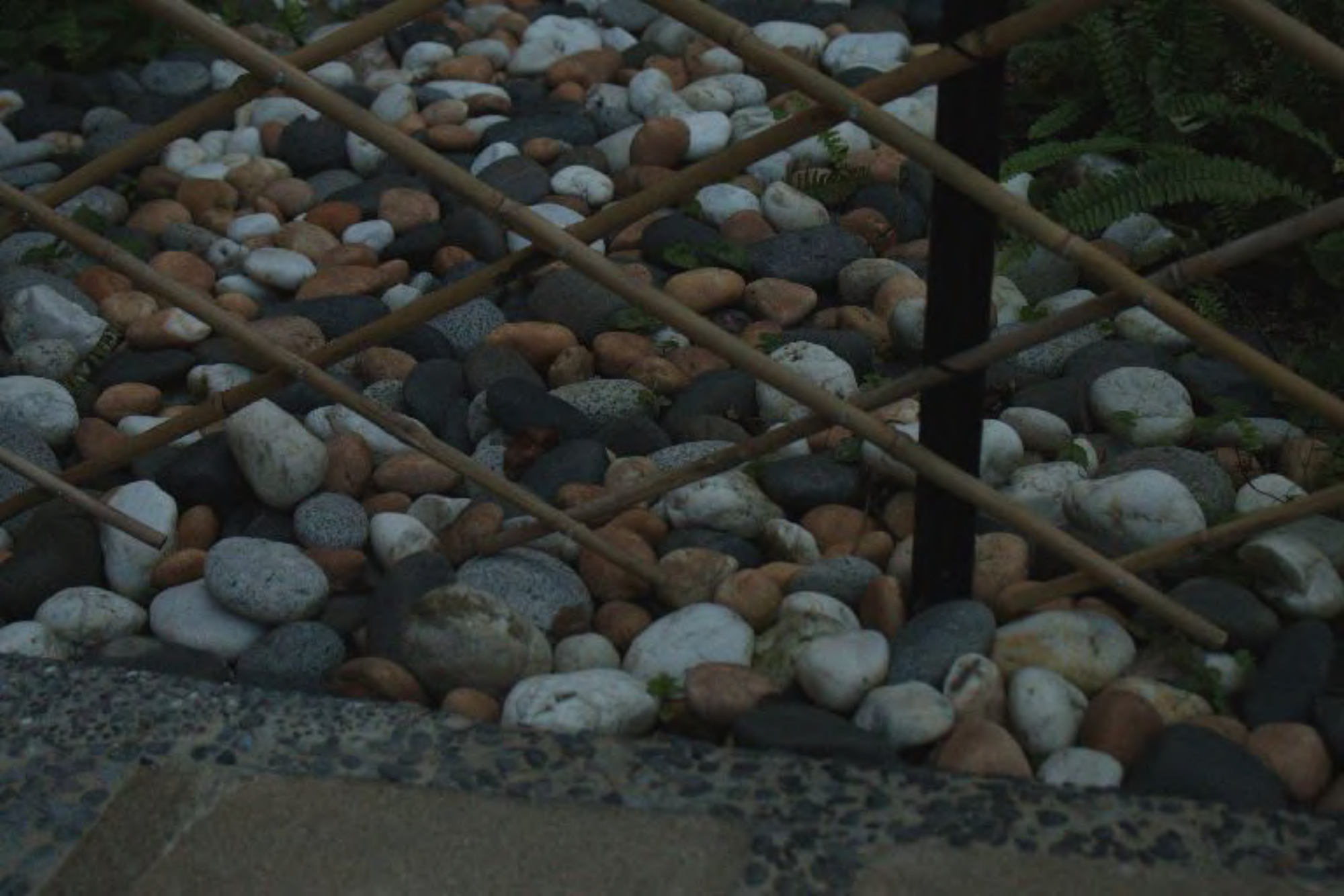} &    \includegraphics[width=0.198\linewidth]{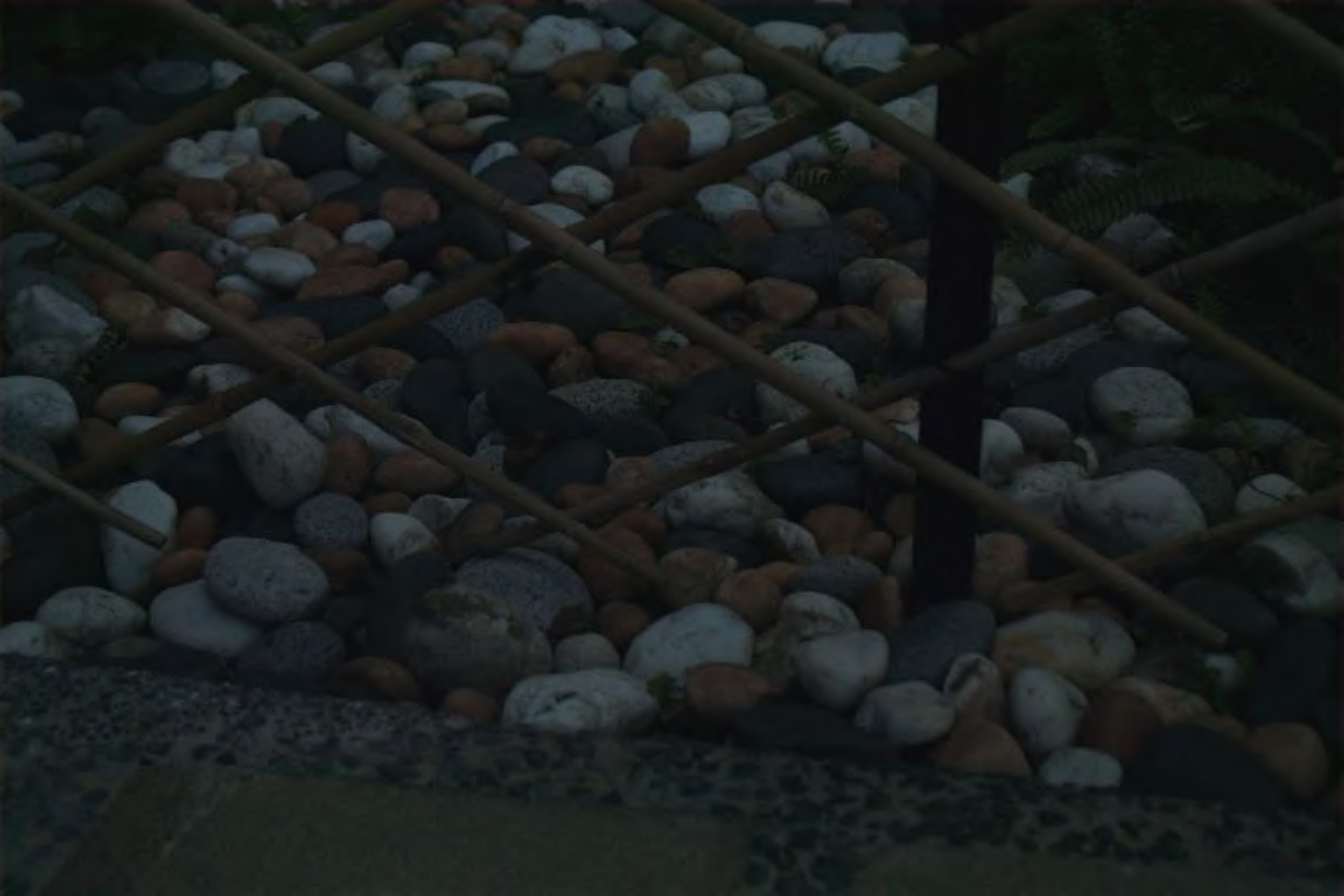} & \includegraphics[width=0.198\linewidth]{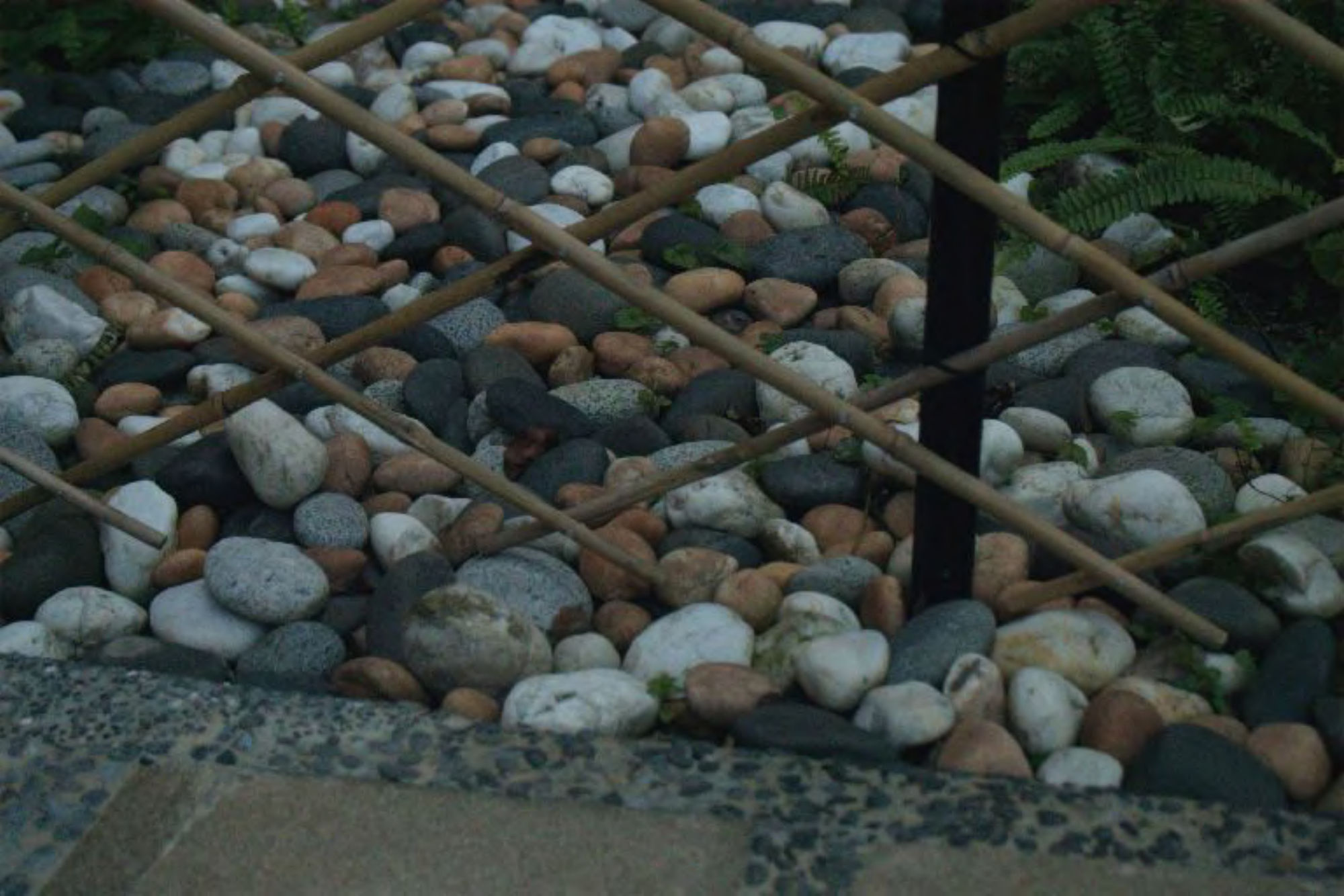} & 
     \includegraphics[width=0.198\linewidth]{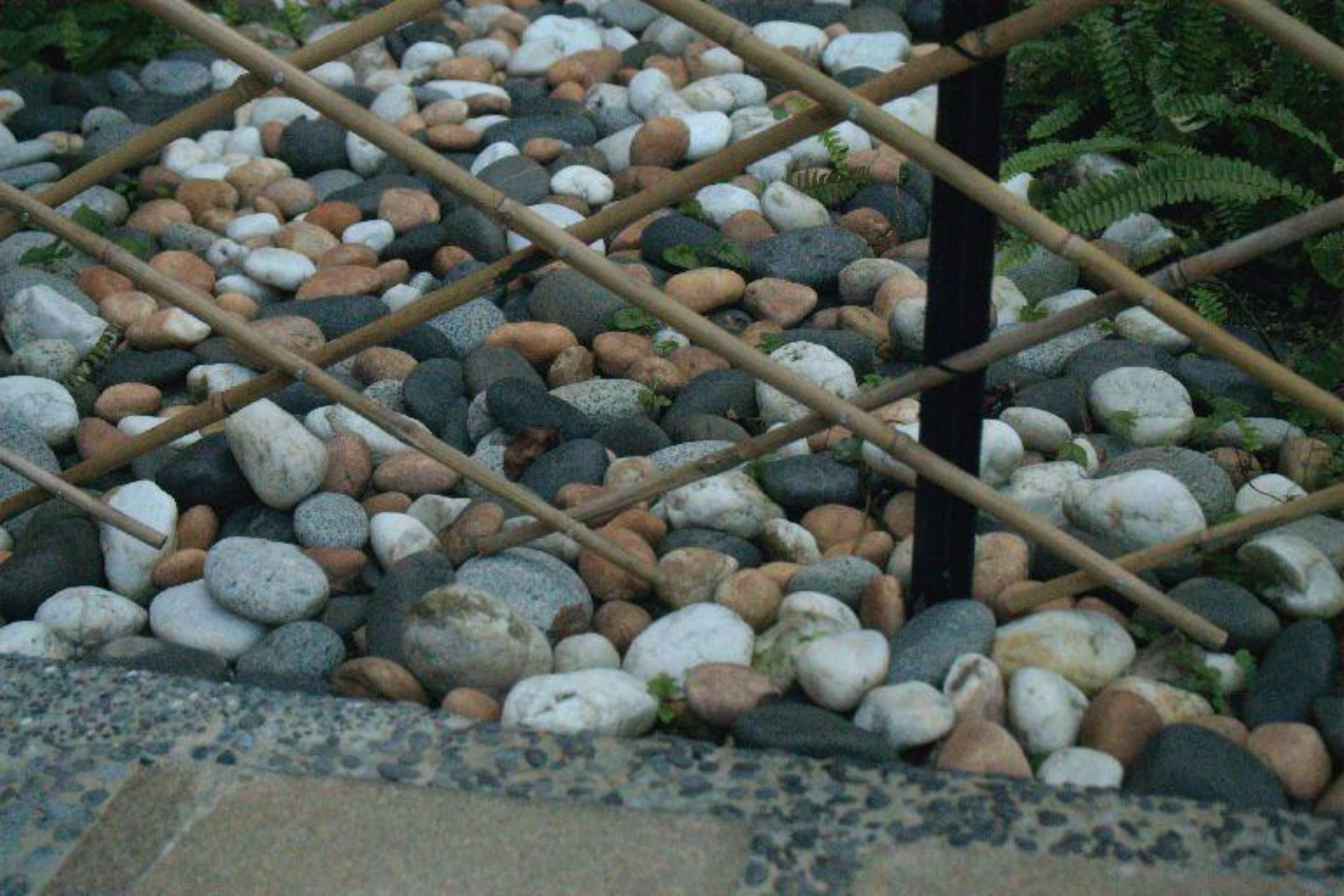} &
     \includegraphics[width=0.198\linewidth]{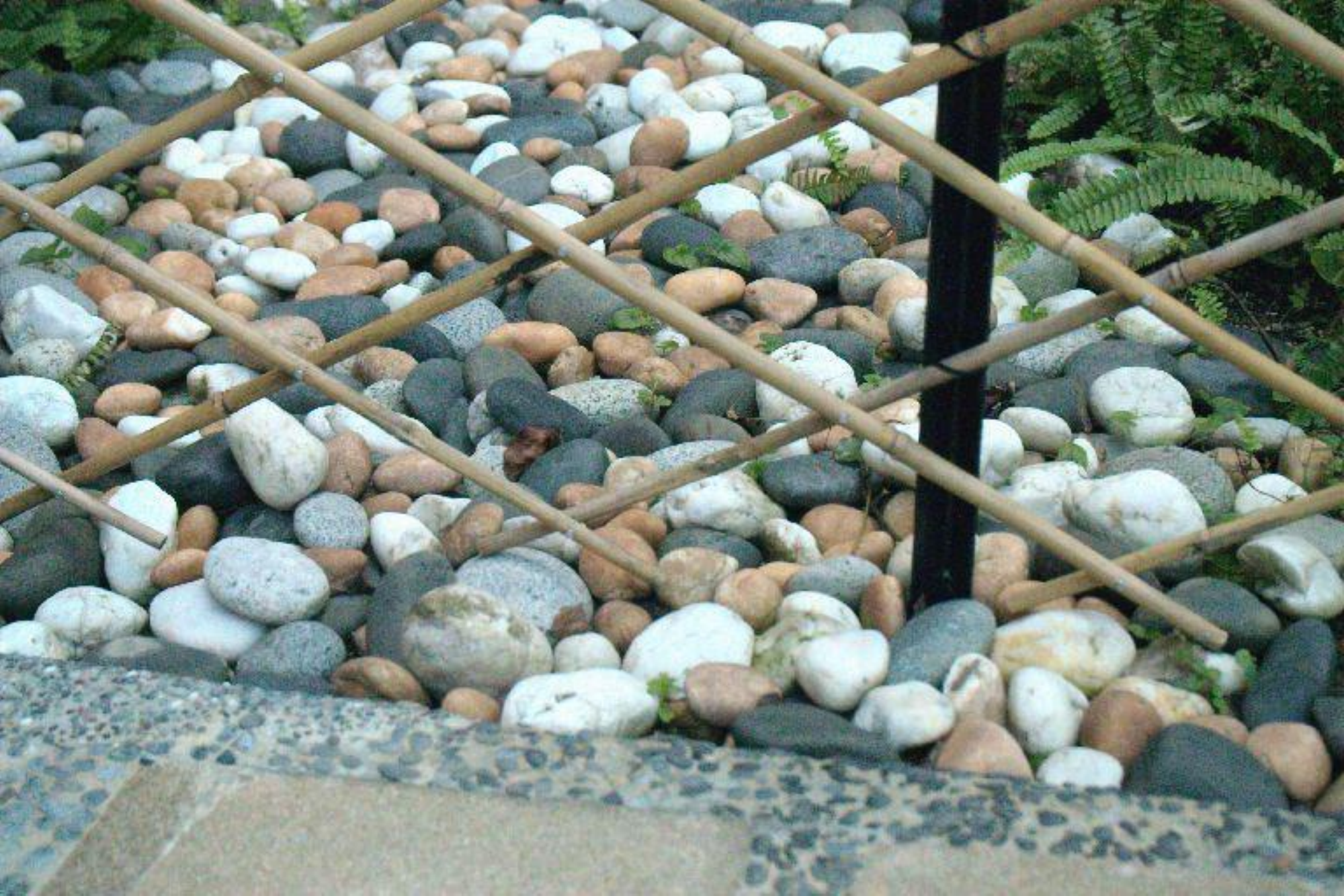} \vspace{-2mm}\\
   \scriptsize 0.627 (original)& \scriptsize 0.705  &\scriptsize 0.428&\scriptsize 0.222 &\scriptsize \textbf{0.200}\\
     \includegraphics[width=0.198\linewidth]{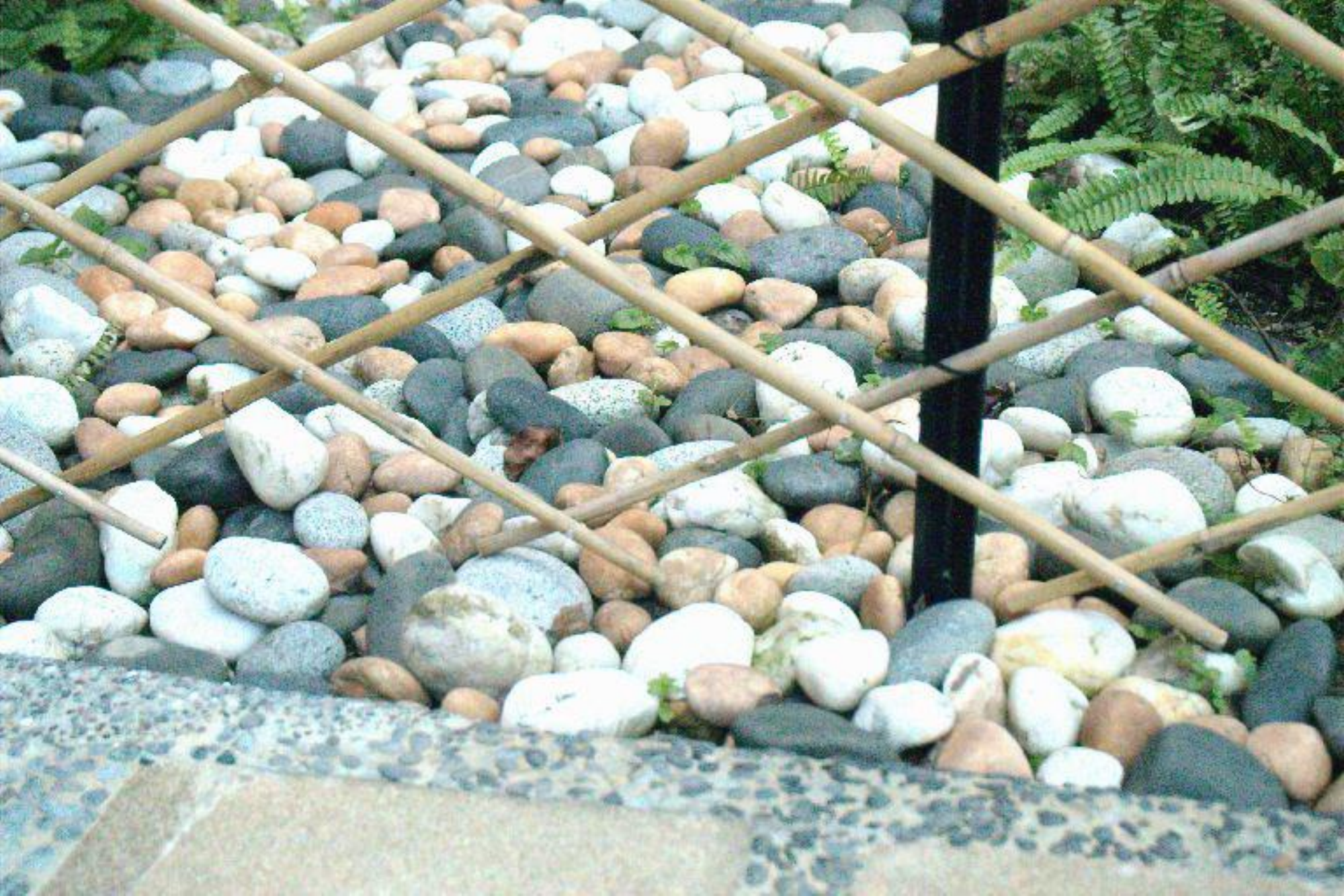} &    \includegraphics[width=0.195\linewidth]{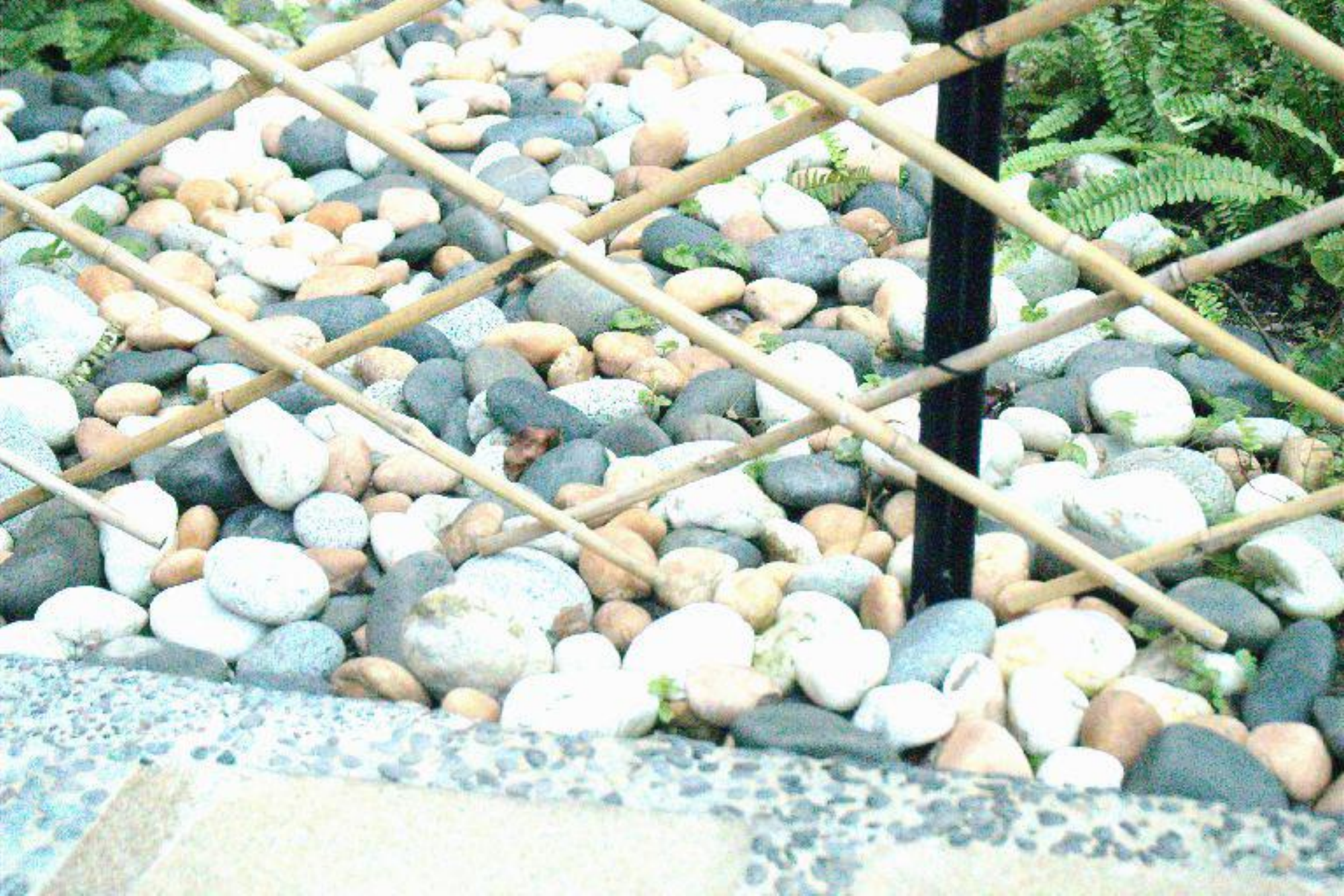} & \includegraphics[width=0.198\linewidth]{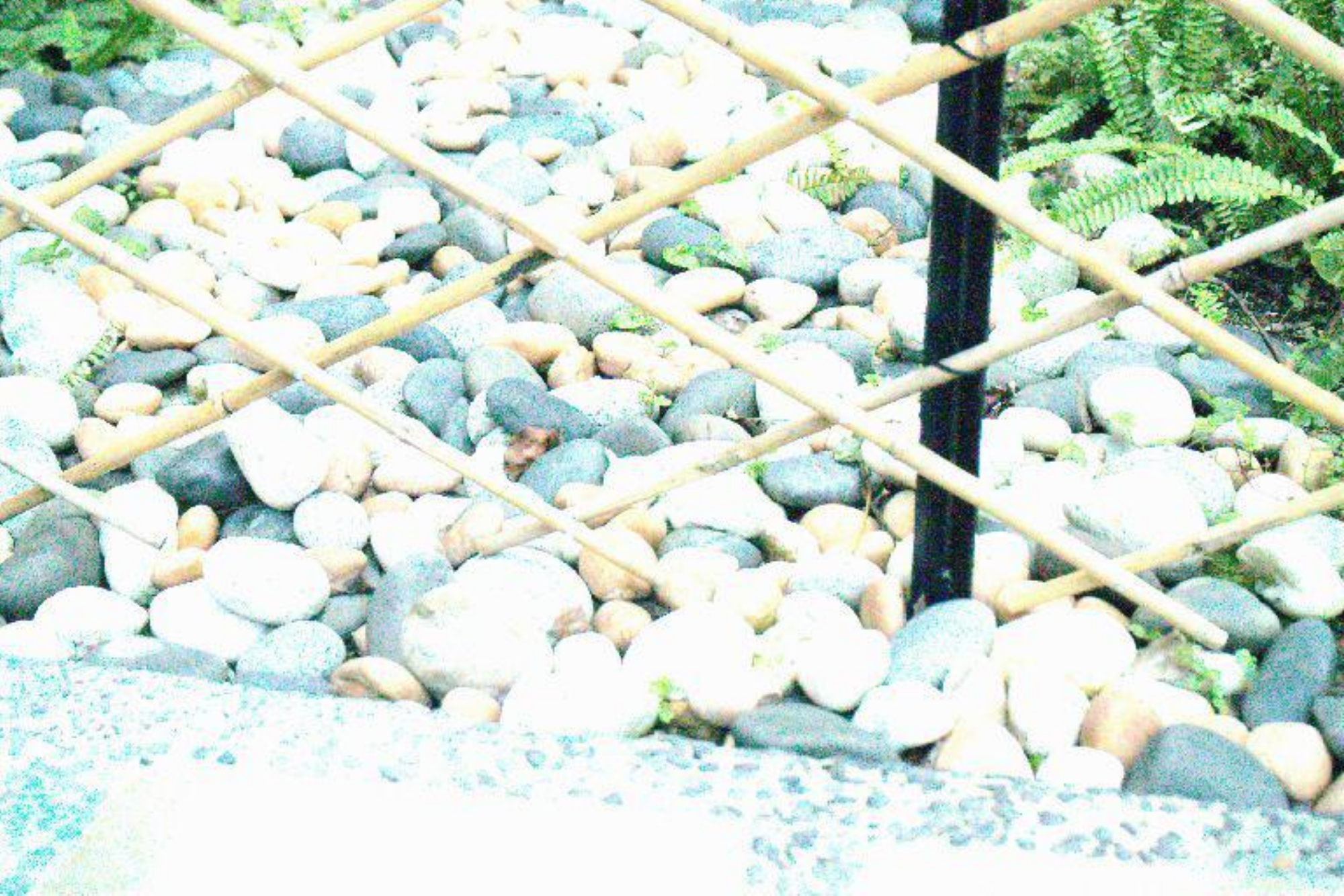} & 
     \includegraphics[width=0.198\linewidth]{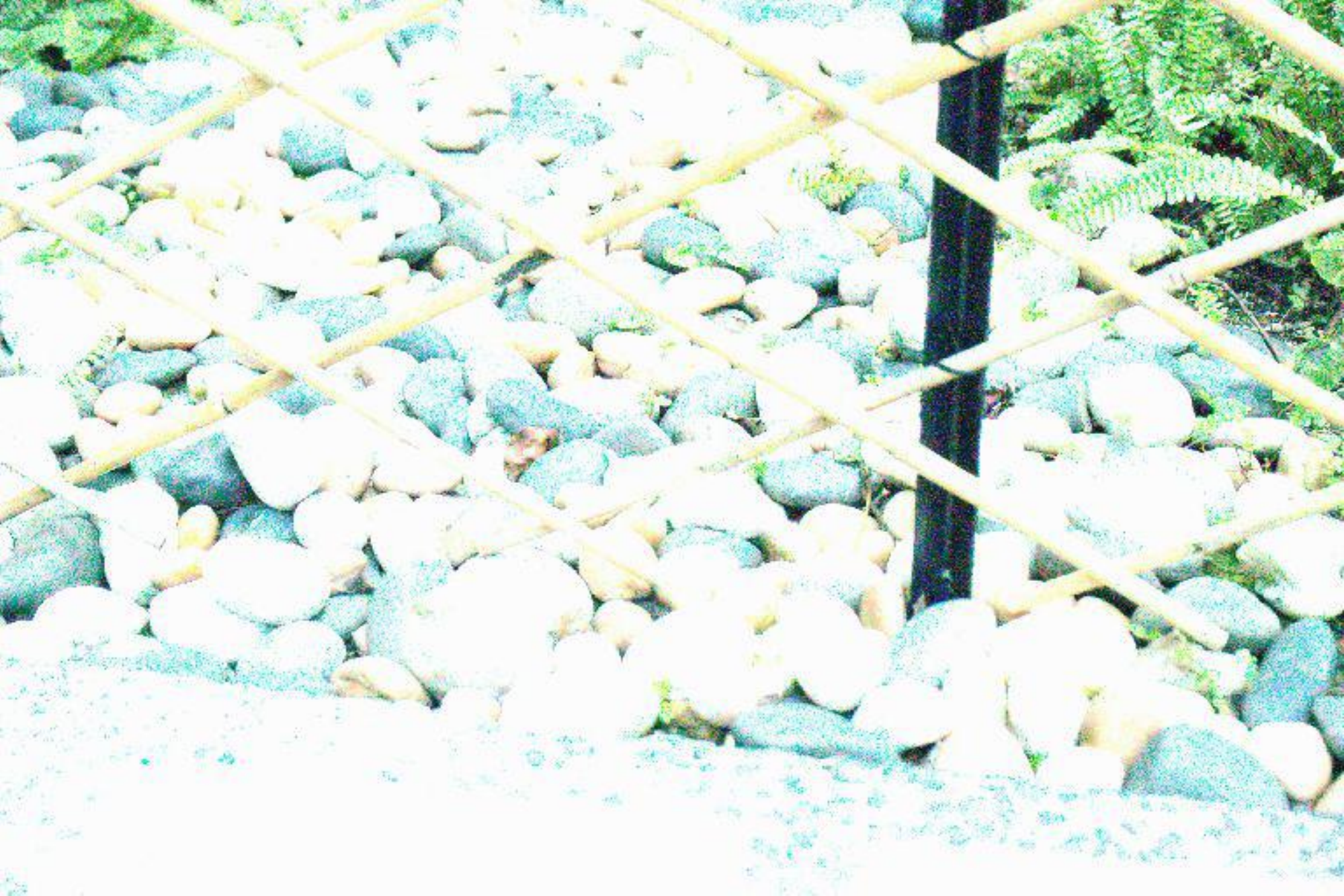} &
     \includegraphics[width=0.198\linewidth]{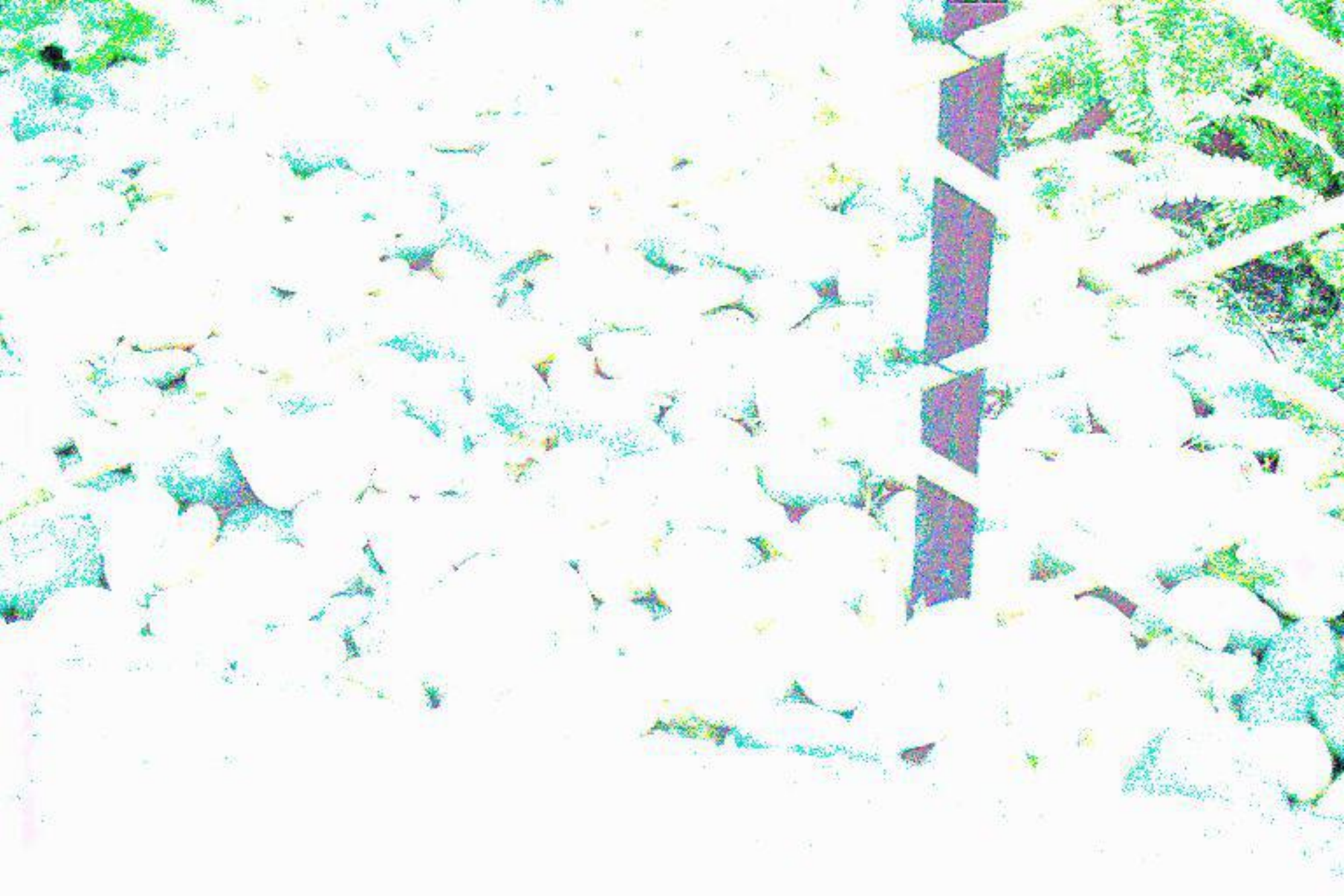} \vspace{-2mm}\\
      \scriptsize 0.208 & \scriptsize 0.254 &\scriptsize  0.410&\scriptsize 0.584 &\scriptsize 0.692\\
    \end{tabular}
    \caption{Defection scores (the smaller, the better) for the original and simulated images evaluated with~\cite{yu2018learning}. We pick nine representative simulated images for display. }
    \label{fig:auto-exposure}
\end{figure}

\mypara{Auto-exposure mode.} Traditional auto-exposure algorithms rely on light metering based on reference regions. However, metering is inaccurate when no gray objects exist for reference (for example, the snow scenes.) Our proposed model can provide a simulation environment for training the auto-exposure algorithm. We can train the selection algorithm by direct regression or by trial-and-error in reinforcement learning. Any aesthetic standards can be adopted for evaluating the quality of output images with new settings. To validate the feasibility of the proposed auto-exposure pipeline, we build a toy exposure selection model. We pre-defined 64 states with different camera settings for selection and simulate images for each state. We test two different standards for assessing the quality of the images: estimating aesthetic score~\cite{talebi2018nima} and detecting the image defections, including noise, exposure, and white balance~\cite{yu2018learning}. We then train a neural network by regression for predicting the score of each state.  In Fig.~\ref{fig:auto-exposure}, we show the simulation results and the final selected settings (more details in the \textbf{supplement}).

\begin{figure}[!t]
    \centering 
    \begin{tabular}{@{\hspace{1mm}}c@{\hspace{1mm}}c@{}}
    
    \small Original & \small Augmented \\
     \raisebox{-.2\height}{\includegraphics[width=0.49\linewidth]{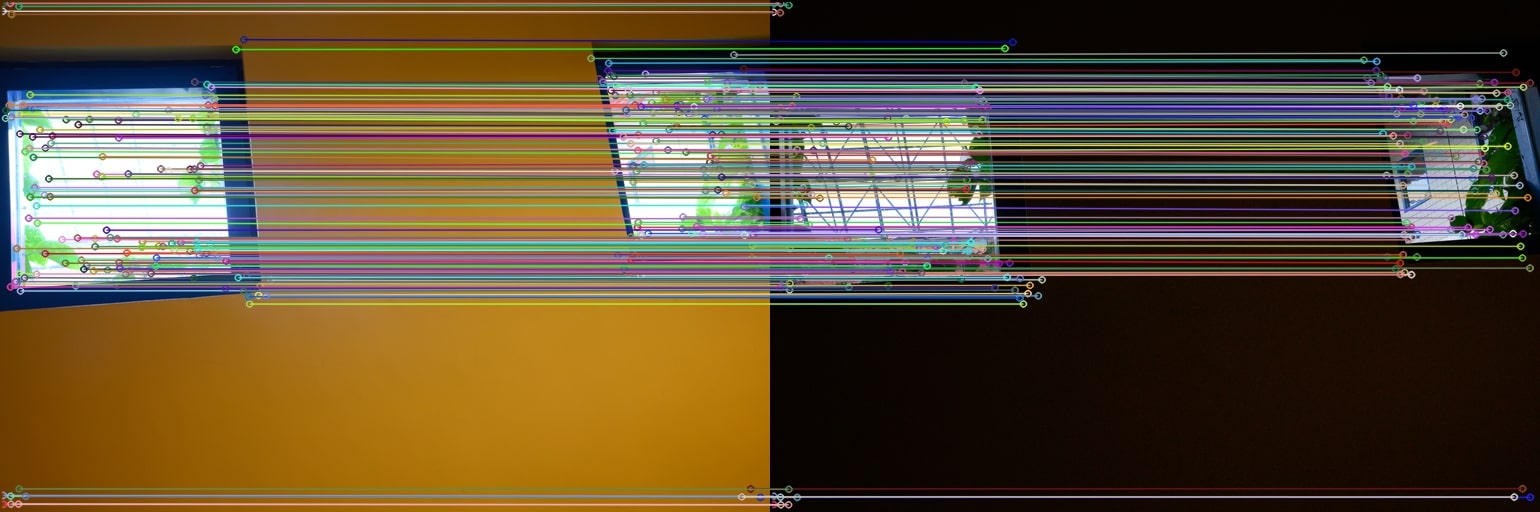}} &    \raisebox{-.2\height}{\includegraphics[width=0.49\linewidth]{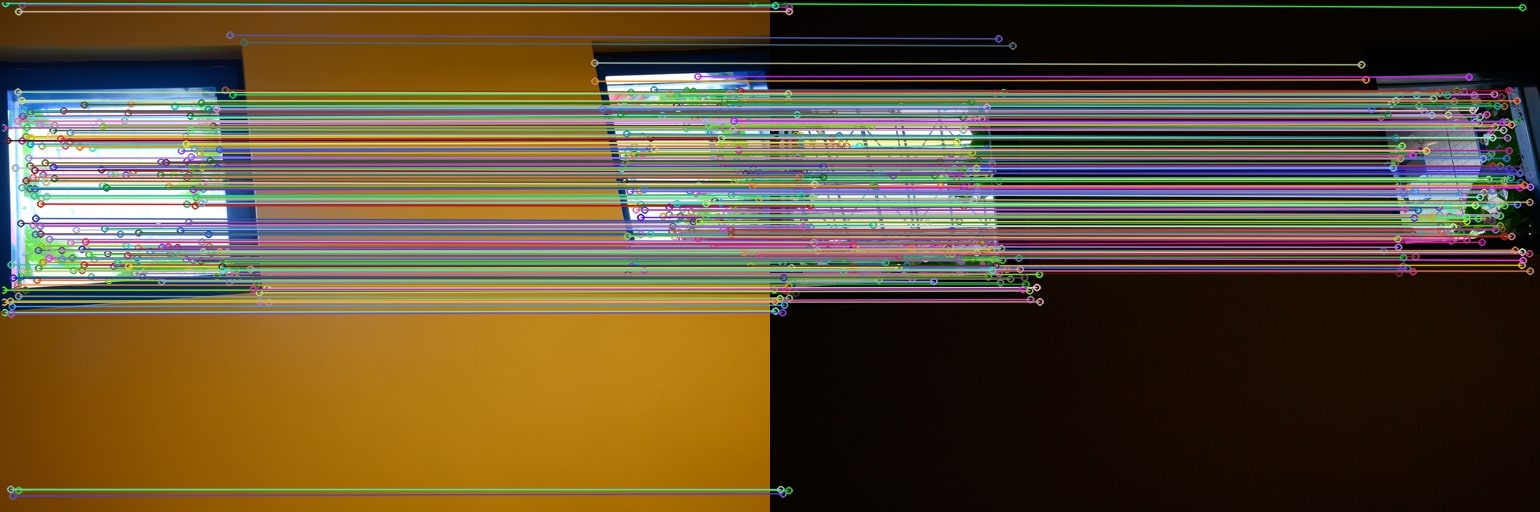}}   \\ 
    \raisebox{-.2\height}{\includegraphics[width=0.49\linewidth]{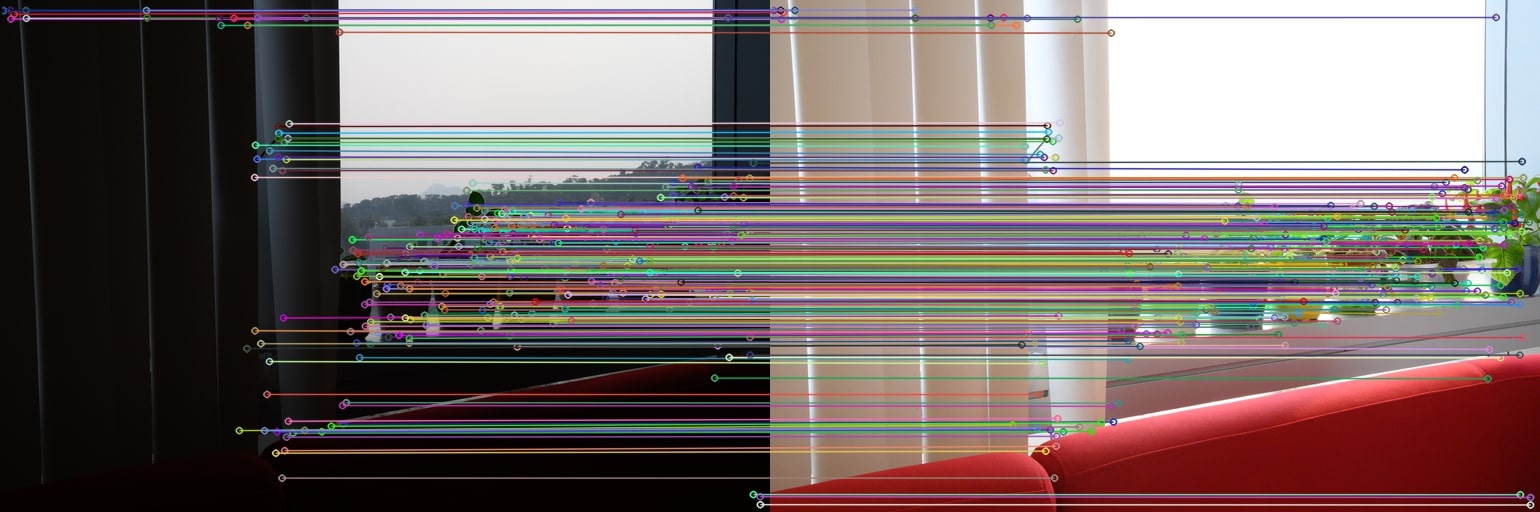}} &    \raisebox{-.2\height}{\includegraphics[width=0.49\linewidth]{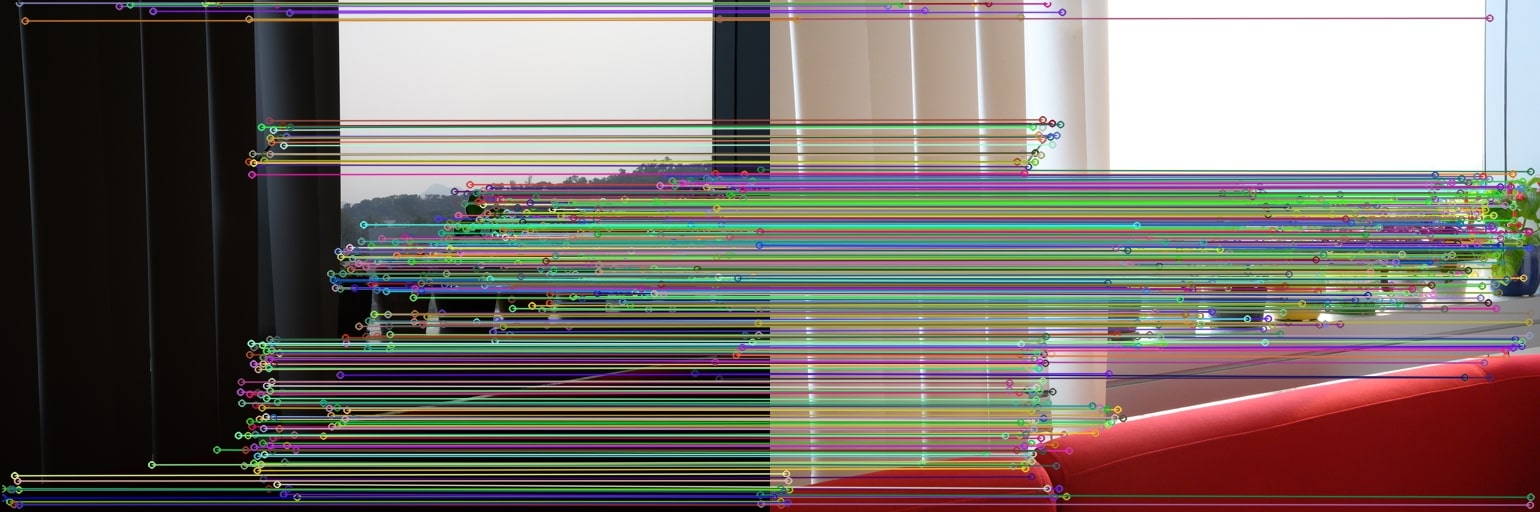}}  
    \end{tabular}
    \vspace*{1mm}
    \caption{Visual results of local feature matching. }
    
    \label{fig:detector}
\end{figure}

\mypara{Data augmentation.} We test the feasibility of using the synthesized images as augmented data for training local image features. We conduct experiments using D2-Net~\cite{dusmanu2019d2}, which is a local-feature extractor where the descriptor and detector are jointly trained. We use image pairs from the same sequences with homography transform to generate the ground truth matching points. Fifty scenes from the Nikon dataset are used for training, and five scenes are used for testing.  We compare the performance of two models: D2-Net trained with only the original augmentation (including color jittering) and D2-Net trained with further augmented data from the simulator. In Fig.~\ref{fig:detector}, the model trained with our augmented data finds more matches in extreme cases. The number of valid matches respectively increases from 253 to 371 and from 358 to 515 for the first and second examples. More details and quantitative results are in the \textbf{supplement}.

\section{Discussion}
In this paper, we systematically study the synthesis of one important component in the physical image pipeline: camera exposure settings.  We address this novel problem by deep models with physical prior. Experiments demonstrate promising results with correctly modified exposure, noise, and defocus blur.  This work also provides many opportunities for future work. The major limitation is that the model only considers static scenes and fails in handling the deblur tasks. In images where no blurry region exists as guidance, our model fails in enhancing the aperture. For noise simulation, we may also adopt a learning-based method~\cite{abdelhamed2019noise} rather than NLF functions. We expect future work to yield further improvements in the quality of simulated raw data.

{\small
\bibliographystyle{ieee_fullname}
\bibliography{cvpr}
}

\end{document}

%% file: vcl-shortcuts.tex
\usepackage{times}
\usepackage{epsfig}
\usepackage{graphicx}
\usepackage{float}
\usepackage{wrapfig}
\usepackage{amsmath,amssymb,amsthm}
\usepackage{algorithm,algorithmicx,algpseudocode}
\usepackage{bm,xspace}
\usepackage{comment}
\usepackage{verbatim}
\usepackage{multirow}
\usepackage{balance}
\usepackage{url}
\usepackage{booktabs}
\usepackage{etoolbox,siunitx}
\usepackage{calc}
\usepackage{pifont,hologo}
\usepackage[usenames, dvipsnames]{color}
\usepackage{nicefrac}

\usepackage[super]{nth}
\usepackage{nicefrac}
\robustify\bfseries

\DeclareMathSymbol{@}{\mathord}{letters}{"3B}

\newcommand\mypara[1]{\vspace{1mm}\noindent\textbf{#1}}

\def\latex/{\LaTeX}
\def\bibtex/{\hologo{BibTeX}}
